\documentclass[letterpaper, 10 pt, conference]{ieeeconf}  

\IEEEoverridecommandlockouts                              

\usepackage{cite}
\usepackage{amsmath,amssymb,amsfonts}
\usepackage{algorithmic}
\usepackage{graphicx}
\usepackage{textcomp}
\usepackage[dvipsnames]{xcolor}
\usepackage{multirow}
\usepackage{adjustbox}
\usepackage{caption}
\usepackage{subcaption}
\usepackage{balance}
\usepackage{float}
\usepackage{url}
\usepackage{hyperref}
\usepackage{multirow}
\usepackage{booktabs}
\usepackage{amssymb}
\usepackage{xcolor}

\newcommand{\namemethod}{V2V-LLM}
\newcommand{\namedataset}{V2V-QA}
\newcommand{\namevsplit}{V2V-split}
\newcommand{\namexsplit}{V2X-split}

\title{\LARGE \bf
\namemethod: Vehicle-to-Vehicle Cooperative Autonomous Driving with Multimodal Large Language Models
}

\author{
Hsu-kuang Chiu$^{1,2}$\quad Ryo Hachiuma$^1$\quad Chien-Yi Wang$^1$\quad Stephen F. Smith$^2$\quad Yu-Chiang Frank Wang$^1$\quad \\ Min-Hung Chen$^1$
\thanks{
$\textsuperscript{\rm 1}$NVIDIA,\quad$\textsuperscript{\rm 2}$Carnegie Mellon University}
\thanks{The authors thank Boyi Li, Zhiding Yu, Boris Ivanovic, and Marco Pavone from NVIDIA for valuable discussions and feedback.}
\thanks{This research was funded by NVIDIA, the CMU Safety21 University
Transportation Center, and CMU Robotics Institute.}
}

\begin{document}

\maketitle
\thispagestyle{empty}
\pagestyle{empty}

\begin{abstract}
Current autonomous driving vehicles rely mainly on their individual sensors to understand surrounding scenes and plan for future trajectories, which can be unreliable when the sensors are malfunctioning or occluded. To address this problem, cooperative perception methods via vehicle-to-vehicle (V2V) communication have been proposed, but they have tended to focus on perception tasks like detection or tracking. How those approaches contribute to overall cooperative planning performance is still under-explored. Inspired by recent progress using Large Language Models (LLMs) to build autonomous driving systems, we propose a novel problem setting that integrates a Multimodal LLM into cooperative autonomous driving, with the proposed Vehicle-to-Vehicle Question-Answering (\namedataset) dataset and benchmark. We also propose our baseline method Vehicle-to-Vehicle Multimodal Large Language Model (\namemethod), which uses an LLM to fuse perception information from multiple connected autonomous vehicles (CAVs) and answer various types of driving-related questions: grounding, notable object identification, and planning. Experimental results show that our proposed \namemethod~
can be a promising unified model architecture for performing various tasks in cooperative autonomous driving, and 
outperforms other baseline methods that use different fusion approaches.
Our work also creates a new research direction that can improve the safety of future autonomous driving systems. Our code and dataset are released to facilitate open-source research at \href{https://eddyhkchiu.github.io/v2vllm.github.io/}{https://eddyhkchiu.github.io/v2vllm.github.io/}.
\end{abstract}

\section{Introduction}
Autonomous driving technology has advanced significantly due to the evolution of deep learning algorithms,  and the release of large-scale real-world driving datasets and benchmarks~\cite{geiger2012kitti, caesar2019nuscenes, sun2020waymo}. However, the perception and planning systems of autonomous vehicles in daily operation rely mainly on their local LiDAR sensors and cameras to detect notable nearby objects and plan for future trajectories. This approach may encounter safety-critical problems when the sensors are occluded by nearby large objects.

To address this safety-critical issue, recent research proposes cooperative perception algorithms~\cite{chen2019fcooper, xu2022opencood, xu2022v2xvit, xu2022cobevt, chiu2024probabilistic, chiu2023selective} via vehicle-to-vehicle (V2V) communication. In cooperative driving scenarios, multiple \textit{Connected Autonomous Vehicles (CAVs)} driving nearby to each other share their perception information via V2V communication. The received perception data from multiple CAVs is then fused to generate better overall detection or tracking results.
A number of cooperative autonomous driving datasets have been released to the public, including simulated ones~\cite{xu2022opencood, li2022v2xsim, xu2022v2xvit, cui2022coopernaut} and real ones~\cite{xu2023v2v4real, xiang2024v2xreal, yu2022dair-v2x, zimmer2024tumtraf}. These datasets also establish benchmarks to evaluate the performance of cooperative perception algorithms. However, to date, cooperative driving research and datasets have mostly focused on perception tasks.
How these state-of-the-art cooperative perception models can be connected with the downstream planning models to generate good cooperative planning results is still under-explored. 

Other recent research has attempted to use LLM-based methods to build end-to-end perception and planning algorithms for an individual autonomous vehicle~\cite{sima2023drivelm, tian2024drivevlm, wang2025omnidrive, tian2024token, chen2024drivingwithllms, wang2023drivemlm} due to their common-sense reasoning and generalization ability from large-scale pre-trained data. These LLM-based models encode the raw sensor inputs
and answer driving-related perception and planning questions. These approaches have shown some promise but have not yet explored the benefits of cooperative perception and planning.

\begin{figure}[!t]
\centering
\includegraphics[width=0.47\textwidth]{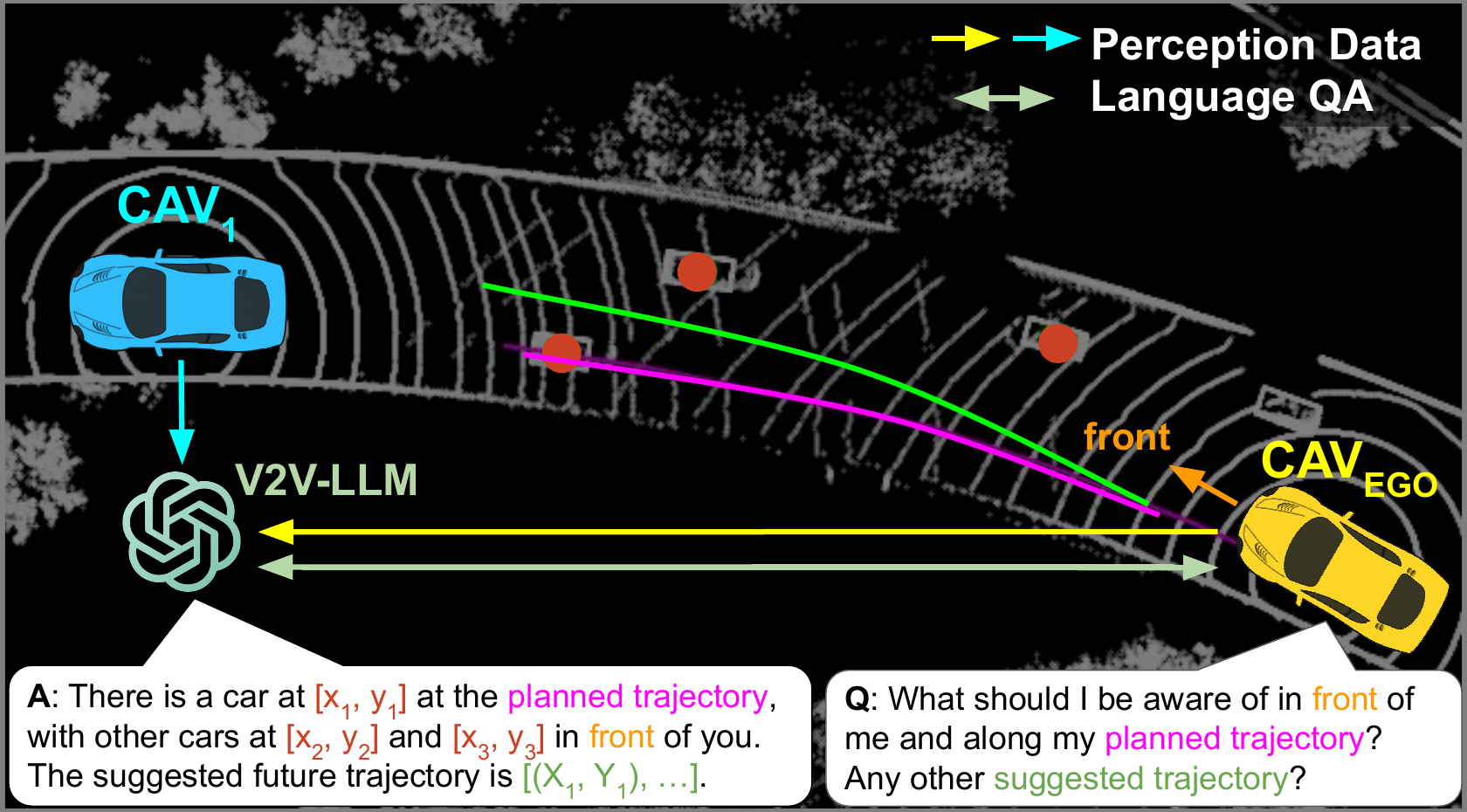}
\caption[]
        {Overview of our problem setting of LLM-based cooperative autonomous driving. All CAVs share their perception information with the LLM. Any CAV can ask the LLM a question to obtain useful information for driving safety.} 
        \vspace{-20pt}
        \label{fig:conceptual_only}
\end{figure}

\begin{table*}[t!]
\caption{
Comparison between our \namedataset~and recent related Autonomous Driving (AD) datasets. 
}
\vspace{-10pt}
\begin{center}
\begin{tabular}{l cccc rrr cc}
  \hline
  \hline
  Dataset & Publication & \# CAVs & RSU & Sim/Real & \# Frames & \# QA  & \# QA/frame & Point Cloud & Planning \\
  \hline
  \hline
  \scriptsize{\textit{Cooperative perception in AD}} \\
  OPV2V~\cite{xu2022opencood} & ICRA 2022 & 2-7 & - & Sim &11K & - & - & \checkmark  \\
   V2X-Sim~\cite{li2022v2xsim} & RA-L 2022 & 2-5 & \checkmark & Sim & 10K & - & - & \checkmark \\
  V2XSet~\cite{xu2022v2xvit} & ECCV 2022 & 2-5 & \checkmark & Sim & 11K & - & - & \checkmark \\
  DAIR-V2X~\cite{yu2022dair-v2x} &  CVPR 2022 & 1 & \checkmark & Real & 71K & - & - & \checkmark \\
  V2V4Real~\cite{xu2023v2v4real} & CVPR 2023 & 2 & - & Real & 20K & - & - & \checkmark \\
  TUMTrafV2X~\cite{zimmer2024tumtraf} & CVPR 2024 & 1 & \checkmark & Real & 2K & - & - & \checkmark \\
  V2X-Real~\cite{xiang2024v2xreal} & ECCV 2024 & 2 & \checkmark & Real & 33K & - & - & \checkmark \\
  \hline
  \scriptsize{\textit{LLM-based AD}} \\  
  NuScenes-QA~\cite{qian2023nuscenesqa} & AAAI 2024 & - & - & Real & 34K & 460K & 13.5 & \checkmark & \\
  Lingo-QA~\cite{marcu2023lingoqa} & ECCV 2024 & - & - & Real & 28K & 420K & 15.3 & & \checkmark \\
  DriveLM~\cite{sima2023drivelm} & ECCV 2024 & - & - & Sim+Real & 69K & 2M & 29.1 & & \checkmark \\
  TOKEN~\cite{tian2024token} & CoRL 2024 & - & - & Real & 28K & 434K & 15.5 & & \checkmark \\
  OmniDrive~\cite{wang2025omnidrive} & CVPR 2025 & - & - & Real & 34K & 450K & 13.2 & & \checkmark \\
  \hline
  \textbf{\namedataset~(Ours)}     & -  & 2 & \checkmark & Real & 48K & 1.45M & \textbf{30.2} & \checkmark & \checkmark \\
  \hline
  \vspace{-20pt}
\end{tabular}
\label{tab:dataset_comparison_all}
\end{center}
\vspace{-10pt}
\end{table*}
    
In this paper, we propose and explore a novel problem setting wherein LLM-based methods are used to build end-to-end perception and planning algorithms for \emph{Cooperative Autonomous Driving},
as illustrated in Figure \ref{fig:conceptual_only}. In this problem setting, we assume that there are multiple CAVs and a centralized LLM computing node. All CAVs share their individual perception information with the LLM. Any CAV can ask the LLM a question in natural language to obtain useful information for driving safety. 
To enable the study of this problem setting, we first create the \textbf{Vehicle-to-Vehicle Question-Answering (\namedataset)} dataset, built upon the V2V4Real~\cite{xu2023v2v4real} and V2X-Real~\cite{xiang2024v2xreal} cooperative perception datasets for autonomous driving. 
Our \namedataset~
includes \textbf{grounding} 
, \textbf{notable object identification} 
, and \textbf{planning}
question-answer pairs. 
There are several differences between our novel problem setting and other existing LLM-based driving research~\cite{tian2024token, wang2025omnidrive, sima2023drivelm, marcu2023lingoqa, qian2023nuscenesqa}. 
First, our LLM can fuse multiple perception information from different CAVs and provide answers to different questions from any CAV, rather than just serving a single self-driving car.
Second, our grounding questions are specially designed to focus on the potential occluded regions of each individual CAV. 
More differences between our \namedataset~and other related datasets are summarized in Table \ref{tab:dataset_comparison_all}.

To establish a benchmark for the \namedataset~dataset, we first propose a strong baseline method: \textbf{Vehicle-to-Vehicle Multimodal Large Language Model (\namemethod)} for cooperative autonomous driving, as illustrated in Figure \ref{fig:model}. Each CAV extracts its own perception features and shares them with \namemethod. The \namemethod~fuses the scene-level feature maps and object-level feature vectors, and then performs vision and language understanding to provide the answer to the input driving-related questions in \namedataset. 
We also compare \namemethod~with other baselines  corresponding to different feature fusion methods: \textit{no fusion}, \textit{early fusion}, and \textit{intermediate fusion}~\cite{xu2023v2v4real, xiang2024v2xreal, xu2022opencood, xu2022v2xvit, xu2022cobevt}. The results show that \namemethod~achieves better performance in the grounding, notable object identification, and planning tasks for the cooperative autonomous driving scenarios in general.

Our contribution can be summarized as follows:
\begin{itemize}
\item  We create and introduce the \namedataset~dataset to support the development and evaluation of LLM-based approaches to end-to-end cooperative autonomous driving.
\namedataset~includes grounding, notable object identification, and planning question-answering tasks.

\item We propose a baseline method \namemethod~for cooperative autonomous driving to provide an initial benchmark for \namedataset. This method fuses scene-level and object-level features  provided by multiple CAVs, and answers different CAV's driving-related questions. 

\item We create a benchmark for \namedataset~and show that \namemethod~outperforms other baselines on the grounding, notable object identification, and planning tasks in general, indicating the potential of \namemethod~to be a foundation model for cooperative autonomous driving.
  
\end{itemize}

\section{Related Work}

\subsection{Cooperative Perception in Autonomous Driving}
Cooperative perception~\cite{huang2024v2xcooperativeperceptionautonomous} algorithms were proposed to address the potential occlusion problem in individual autonomous vehicles. Pioneering work F-Cooper~\cite{chen2019fcooper} proposes the first intermediate fusion approach that merges feature maps to achieve good cooperative detection performance.
More recent work, AttFuse~\cite{xu2022opencood}, V2X-ViT~\cite{xu2022v2xvit}, and CoBEVT~\cite{xu2022cobevt} integrate attention-based models to aggregate features for cooperative detection and tracking. 

From a dataset perspective~\cite{liu2024asurveyon, yazgan2024collaborativeperceptiondatasetsautonomous}, simulation datasets OPV2V~\cite{xu2022opencood}, V2X-Sim~\cite{li2022v2xsim}, and V2XSet~\cite{xu2022v2xvit} were first generated for cooperative perception research. More recently, real datasets have been collected. V2V4Real~\cite{xu2023v2v4real} is the first worldwide available real vehicle-to-vehicle cooperative perception dataset with perception benchmarks. V2X-Real~\cite{xiang2024v2xreal}, DAIR-V2X~\cite{yu2022dair-v2x}, and TUMTraf-V2X~\cite{zimmer2024tumtraf} further include sensor data from roadside infrastructures.

Different from this group of research, our problem setting and proposed dataset include both perception and planning question-answering tasks for multiple CAVs. Our proposed \namemethod~also adopts a novel LLM-based fusion approach.

\subsection{LLM-based Autonomous Driving}
Language-based planning models~\cite{mao2023gpt, mao2023agentdriver, li2024drivingeverywhere} and more recent Multimodal Large Language Model (MLLM)-based approaches~\cite{tian2024token, wang2025omnidrive, sima2023drivelm, tian2024drivevlm, wang2023driveanywhere} encode the driving scene and ego-vehicle's state into text and visual features and use them as input to the LLM. Then the LLM generates text output including the suggested action or future trajectory. 

From a dataset perspective, several LLM-based autonomous driving datasets have been built on top of existing autonomous driving datasets. For example, 
NuPrompt~\cite{wu2023nuprompt}, NuScenes-QA~\cite{qian2023nuscenesqa}, and NuInstruct~\cite{ding2024nuinstruct}  
create captioning, perception, prediction, and planning QA pairs based on the NuScenes~\cite{caesar2019nuscenes} dataset.
DriveLM~\cite{sima2023drivelm} adopts real data from NuScenes~\cite{caesar2019nuscenes} and simulated data from CARLA~\cite{dosovitskiy2017carla} to have larger-scale and more diverse driving QAs.

Different from those LLM-based driving research that only supports individual autonomous vehicles, our problem setting and proposed \namedataset~dataset are designed for cooperative driving with multiple CAVs. 

\begin{figure*}[!t]
        \centering
        \begin{subfigure}[t]{0.32\textwidth}
            \centering 
            \includegraphics[width=\textwidth]{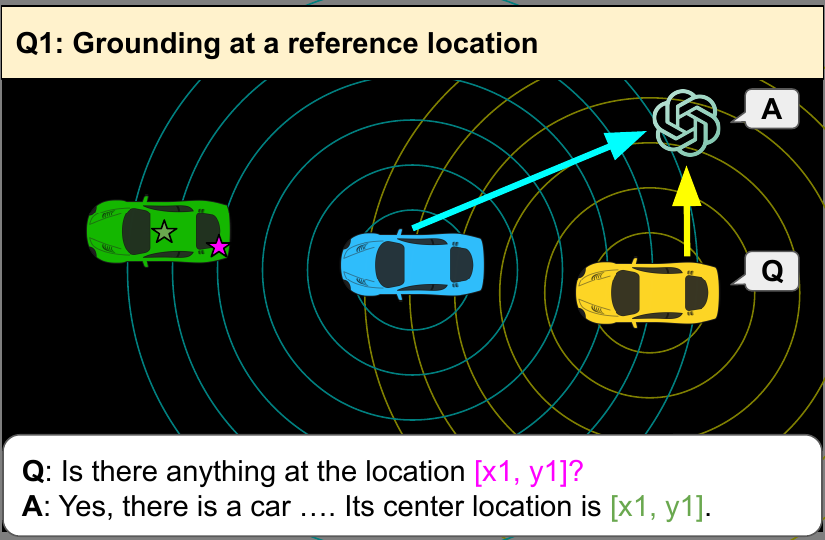}
            \caption[]%
            {{Q1: Grounding at a reference location.}}    
            \label{fig:q1_illustration}
        \end{subfigure}
        \hfill
        \begin{subfigure}[t]{0.32\textwidth}  
            \centering 
            \includegraphics[width=\textwidth]{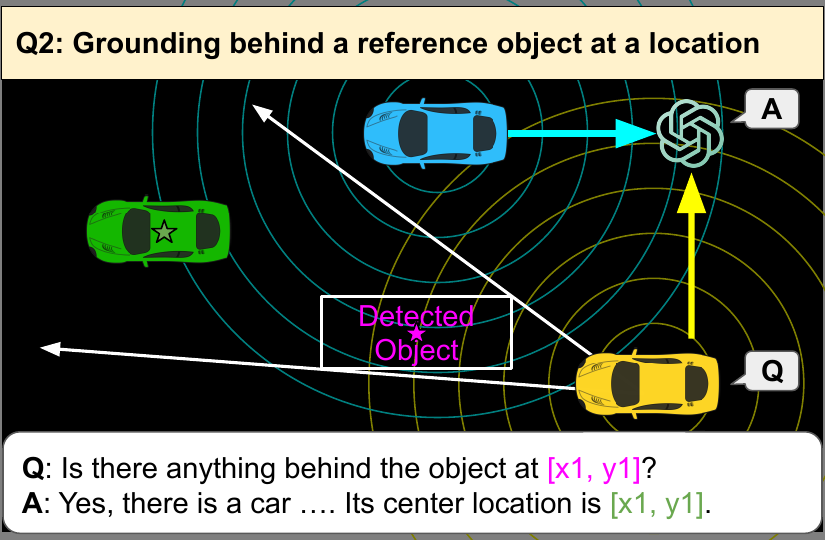}
            \caption[]%
            {{Q2: Grounding behind a reference object at a location.}}    
            \label{fig:q2_illustration}
        \end{subfigure}
        \hfill
        \begin{subfigure}[t]{0.32\textwidth}
            \centering 
            \includegraphics[width=\textwidth]{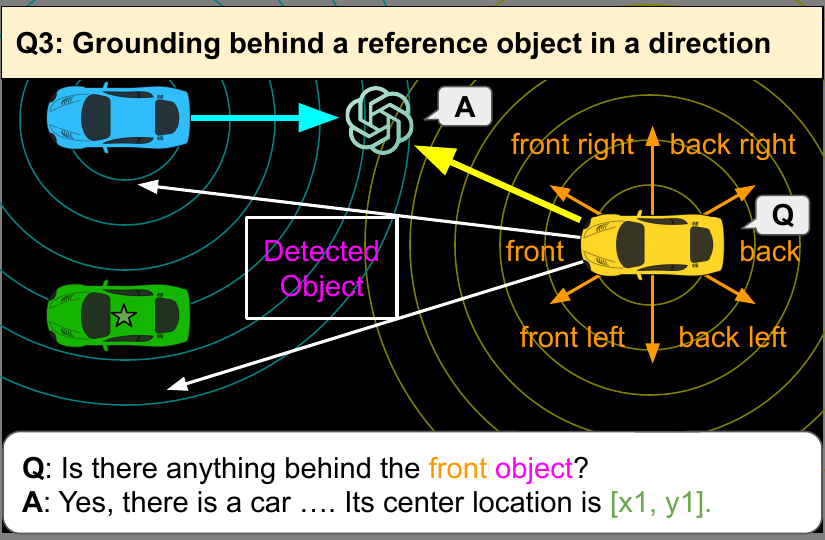}
            \caption[]%
            {{Q3: Grounding behind a reference object in a direction.}}    
            \label{fig:q3_illustration}
        \end{subfigure}

        \begin{subfigure}[t]{0.32\textwidth}
            \centering 
            \includegraphics[width=\textwidth]{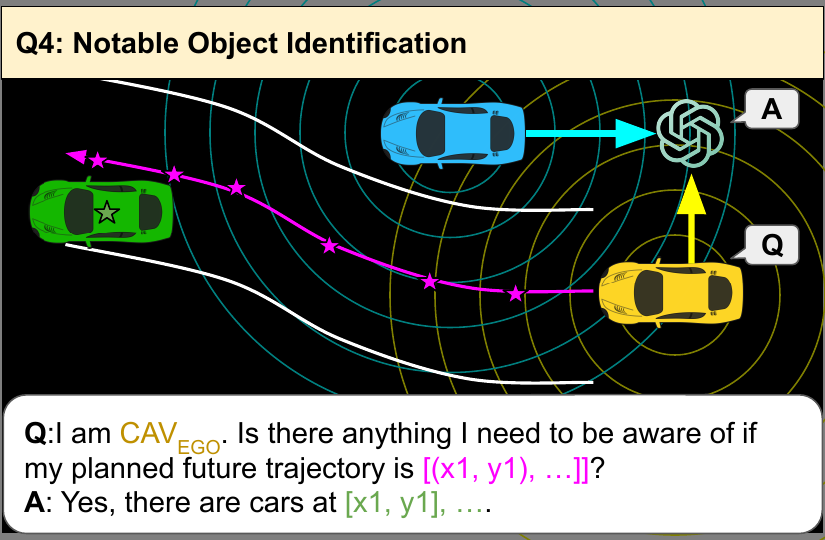}
            \caption[]%
            {{Q4: Notable object identification.}}    
            \label{fig:q4_illustration}
        \end{subfigure}
        \hspace*{0.0175\textwidth}%
        \begin{subfigure}[t]{0.32\textwidth}
            \centering 
            \includegraphics[width=\textwidth]{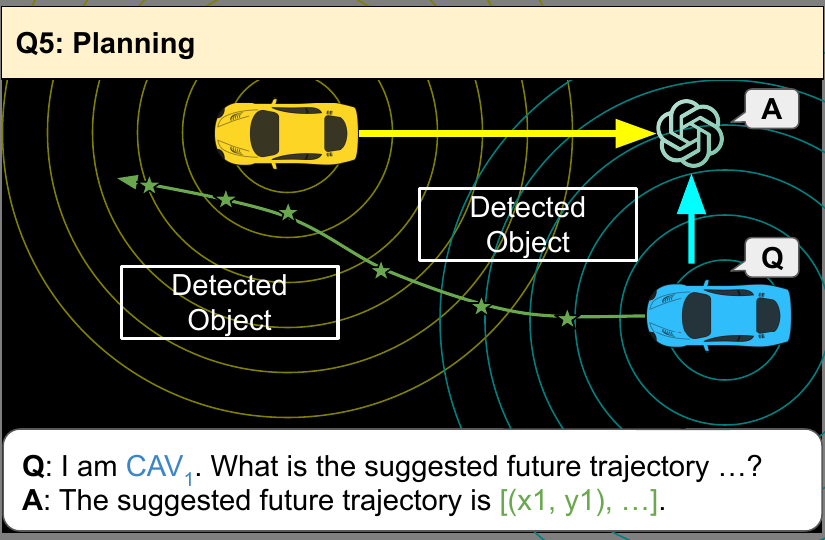}
            \caption[]%
            {{Q5: Planning.}}    
            \label{fig:q5_illustration}
        \end{subfigure}
        \hfill
        
        \caption[]
        {
        Illustration of \namedataset's $5$ types of QA pairs. The arrows pointing at LLM indicate the perception data from CAVs.
        } 
        \label{fig:all_q_illustration}
        \vspace{-15pt}
\end{figure*}

\section{\namedataset~Dataset}
To enable the research in our proposed novel problem setting: LLM-based cooperative autonomous driving, we create the \textbf{Vehicle-to-Vehicle Question-Answering (\namedataset)} dataset to benchmark different models' performance on fusing perception information and answering safety-critical driving-related questions.

\subsection{Problem Setting}
Our proposed V2V cooperative autonomous driving with LLM problem is illustrated in Figure \ref{fig:conceptual_only}. In this setting, we assume there are multiple Connected Autonomous Vehicles (CAVs) and a centralized LLM computing node. All CAVs share their individual perception information, such as scene-level or object-level features, with the centralized LLM. Any CAV can ask the LLM a question in natural language to obtain information for driving safety. The LLM aggregates the received perception information from multiple CAVs and provides a natural language answer to the CAV's question. In this research, the questions and answers include \textbf{grounding (Q1-3)}, \textbf{notable object identification (Q4)}, and \textbf{planning (Q5)}, as illustrated in Figure \ref{fig:all_q_illustration}. 

\subsection{Dataset Details}
Our \namedataset~dataset contains two splits: \textbf{\namevsplit}~and \textbf{\namexsplit}, which are built on top of V2V4Real~\cite{xu2023v2v4real} and V2X-Real~\cite{xiang2024v2xreal} datasets, respectively. These base datasets are collected by driving two vehicles with LiDAR sensors simultaneously near to each other.
These datasets also includes 3D bounding box annotations for other objects in the driving scenes. In V2V4Real~\cite{xu2023v2v4real}, the training set has $32$ driving sequences and a total of $7105$ frames of data per CAV, and the testing set has $9$ driving sequences and a total of $1993$ frames of data per CAV. In V2X-Real~\cite{xiang2024v2xreal}, the training set has $43$ driving sequences and a total of $5772$ frames of data per CAV, and the testing set has $9$ driving sequences and a total of $1253$ frames of data per CAV. The frame rate is $10$Hz. In V2X-Real~\cite{xiang2024v2xreal}, some driving scenes also provide LiDAR point clouds from roadside infrastructures. We also include them as perception inputs to the LLM with the same approach as using CAVs' LiDAR point clouds. We follow the same training and testing settings from V2V4Real~\cite{xu2023v2v4real} and V2X-Real~\cite{xiang2024v2xreal} when building our \namevsplit~and \namexsplit.
Table \ref{tab:dataset_stats} summarizes the numbers of QA pairs in our proposed \namedataset's \namevsplit~and \namexsplit. We have $1.45$M QA pairs in total and $30.2$ QA pairs per frame on average. 
 
\begin{table}[!t]
\caption{
Dataset statistics of our \namedataset's \namevsplit~and \namexsplit. Q1: Grounding at a reference location. Q2: Grounding behind a reference object at a location. Q3: Grounding behind a reference object in a direction. Q4: Notable object identification. Q5: Planning.
\vspace{-10pt}
}
\small
\setlength{\tabcolsep}{4pt}
\begin{center}
\begin{tabular}{c | rr | rr | r}
  \hline
  \hline
  \multirow{2}{*}{QA type} & \multicolumn{2}{c|}{\namevsplit} & \multicolumn{2}{c|}{\namexsplit} & Total \\
  & Training & Testing & Training & Testing \\
  \hline
  \hline

  

  Q1 & 354820 & 121383 & 495290 & 128711 & 1100204 \\
  Q2 &  35700 &  13882 & 167694 &  35233 &  252509 \\
  Q3 &  14339 &   5097 &  28740 &   6465 &   54641 \\
  Q4 &  12290 &   3446 &   6274 &   1708 &   23718 \\
  Q5 &  12290 &   3446 &   6274 &   1708 &   23718 \\
  \hline
  Total & 429439 & 147254 & 704272 & 173825 & 1454790
\end{tabular}
\vspace{-20pt}
\label{tab:dataset_stats}
\end{center}
\end{table}

\subsection{Question and Answer Pairs Curation}
For each frame of V2V4Real~\cite{xu2023v2v4real} and V2X-Real~\cite{xiang2024v2xreal} datasets, we create $5$ different types of QA pairs, including $3$ types of grounding questions,  $1$ type of notable object identification question, and $1$ type of planning question. These QAs are designed for cooperative driving scenarios. To generate instances of these QA pairs, we use V2V4Real~\cite{xu2023v2v4real} and V2X-Real~\cite{xiang2024v2xreal}'s ground-truth bounding box annotations, each CAV's ground-truth trajectories, and individual detection results as the source information. Then we use different manually designed rules based on the geometric relationship among the aforementioned entities and text templates to generate our QA pairs. The text template can be seen in Figures \ref{fig:qualitative_result_grounding} and  \ref{fig:qualitative_result_notable_planning}. The generation rule of each QA type is described as follows.

\noindent\textbf{Q1. Grounding at a reference location (\ref{fig:q1_illustration}):}
In this type of question, we ask the LLM to identify whether an object that occupies a specific query 2D location exists. 
If so, the LLM is expected to provide the center location of the object. Otherwise, the LLM should indicate that there is nothing at the reference location. 
We use the center locations of ground-truth boxes and every CAV's individual detection result boxes as the query locations in the questions. By doing so, we can focus more on evaluating each model's cooperative grounding ability on the potential false positive and false negative detection results.


\noindent\textbf{Q2. Grounding behind a reference object at a location (\ref{fig:q2_illustration}):}
When a CAV's field of view is occluded by a nearby large detected object, this CAV may want to ask the centralized LLM to determine whether there exists any object behind that occluding large object given the fused perception information from all CAVs. If so, the LLM is expected to return the object's location and the asking CAV may need to drive more defensively or adjust its planning. Otherwise, the LLM should indicate that there is nothing behind the reference object.
We use the center location of each detection result box as the query locations in these questions. We draw a sector region based on the relative pose of the asking CAV and the reference object, and select the closest ground-truth object in the region as the answer.


\noindent\textbf{Q3. Grounding behind a reference object in a direction (\ref{fig:q3_illustration}):}
We further challenge the LLM on language and spatial understanding ability by replacing Q2's reference 2D location with a reference directional keyword.
We first get the closest detection result box in each of the $6$ directions of a CAV as the reference object. Then we follow the same approach in Q2 to get the closest ground-truth box in the corresponding sector region as the answer.


\noindent\textbf{Q4. Notable object identification (\ref{fig:q4_illustration}):}
The aforementioned grounding tasks are intermediate tasks in the autonomous driving pipeline. More critical abilities of autonomous vehicles involve both identifying notable objects near planned future trajectories and adjusting future planning to avoid potential collisions. 
We extract $6$ waypoints from the ground-truth trajectory in the next $3$ seconds as the reference future waypoints in the questions. Then we get, at most, the $3$ closest ground-truth objects within $10$ meters of the reference future trajectory as the answer.


\noindent\textbf{Q5. Planning (\ref{fig:q5_illustration}):}
Planning is important because the ultimate goal of autonomous vehicles is to navigate through complex environments safely and avoid any potential collision in the future. 
To generate the planning QAs, we extract $6$ future waypoints, evenly distributed in the next $3$ seconds, from each CAV's ground-truth future trajectory as the answer. 
Our \namedataset's planning task is more challenging than other NuScenes~\cite{caesar2019nuscenes}-based LLM-driving related works for a couple reasons. First, we support multiple CAVs in cooperative driving scenarios. The LLM model needs to provide different answers depending on which CAV is asking, while prior works only need to generate planning results for a single autonomous vehicle. Second, our \namedataset's ground-truth planning trajectories are more diverse. \namedataset~contains both urban and highway driving scenarios, while NuScenes~\cite{caesar2019nuscenes} only includes urban driving scenarios. 


\subsection{Evaluation Metrics}
We follow prior works~\cite{tian2024token, wang2025omnidrive}'s approach to evaluate model performance. For the grounding questions (Q1, Q2, Q3) and the notable object identification question (Q4), the evaluation metric is F1 score.
For the planning question (Q5), the evaluation metrics are L2 errors and collision rates. 


\begin{figure}[!t]
\centering
\includegraphics[width=0.47\textwidth]{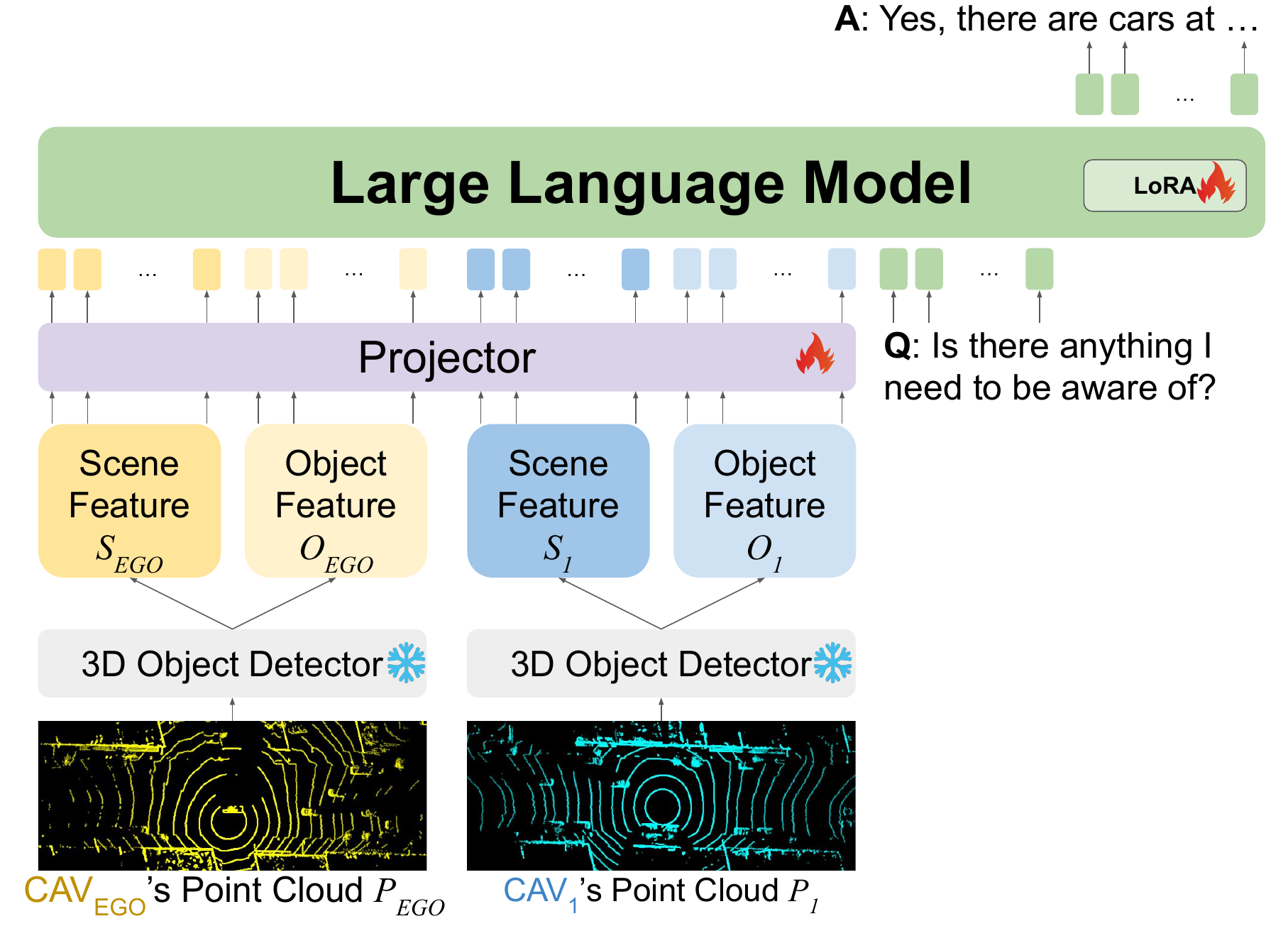}
\caption[]
        {Model diagram of our proposed \namemethod~for cooperative autonomous driving.}         
        \label{fig:model}
        \vspace{-15pt}
\end{figure}

\section{\namemethod}

We propose a competitive baseline model, \textbf{\namemethod}, for this LLM-based collaborative driving problem, as shown in Figure \ref{fig:model}. Our model is a Multimodal LLM (MLLM) that takes the individual perception features of every CAV as the vision input, a question as the language input, and generates an answer as the language output. 

\subsection{LiDAR-based Input Features}
For extracting the perception input features, each CAV applies a 3D object detection model to its individual LiDAR point cloud: $P_{EGO}$ and $P_{1}$. We extract the scene-level feature map $S_{EGO}$ and $S_{1}$ from the 3D object detection model and transform the 3D object detection results as the object-level feature vectors $O_{EGO}$ and $O_{1}$. Following prior works V2V4Real~\cite{xu2023v2v4real} and V2X-Real~\cite{xiang2024v2xreal}, we use PointPillars~\cite{lang2019pointpillar} as the 3D object detector for fair comparisons.

\subsection{LiDAR-based Multimodal LLM}
\noindent\textbf{Model architecture:} We utilize LLaVA~\cite{liu2023llava} to develop our MLLM, given its superior performance on visual question-answering tasks. However, since the perception features of our cooperative driving tasks are LiDAR-based instead of RGB images used by LLaVA~\cite{liu2023llava}, we use a LiDAR-based 3D object detector as the point cloud feature encoder, as described in the previous section, instead of LLaVA~\cite{liu2023llava}'s CLIP~\cite{radford2021clip} image feature encoders.
We then feed the resulting features to a multi-layer perceptron-based projector network for feature alignment from the point cloud embedding space to the language embedding space. The aligned perception features are the input perception tokens digested by the LLM together with the input language tokens from the question. Finally, the LLM aggregates the perception information from all CAVs and returns an answer based on the question. 

\noindent\textbf{Training:} We use 8 NVIDIA A100-80GB GPUs to train our model. Our \namemethod~uses LLaVA-v1.5-7b~\cite{liu2023llava}'s Vicuna~\cite{chiang2023vicuna} as the LLM backbone. To train our model, we initialize it by loading the pre-trained LLaVA-v1.5-7b~\cite{liu2023llava}'s checkpoint. We freeze the LLM and the point cloud feature encoder, and finetune the projector and the LoRA~\cite{hu2022lora} parts of the model. During training, we use batch size 32. Adam optimizer is adopted for training with a starting learning rate $2e-5$ and a cosine learning rate scheduler with a 3\% warm-up ratio. For all other training settings and hyperparameters, we use the same ones from LLaVA-v1.5-7b~\cite{liu2023llava}.

\section{Experiment}

\subsection{Baseline Methods}
We follow V2V4Real~\cite{xu2023v2v4real} and V2X-Real~\cite{xiang2024v2xreal} to establish a benchmark for our proposed \namedataset~dataset with experiments on baseline methods using different fusion approaches: \textbf{no fusion}, \textbf{early fusion}, \textbf{intermediate fusion}, and our proposed baseline, \textbf{LLM fusion} (\ref{fig:model}). All baseline methods also adopt the same projector and LLM architecture as in our \namemethod~but with different point cloud feature encoders. 
In some driving sequences of \namexsplit~that have point clouds from roadside infrastructures, we include them as perception input as well in the same way as using CAVs' point clouds.

\noindent \textbf{No fusion}: Only a single CAV's LiDAR point cloud is fed to a single 3D object detector to extract the scene-level and object-level features as the LLM's visual input. The performance is expected to be worse than all other cooperative perception approaches.

\noindent \textbf{Early fusion}: The LiDAR point cloud from two CAVs is merged first. Then the merged point cloud is used as input to a 3D object detector to extract the visual features as the visual input to the LLM. 
This approach requires much higher communication cost and is less practical for deployment on real-world autonomous vehicles.

\noindent \textbf{Intermediate fusion}: Prior work CoBEVT~\cite{xu2022cobevt}, V2X-ViT~\cite{xu2022v2xvit}, and AttFuse~\cite{xu2022opencood} propose  cooperative detection models that merge feature maps from multiple CAVs via attention mechanisms. Such approaches require less communication cost and can still achieve good performance. In our benchmark, we extract both of the scene-level and object-level features from those cooperative detection models as the input tokens to the LLM.

\noindent \textbf{LLM fusion}: We categorize our proposed \namemethod~as a new type of fusion method, \textit{LLM fusion}, which lets each CAV perform its individual 3D object detection to extract the scene-level feature maps and object-level feature vectors, and uses the Multimodal LLM to fuse the features from multiple CAVs. This approach is related to the traditional \textit{late fusion} method that performs individual 3D object detection and aggregates the results by non-maximum suppression (NMS). Instead of applying NMS, our method adopts LLM to perform more tasks than just detection.

\begin{table*}[t!]
\caption{
\namemethod's testing performance in \namedataset's \namevsplit~and \namexsplit~in comparison with baseline methods. Q1: Grounding at a reference location. Q2: Grounding behind a reference object at a location. Q3: Grounding behind a reference object in a direction. Q\textsubscript{Gr}: Average of grounding (Q1, Q2, and Q3). Q4: Notable object identification. Q5: Planning. L2: L2 distance error. CR: Collision rate. Comm: Communication cost. In each column, the \textbf{best} results are in boldface, and the \underline{second-best} results are in underline. 
\vspace{-10pt}
}
\small
\setlength{\tabcolsep}{2pt}
\begin{center}
\begin{tabular}{l ccccccc ccccccc c}
  \hline
  \hline
  \multirow{3}{*}{Method} &
  \multicolumn{7}{c}{\namevsplit} & \multicolumn{7}{c}{\namexsplit} & \multirow{3}{*}{Comm(MB) $\downarrow$} \\

  \cmidrule(lr){2-8} \cmidrule(lr){9-15}
  & \multicolumn{1}{c}{Q1} & \multicolumn{1}{c}{Q2} & \multicolumn{1}{c}{Q3} & \multicolumn{1}{c}{Q\textsubscript{Gr}} & \multicolumn{1}{c}{Q4} & \multicolumn{2}{c}{Q5} & \multicolumn{1}{c}{Q1} & \multicolumn{1}{c}{Q2} & \multicolumn{1}{c}{Q3} & \multicolumn{1}{c}{Q\textsubscript{Gr}} & \multicolumn{1}{c}{Q4} & \multicolumn{2}{c}{Q5} \\
  
  \cmidrule(lr){2-2} \cmidrule(lr){3-3} \cmidrule(lr){4-4} \cmidrule(lr){5-5} \cmidrule(lr){6-6} \cmidrule(lr){7-8} \cmidrule(lr){9-9} \cmidrule(lr){10-10} \cmidrule(lr){11-11} \cmidrule(lr){12-12} \cmidrule(lr){13-13} \cmidrule(lr){14-15}
  &
  F1 $\uparrow$ & F1 $\uparrow$ & F1 $\uparrow$ & F1 $\uparrow$ & F1 $\uparrow$ &  L2 (m) $\downarrow$ & CR (\%) $\downarrow$ & F1 $\uparrow$ & F1 $\uparrow$ & F1 $\uparrow$ & F1 $\uparrow$ & F1 $\uparrow$ &  L2 (m) $\downarrow$ & CR (\%) $\downarrow$\\
  \hline
  \hline
  \textit{No Fusion}         & 66.6 & 22.6 & 17.2 & 35.5 & 47.3 & 6.55 & 4.57 & 55.7 & 21.4 & 25.2 & 34.1 & 64.4 & 2.31 & 9.21 & \textbf{0} \\
  \textit{Early Fusion}      & \textbf{73.5} & 23.3 & 20.8 & 39.2 & 53.9 & \underline{6.20} & \underline{3.55} & \underline{59.7} & 23.3 & 26.1 & 36.4 & \underline{67.6} & \underline{2.12} & 8.61 & 1.9208 \\
  \hline
  \scriptsize{\textit{Intermediate Fusion}} \\ 
  AttFuse~\cite{xu2022opencood}         & 70.7 & 26.4 & 18.4 & 38.5 & 56.9 & 6.83 & 4.12 & 58.9 & 23.9 & \underline{26.3} & 36.4 & 65.9 & 2.19 & \underline{8.39} & \underline{0.4008} \\
  V2X-ViT~\cite{xu2022v2xvit}           & 70.8 & 28.0 & \textbf{22.6} & 40.5 & \underline{57.6} & 7.08 & 4.33 & 59.6 & \underline{24.2} & 26.1 & \underline{36.6} & 65.0 & 2.29 & 8.86 & \underline{0.4008} \\
  CoBEVT~\cite{xu2022cobevt}            & \underline{72.2} & \underline{29.3} & \underline{21.3} & \textbf{40.9} & \underline{57.6} & 6.72 & 3.88 & - & - & -& - & - & - & - & \underline{0.4008} \\
  \hline
  \scriptsize{\textit{LLM Fusion}} \\
  \namemethod~(Ours)     & 70.0 & \textbf{30.8} & 21.2 & \underline{40.7} & \textbf{59.7} & \textbf{4.99} & \textbf{3.00} & \textbf{60.5} & \textbf{25.3} & \textbf{26.7} & \textbf{37.5} & \textbf{69.3} & \textbf{1.71} & \textbf{6.89} & 0.4068 \\
  \hline
\end{tabular}
\label{tab:all_v2v_v2x_result}
\end{center}
\vspace{-10pt}
\end{table*}


\subsection{Quantitative Results}

\subsubsection{Performance}

\noindent \textbf{Grounding}:
Our \namemethod~and baseline methods' performance on \namedataset's $3$ types of grounding questions can be seen in Table \ref{tab:all_v2v_v2x_result} for \namevsplit~and \namexsplit, respectively. CoBEVT~\cite{xu2022cobevt} is not included in \namexsplit's result because V2X-Real~\cite{xiang2024v2xreal} does not release its CoBEVT~\cite{xu2022cobevt} baseline model. In average, \namemethod~ achieves similar performance in \namevsplit~and outperforms all other baseline methods in \namexsplit. Such results indicate that our \namemethod~has a promising capability of fusing perception features from multiple CAVs to answer grounding questions.

\noindent \textbf{Notable Object Identification}:
Table \ref{tab:all_v2v_v2x_result} show the performance on the notable object identification task (Q4). Our proposed \namemethod~outperforms all other methods in both \namevsplit~and \namexsplit. Compared with the aforementioned grounding tasks, this notable object identification task requires more spatial understanding ability to identify the objects close to the planned future waypoints. For such a task, our \namemethod, which lets the Multimodal LLM perform both perception feature fusion and question answering, achieves the best results.

\noindent \textbf{Planning}:
Table \ref{tab:all_v2v_v2x_result} show the performance of the planning task (Q5) for \namevsplit~and \namexsplit, respectively. Our proposed \namemethod~outperforms other methods in this safety-critical task to generate a future trajectory that aims to avoid potential collisions. 

\subsubsection{Communication Cost and Scaling Analysis}
In our centralized setting, each CAV sends one scene-level feature map ($\leq 0.2$MB), one set of individual object detection result parameters ($\leq 0.003$MB), one question ($\leq 0.0002$MB) to the LLM computing node and receives one answer ($\leq 0.0002$MB) at each timestep.  If there are $N_v$ CAVs and each asks $N_q$ questions, the communication cost of each CAV is $(0.2 + 0.003 + (0.0002 + 0.0002)N_q) = (0.203 + 0.0004N_q)$ MB, and the cost of the LLM is $(0.2 + 0.003 + (0.0002 + 0.0002)N_q) N_v = (0.203N_v + 0.0004N_qN_v)$ MB, as shown in Table \ref{tab:communication_cost}.
Note that each CAV only needs to send the same features to the LLM computing node once at each timestep because the LLM node can save and reuse them to answer multiple questions from the same or different CAVs at the same timestep.

\begin{table}[!t]
\caption{
Communication cost (MB) and scaling analysis. $N_v$: number of CAVs. $N_q$: number of questions asked by each CAV at each timestep.
\vspace{-10pt}
}
\small
\setlength{\tabcolsep}{3pt}
\begin{center}
\begin{tabular}{l c c }
  \hline
  \hline
  Setting & Each CAV & Centralized LLM  \\
  \hline
  \hline
  Centralized  & $0.203 + 0.0004N_q$ & $0.203N_v + 0.0004N_qN_v$ \\
  \hline
\end{tabular}
\label{tab:communication_cost}
\end{center}
\vspace{-10pt}
\end{table}

\subsubsection{Summary}
In general, \namemethod~achieves the best results in the notable object identification and planning tasks, which are critical in autonomous driving applications. \namemethod~also achieves competitive results in the grounding tasks. In terms of communication costs, \namemethod~only increases communication costs by $1.5\%$ in comparison to other intermediate fusion baseline methods. 

\subsection{Comparison to Non-LLM baseline}
To compare our \namemethod~with non-LLM baseline in \namedataset, for perception, we use the detection results from V2V4Real~\cite{xu2023v2v4real}'s best cooperative detection model CoBEVT~\cite{xu2022cobevt} checkpoint (the same one as the feature extractor in our experiment) and evaluate them in our Q1 (grounding at a location). For planning, since there is no prior cooperative planning research on V2V4Real~\cite{xu2023v2v4real}, we use the BEV features from the same CoBEVT~\cite{xu2022cobevt} model checkpoint as input to initialize and train a BEV-planner~\cite{li2024bevplanner}, which is the best non-LLM planning baseline method in prior work OmniDrive~\cite{wang2025omnidrive}. Our \namemethod~still outperforms the non-LLM cooperative perception and planning baseline, as shown in Table~\ref{tab:ablation_non_llm}.

\begin{table}[!t]
\caption{
Perception and planning performance comparison with non-LLM baseline method.
\vspace{-10pt}
}
\small
\setlength{\tabcolsep}{1pt}
\begin{center}
{
\begin{tabular}{l c cc}
  \hline
  \hline
  \multirow{2}{*}{Method} &
  \multicolumn{1}{c}{Q1} & \multicolumn{2}{c}{Q5}\\
  \cmidrule(lr){2-2}  \cmidrule(lr){3-4}
  &
  F1 $\uparrow$ &
  L2$_{avg}$ (m) $\downarrow$ & CR$_{avg}$ (\%) $\downarrow$ \\
  \hline
  \hline
   CoBEVT~\cite{xu2022cobevt} + BEV-planner~\cite{li2024bevplanner}        & 65.7 & 5.82 & 11.59 \\
  \hline
  \namemethod~(ours)     & \textbf{70.0} & \textbf{4.99} & \textbf{3.00} \\
  \hline
\end{tabular}
}
\label{tab:ablation_non_llm}
\end{center}
\vspace{-10pt}
\end{table} 

\subsection{Robustness Assessment}
We follow V2X-ViT~\cite{xu2022v2xvit} to experiment on the impact of latency and sensor noise on positional errors. Our model is robust to these factors, as shown in Tables \ref{tab:ablation_latency} and \ref{tab:ablation_positional_error}.

\begin{table}[!t]
\caption{
Experiments in \namevsplit~with communication latency.
\vspace{-10pt}
}
\small
\setlength{\tabcolsep}{1pt}
\begin{center}
{
\begin{tabular}{l c c c c c cc}
  \hline
  \hline
  \multirow{2}{*}{Latency (s)} &
  \multicolumn{1}{c}{Q1} & \multicolumn{1}{c}{Q2} & \multicolumn{1}{c}{Q3} & \multicolumn{1}{c}{Q\textsubscript{Gr}} & \multicolumn{1}{c}{Q4} & \multicolumn{2}{c}{Q5}\\
  \cmidrule(lr){2-2} \cmidrule(lr){3-3} \cmidrule(lr){4-4} \cmidrule(lr){5-5} \cmidrule(lr){6-6} \cmidrule(lr){7-8}
  &
  F1 $\uparrow$ &
  F1 $\uparrow$ &
  F1 $\uparrow$ &
  F1 $\uparrow$ &
  F1 $\uparrow$ & 
  L2$_{avg}$ (m) $\downarrow$ & CR$_{avg}$ (\%) $\downarrow$ \\
  \hline
  \hline
  1.0 & 69.3 & 29.7 &  18.9 & 39.3 & 55.0 & 5.26 & 4.09 \\
  0.4 & 69.7 & 30.3 &  20.1 & 40.0 & 56.0 & 5.09 & 3.49 \\
  0.3 & 69.8 & 30.7 &  20.6 & 40.4 & 57.2 & 5.07 & 3.31 \\
  0.2 & 69.8 & \textbf{30.8} &  20.8 & 40.5 & 57.7 & 5.05 & 3.21 \\
  0.1 & 69.8 & 30.7 &  21.1 & 40.5 & 59.4 & 5.02 & 3.05 \\
  \hline
  0 & \textbf{70.0} & \textbf{30.8} &  \textbf{21.2} & \textbf{40.7} & \textbf{59.7} &  \textbf{4.99} & \textbf{3.00} \\
  \hline
\end{tabular}
}
\label{tab:ablation_latency}
\end{center}
\vspace{-10pt}
\end{table}

\begin{table}[!t]
\caption{
Experiments in \namevsplit~with positional errors. STD: standard deviation.
\vspace{-10pt}
}
\small
\setlength{\tabcolsep}{1pt}
\begin{center}
{
\begin{tabular}{l c c c c c cc}
  \hline
  \hline
  \multirow{2}{*}{STD (m)} &
  \multicolumn{1}{c}{Q1} & \multicolumn{1}{c}{Q2} & \multicolumn{1}{c}{Q3} & \multicolumn{1}{c}{Q\textsubscript{Gr}} & \multicolumn{1}{c}{Q4} & \multicolumn{2}{c}{Q5}\\
  \cmidrule(lr){2-2} \cmidrule(lr){3-3} \cmidrule(lr){4-4} \cmidrule(lr){5-5} \cmidrule(lr){6-6} \cmidrule(lr){7-8}
  &
  F1 $\uparrow$ &
  F1 $\uparrow$ &
  F1 $\uparrow$ &
  F1 $\uparrow$ &
  F1 $\uparrow$ & 
  L2$_{avg}$ (m) $\downarrow$ & CR$_{avg}$ (\%) $\downarrow$ \\
  \hline
  \hline
  1.0 & 69.8 & 29.9 & \textbf{21.7} & 40.5 & 57.2 & 5.21 & 3.86 \\ 
  0.4 & 69.8 & \textbf{30.9} & 21.5 & \textbf{40.7} & 59.2 & 5.03 & 3.27 \\
  0.3 & 69.8 & 30.7 &  21.0 & 40.5 & 59.1 & 5.00 & 3.20 \\
  0.2 & 69.8 & \textbf{30.9} &  21.0 & 40.6 & \textbf{60.0} & 4.99 & 3.10 \\
  0.1 & 69.8 & 30.8 &  21.3 & 40.6 & 59.8 & \textbf{4.98} & 3.05 \\
  \hline
  0 & \textbf{70.0} & 30.8 &  21.2 & \textbf{40.7} & 59.7 &  4.99 & \textbf{3.00} \\
  \hline
\end{tabular}
}
\label{tab:ablation_positional_error}
\end{center}
\vspace{-10pt}
\end{table}
  
\subsection{Ablation Study}

\noindent \textbf{Input Features}: We experiment with variants of our \namemethod~model that use either only the scene-level feature maps or only the object-level feature vectors as the visual input. The ablation results can be seen in Table \ref{tab:small_ablation_v2v}. Both types of features contribute to final performance in all QA tasks. In general, the object-level-only model outperforms the scene-level-only model. This implies that the object-level features are easier for LLM to digest, which is consistent with the results observed in the previous work with the TOKEN model~\cite{tian2024token}.

\begin{table}[!t]
\caption{
Ablation study in \namevsplit.
\vspace{-10pt}
}
\small
\setlength{\tabcolsep}{2pt}
\begin{center}
\begin{tabular}{l c c c c c cc }
  \hline
  \hline
  \multirow{2}{*}{Method} &
  \multicolumn{1}{c}{Q1} & \multicolumn{1}{c}{Q2} & \multicolumn{1}{c}{Q3} & \multicolumn{1}{c}{Q\textsubscript{Gr}} & \multicolumn{1}{c}{Q4} & \multicolumn{2}{c}{Q5} \\
  \cmidrule(lr){2-2} \cmidrule(lr){3-3} \cmidrule(lr){4-4} \cmidrule(lr){5-5} \cmidrule(lr){6-6} \cmidrule(lr){7-8}
  &
  F1 $\uparrow$ &
  F1 $\uparrow$ &
  F1 $\uparrow$ &
  F1 $\uparrow$ &
  F1 $\uparrow$ & 
  L2 (m) $\downarrow$ & CR (\%) $\downarrow$ \\
  \hline
  \hline
  Scene only         & 69.9 & 15.4 & 17.9 & 34.4 & 43.2 & 7.21 & 15.55 \\
  Object only      & 69.0 & 26.9 & 17.6 & 37.8 & 52.6 & 5.24 & 7.78 \\
  \hline
  Scratch         & 67.6 & 26.5 & 17.2 &  37.1 & 49.3 & 6.30 & 5.01 \\
  \hline
  \namemethod     & \textbf{70.0} & \textbf{30.8} & \textbf{21.2} & \textbf{40.7} & \textbf{59.7} & \textbf{4.99} & \textbf{3.00} \\
  \hline
\end{tabular}
\vspace{-5pt}
\label{tab:small_ablation_v2v}
\end{center}
\vspace{-15pt}
\end{table}

\noindent \textbf{Training from Scratch}: Table \ref{tab:small_ablation_v2v} also shows that training from scratch achieves worse performance, meaning that pre-training with LLaVA's VQA tasks improves our \namemethod's performance in \namedataset. 

\begin{figure*}[!t]
\centering
\includegraphics[width=1\textwidth]{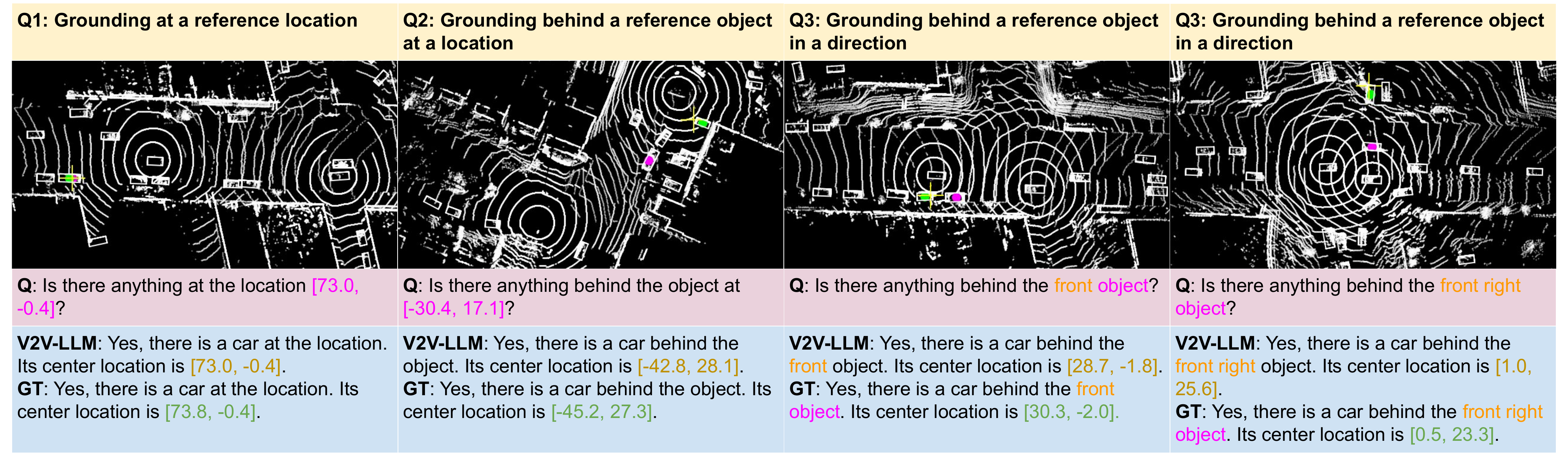}
\vspace{-10pt}
\caption[]
        {\namemethod's \textit{grounding} results on \namedataset's testing set.~\textcolor{magenta}{Magenta $\circ$}: reference locations in questions. \textcolor{olive}{Yellow $+$}: model output locations. \textcolor{Green}{Green $\circ$}: ground-truth answers.} 
        \label{fig:qualitative_result_grounding}
        \vspace{-5pt}
\end{figure*}

\begin{figure*}[!t]
\centering
\includegraphics[width=1\textwidth]{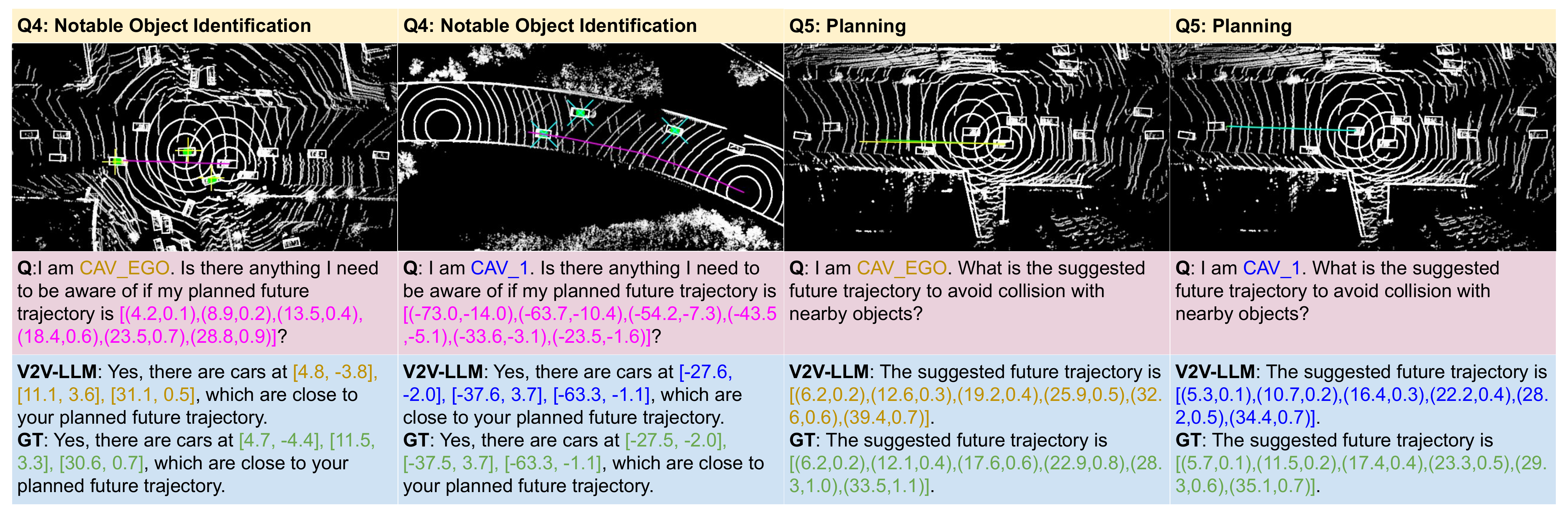}
\vspace{-10pt}
\caption[]
        {\namemethod's \textit{notable object identification} and \textit{planning} results on \namedataset's testing set. For notable object identification,~\textcolor{magenta}{Magenta curve}: planned future trajectories in questions. \textcolor{Green}{Green $\circ$}: ground-truth notable object locations. \textcolor{olive}{Yellow $+$} and \textcolor{cyan}{Cyan $\times$}: model identification outputs corresponding to \textcolor{olive}{CAV\_EGO} and \textcolor{cyan}{CAV\_1}, respectively.
        For planning, \textcolor{Green}{Green line}: future trajectories in ground-truth answers. \textcolor{olive}{Yellow curve} and \textcolor{cyan}{Cyan curve}: model planning outputs corresponding to \textcolor{olive}{CAV\_EGO} and \textcolor{cyan}{CAV\_1}, respectively.}         
        \label{fig:qualitative_result_notable_planning}
        \vspace{-10pt}
\end{figure*}

\subsection{Qualitative Results}
Figure \ref{fig:qualitative_result_grounding} shows our \namemethod's \textit{grounding} results and the ground truth with visualization on \namedataset's testing set. We can observe that our \namemethod~ is able to locate the objects given the provided reference information for each of the $3$ types of grounding questions: grounding at a reference location, grounding behind a reference object
at a location, and grounding behind a reference object in a direction. 
Figure \ref{fig:qualitative_result_notable_planning}'s left part shows our \namemethod's \textit{notable object identification} results. \namemethod~ demonstrate its capability of identifying multiple objects near the planned future trajectories specified in the questions for each CAV. 
Figure \ref{fig:qualitative_result_notable_planning}'s right part shows \namemethod's \textit{planning} results. Our model is able to suggest future trajectories that avoid potential collisions with nearby objects. Overall, the outputs of our model closely align with the ground-truth answers across all question types, indicating its robustness in cooperative autonomous driving tasks.

\section{Conclusion}
In this work, we expand the research scope of cooperative autonomous driving by integrating the use of Multimodal LLM-based methods, aimed at improving the safety of future autonomous driving systems. We propose a new problem setting and create a novel \namedataset~dataset and benchmark that includes grounding, notable object identification, and planning question-answering tasks designed for varieties of cooperative driving scenarios. We propose a baseline model \namemethod~that fuses each CAV's individual perception information and performs visual and language understanding to answer driving-related questions from any CAV. 
Our proposed \namemethod~outperforms  other baselines adopted from state-of-the-art cooperative perception algorithms in the grounding, notable object identification, and planning. Our method also outperforms non-LLM baseline and is robust to communication latency and noise. 
These experimental results indicate that \namemethod~is promising as a unified multimodal foundation model that can effectively perform perception and planning tasks for cooperative autonomous driving.
We have publicly released our \namedataset~ dataset and \namemethod~code to facilitate open-source research, and believe it will bring cooperative driving research to the next stage.


{\small
\bibliographystyle{IEEEtran}
\bibliography{egbib}

@String(CVPR= {IEEE Conf. Comput. Vis. Pattern Recog.})

@String(ECCV= {Eur. Conf. Comput. Vis.})

@String(ICLR = {Int. Conf. Learn. Represent.})

@String(AAAI = {AAAI})

@String(CVPR  = {CVPR})

@String(ECCV  = {ECCV})

@String(ICLR  = {ICLR})

@inproceedings{caesar2019nuscenes,
  title={nuscenes: A multimodal dataset for autonomous driving},
  author={Caesar, Holger and Bankiti, Varun and Lang, Alex H and Vora, Sourabh and Liong, Venice Erin and Xu, Qiang and Krishnan, Anush and Pan, Yu and Baldan, Giancarlo and Beijbom, Oscar},
  booktitle={IEEE/CVF Conference on Computer Vision and Pattern Recognition (CVPR)},
  year={2020}
}

@inproceedings{lang2019pointpillar,
  title = {Pointpillars: Fast encoders for object detection from point clouds},
  author = {Lang, Alex H. and Vora, Sourabh and Caesar, Holger and Zhou, Lubing and Yang, Jiong and Beijbom, Oscar},
  booktitle = {IEEE/CVF Conference on Computer Vision and Pattern Recognition (CVPR)},
  year = {2019}
}

@inproceedings{geiger2012kitti,
  author = {Geiger, Andreas and Lenz, Philip and Urtasun, Raquel},
  title = {Are we ready for Autonomous Driving? The KITTI Vision Benchmark Suite},
  booktitle = {IEEE/CVF Conference on Computer Vision and Pattern Recognition (CVPR)},
  year = {2012}
}

@inproceedings{sun2020waymo,
  title={Scalability in perception for autonomous driving: Waymo open dataset},
  author={Sun, Pei and Kretzschmar, Henrik and Dotiwalla, Xerxes and Chouard, Aurelien and Patnaik, Vijaysai and Tsui, Paul and Guo, James and Zhou, Yin and Chai, Yuning and Caine, Benjamin and others},
  booktitle={IEEE/CVF Conference on Computer Vision and Pattern Recognition (CVPR)},
  year={2020}
}

@inproceedings{cui2022coopernaut,
  title = {Coopernaut: End-to-End Driving with Cooperative Perception for Networked Vehicles},
  author={Cui, Jiaxun and Qiu, Hang and Chen, Dian and Stone, Peter and Zhu, Yuke},
  booktitle = {IEEE/CVF Conference on Computer Vision and Pattern Recognition (CVPR)},
  year = {2022}
}

@inproceedings{xu2022opencood,
 author = {Xu, Runsheng and Xiang, Hao and Xia, Xin and Han, Xu and Li, Jinlong  and Ma, Jiaqi},
 title = {OPV2V: An Open Benchmark Dataset and Fusion Pipeline for Perception with Vehicle-to-Vehicle Communication},
 booktitle = {IEEE International Conference on Robotics and Automation (ICRA)},
 year = {2022}
}

@inproceedings{xu2023v2v4real,
  title = {V2V4Real: A Real-world Large-scale Dataset for Vehicle-to-Vehicle Cooperative Perception},
  author = {Xu, Runsheng and Xia, Xin and Li, Jinlong and Li, Hanzhao and Zhang, Shuo and Tu, Zhengzhong and Meng, Zonglin and Xiang, Hao and Dong, Xiaoyu and Song, Rui and Yu, Hongkai and Zhou, Bolei and Ma, Jiaqi},
  booktitle = {IEEE/CVF Conference on Computer Vision and Pattern Recognition (CVPR)},
  year = {2023}
}

@inproceedings{dosovitskiy2017carla,
  title = {{CARLA}: {An} Open Urban Driving Simulator},
  author = {Dosovitskiy, Alexey and Ros, German and Codevilla, Felipe and Lopez, Antonio and Koltun, Vladlen},
  booktitle = {Conference on Robot Learning (CoRL)},
  year = {2017}
}

@inproceedings{chen2019fcooper,
author = {Chen, Qi and Ma, Xu and Tang, Sihai and Guo, Jingda and Yang, Qing and Fu, Song},
title = {F-Cooper: Feature Based Cooperative Perception for Autonomous Vehicle Edge Computing System Using 3D Point Clouds},
year = {2019},
booktitle = {ACM/IEEE Symposium on Edge Computing (SEC)},
}

@inproceedings{xu2022v2xvit,
 author = {Xu, Runsheng and Xiang, Hao and Tu, Zhengzhong and Xia, Xin and Yang, Ming-Hsuan and Ma, Jiaqi},
 title = {V2X-ViT: Vehicle-to-Everything Cooperative Perception with Vision Transformer},
 booktitle={European Conference on Computer Vision (ECCV)},
 year = {2022}
}

@inproceedings{xu2022cobevt,
 author = {Xu, Runsheng and Tu, Zhengzhong and Xiang, Hao and Shao, Wei and Zhou, Bolei and Ma, Jiaqi},
 title = {CoBEVT: Cooperative Bird's Eye View Semantic Segmentation with Sparse Transformers},
 booktitle={Conference on Robot Learning (CoRL)},
 year = {2022}}

@article{li2022v2xsim,
  title={V2X-Sim: Multi-agent collaborative perception dataset and benchmark for autonomous driving},
  author={Li, Yiming and Ma, Dekun and An, Ziyan and Wang, Zixun and Zhong, Yiqi and Chen, Siheng and Feng, Chen},
  journal={IEEE Robotics and Automation Letters (RA-L)},
  year={2022}
}

@inproceedings{xiang2024v2xreal,
  title = {V2X-Real: a Largs-Scale Dataset for Vehicle-to-Everything Cooperative Perception},
  author = {Xiang, Hao and Zheng, Zhaoliang and Xia, Xin and Xu, Runsheng and Gao, Letian and Zhou, Zewei and Han, Xu and Ji, Xinkai and Li, Mingxi and Meng, Zonglin and Jin, Li and Lei, Mingyue and Ma, Zhaoyang and He, Zihang and Ma, Haoxuan and Yuan, Yunshuang and Zhao, Yingqian and Ma, Jiaqi},
  booktitle = {Europian Conference on Computer Vision (ECCV)},
  year = {2024}
}

@inproceedings{chiu2024probabilistic,
  title={Probabilistic 3d multi-object cooperative tracking for autonomous driving via differentiable multi-sensor kalman filter},
  author={Chiu, Hsu-kuang and Wang, Chien-Yi and Chen, Min-Hung and Smith, Stephen F.},
  booktitle={IEEE International Conference on Robotics and Automation (ICRA)},
  year={2024}
}

@inproceedings{yu2022dair-v2x,
  title={Dair-v2x: A large-scale dataset for vehicle-infrastructure cooperative 3d object detection},
  author={Yu, Haibao and Luo, Yizhen and Shu, Mao and Huo, Yiyi and Yang, Zebang and Shi, Yifeng and Guo, Zhenglong and Li, Hanyu and Hu, Xing and Yuan, Jirui and Nie, Zaiqing},
  booktitle={IEEE/CVF Conference on Computer Vision and Pattern Recognition (CVPR)},
  year={2022}
}

@inproceedings{zimmer2024tumtraf,
  title={Tumtraf v2x cooperative perception dataset},
  author={Zimmer, Walter and Wardana, Gerhard Arya and Sritharan, Suren and Zhou, Xingcheng and Song, Rui and Knoll, Alois C},
  booktitle={IEEE/CVF Conference on Computer Vision and Pattern Recognition (CVPR)},
  year={2024}
}

@inproceedings{sima2023drivelm,
  title={DriveLM: Driving with Graph Visual Question Answering},
  author={Sima, Chonghao and Renz, Katrin and Chitta, Kashyap and Chen, Li and Zhang, Hanxue and Xie, Chengen and Luo, Ping and Geiger, Andreas and Li, Hongyang},
  booktitle = {Europian Conference on Computer Vision (ECCV)},
  year={2024}
}

@article{tian2024drivevlm,
    title={DriveVLM: The Convergence of Autonomous Driving and Large Vision-Language Models},
    author={Tian, Xiaoyu and Gu, Junru and Li, Bailin and Liu, Yicheng and Wang, Yang and Zhao, Zhiyong and Zhan, Kun and Jia, Peng and Lang, Xianpeng and Zhao, Hang},
    journal={arXiv preprint arXiv:2402.12289},
    year={2024}
}

@inproceedings{wang2025omnidrive,
  title={{OmniDrive}: A Holistic Vision-Language Dataset for Autonomous Driving with Counterfactual Reasoning},
  author={Shihao Wang and Zhiding Yu and Xiaohui Jiang and Shiyi Lan and Min Shi and Nadine Chang and Jan Kautz and Ying Li and Jose M. Alvarez},
  booktitle={CVPR},
  year={2025}
}

@inproceedings{tian2024token,
  title={Tokenize the World into Object-level Knowledge to Address Long-tail Events in Autonomous Driving},
  author={Tian, Ran and Li, Boyi and Weng, Xinshuo and Chen, Yuxiao and Schmerling, Edward and Wang, Yue and Ivanovic, Boris and Pavone, Marco},
  booktitle={Conference on Robot Learning (CoRL)},
  year={2024}
  }

@article{wu2023nuprompt,
  title={Language Prompt for Autonomous Driving},
  author={Wu, Dongming and Han, Wencheng and Wang, Tiancai and Liu, Yingfei and Zhang, Xiangyu and Shen, Jianbing},
  journal={arXiv preprint},
  year={2023}
}

@inproceedings{qian2023nuscenesqa,
  title={Nuscenes-qa: A multi-modal visual question answering benchmark for autonomous driving scenario},
  author={Qian, Tianwen and Chen, Jingjing and Zhuo, Linhai and Jiao, Yang and Jiang, Yu-Gang},
  booktitle={AAAI Conference on Artificial Intelligence (AAAI)},
  year={2024}
}

@inproceedings{marcu2023lingoqa,
  title={LingoQA: Visual question answering for autonomous driving},
  author={Marcu, Ana-Maria and Chen, Long and H{\"u}nermann, Jan and Karnsund, Alice and Hanotte, Benoit and Chidananda, Prajwal and Nair, Saurabh and Badrinarayanan, Vijay and Kendall, Alex and Shotton, Jamie and others},
  booktitle={European Conference on Computer Vision (ECCV)},
  year={2024}
}

@inproceedings{mao2023gpt,
  title={GPT-Driver: Learning to Drive with GPT},
  author={Mao, Jiageng and Qian, Yuxi and Ye, Junjie and Zhao, Hang and Wang, Yue},
  booktitle={Advances in Neural Information Processing Systems (NeurIPS) Workshop (Foundation Models for Decision Making)},
  year={2023}
}

@inproceedings{mao2023agentdriver,
  title={A Language Agent for Autonomous Driving},
  author={Mao, Jiageng and Ye, Junjie and Qian, Yuxi and Pavone, Marco and Wang, Yue},
  booktitle={Conference On Language Modeling (COLM)},
  year={2024}
}

@inproceedings{li2024drivingeverywhere,
  title={Driving everywhere with large language model policy adaptation},
  author={Li, Boyi and Wang, Yue and Mao, Jiageng and Ivanovic, Boris and Veer, Sushant and Leung, Karen and Pavone, Marco},
  booktitle={IEEE/CVF Conference on Computer Vision and Pattern Recognition (CVPR)},
  year={2024}
}

@inproceedings{wang2023driveanywhere,
  title={Drive Anywhere: Generalizable End-to-end Autonomous Driving with Multi-modal Foundation Models}, 
  author={Wang, Tsun-Hsuan and Maalouf, Alaa and Xiao, Wei and Ban, Yutong and Amini, Alexander and Rosman, Guy and Karaman, Sertac and Rus, Daniela},
  booktitle={IEEE International Conference on Robotics and Automation (ICRA)},
  year={2023}
}

@inproceedings{liu2023llava,
    author      = {Liu, Haotian and Li, Chunyuan and Wu, Qingyang and Lee, Yong Jae},
    title       = {Visual Instruction Tuning},
    booktitle   = {Advances in Neural Information Processing Systems (NeurIPS)},
    year        = {2023}
  }

@inproceedings{radford2021clip,
  author       = {Radford, Alec and Kim, Jong Wook and Hallacy, Chris and Ramesh, Aditya and Goh, Gabriel and Agarwal, Sandhini and Sastry, Girish and Askell, Amanda and Mishkin, Pamela and Clark, Jack and Krueger, Gretchen and Sutskever, Ilya},
  title        = {Learning Transferable Visual Models From Natural Language Supervision},
  booktitle={International Conference on Machine Learning (ICML)},
  year         = {2021}
}

@inproceedings{hu2022lora,
  title={Lo{RA}: Low-Rank Adaptation of Large Language Models},
  author={Edward J Hu and Yelong Shen and Phillip Wallis and Zeyuan Allen-Zhu and Yuanzhi Li and Shean Wang and Lu Wang and Weizhu Chen},
  booktitle={International Conference on Learning Representations (ICLR)},
  year={2022}
}

@misc{chiang2023vicuna,
    title = {Vicuna: An Open-Source Chatbot Impressing GPT-4 with 90\%* ChatGPT Quality},
    url = {https://lmsys.org/blog/2023-03-30-vicuna/},
    author = {Chiang, Wei-Lin and Li, Zhuohan and Lin, Zi and Sheng, Ying and Wu, Zhanghao and Zhang, Hao and Zheng, Lianmin and Zhuang, Siyuan and Zhuang, Yonghao and Gonzalez, Joseph E. and Stoica, Ion and Xing, Eric P.},
    year = {2023}
}

@inproceedings{chen2024drivingwithllms,
  title={Driving with LLMs: Fusing Object-Level Vector Modality for Explainable Autonomous Driving},
  author={Chen, Long and Sinavski, Oleg and H{\"u}nermann, Jan and Karnsund, Alice and Willmott, Andrew James and Birch, Danny and Maund, Daniel and Shotton, Jamie},
  booktitle={IEEE International Conference on Robotics and Automation (ICRA)},
  year={2024}
}

@article{wang2023drivemlm,
  title={DriveMLM: Aligning Multi-Modal Large Language Models with Behavioral Planning States for Autonomous Driving},
  author={Wang, Wenhai and Xie, Jiangwei and Hu, ChuanYang and Zou, Haoming and Fan, Jianan and Tong, Wenwen and Wen, Yang and Wu, Silei and Deng, Hanming and Li, Zhiqi and others},
  journal={arXiv preprint arXiv:2312.09245},
  year={2023}
}

@inproceedings{ding2024nuinstruct,
  title={Holistic Autonomous Driving Understanding by Bird's-Eye-View Injected Multi-Modal Large Models},
  author={Xinpeng, Ding and Jinahua, Han and Hang, Xu and Xiaodan, Laing and Xu, Hang and Wei, Zhang and Xiaomeng, Li},
  booktitle = {IEEE/CVF Conference on Computer Vision and Pattern Recognition (CVPR)},
  year={2024}
}

@inproceedings{chiu2023selective,
      title={Selective Communication for Cooperative Perception in End-to-End Autonomous Driving}, 
      author={Chiu, Hsu-kuang and Smith, Stephen F.},
      booktitle={IEEE International Conference on Robotics and Automation (ICRA) Workshop},
      year={2023} 
}

@article{huang2024v2xcooperativeperceptionautonomous,
  title={V2X cooperative perception for autonomous driving: Recent advances and challenges},
  author={Huang, Tao and Liu, Jianan and Zhou, Xi and Nguyen, Dinh C and Azghadi, Mostafa Rahimi and Xia, Yuxuan and Han, Qing-Long and Sun, Sumei},
  journal={arXiv preprint arXiv:2310.03525},
  year={2023}
}

@inproceedings{yazgan2024collaborativeperceptiondatasetsautonomous,
  title={Collaborative perception datasets in autonomous driving: A survey},
  author={Yazgan, Melih and Akkanapragada, Mythra Varun and Z{\"o}llner, J Marius},
  booktitle={IEEE Intelligent Vehicles Symposium (IV)},
  year={2024}
}

@article{liu2024asurveyon,
  title={A survey on autonomous driving datasets: Statistics, annotation quality, and a future outlook},
  author={Liu, Mingyu and Yurtsever, Ekim and Fossaert, Jonathan and Zhou, Xingcheng and Zimmer, Walter and Cui, Yuning and Zagar, Bare Luka and Knoll, Alois C},
  journal={IEEE Transactions on Intelligent Vehicles (T-IV)},
  year={2024}
}

@inproceedings{li2024bevplanner,
  title={Is Ego Status All You Need for Open-Loop End-to-End Autonomous Driving?},
  author={Zhiqi Li and Zhiding Yu and Shiyi Lan and Jiahan Li and Jan Kautz and Tong Lu and Jose M. Alvarez},
  booktitle = {IEEE/CVF Conference on Computer Vision and Pattern Recognition (CVPR)},
  year={2024}
}
}

\clearpage
\section*{APPENDIX}


\section{Detailed Evaluation Results}
Tables \ref{tab:all_v2v_result_supp} and \ref{tab:all_v2x_result_supp} summarize the detailed evaluation results of our \namemethod~and other baseline methods in \namedataset's \namevsplit~and \namexsplit. In addition, Tables \ref{tab:v2v_planning} and \ref{tab:v2x_planning} show the detailed planning performance. 
For the grounding task, our \namemethod~achieves competitive results in \namevsplit~and outperforms all other baseline methods in \namexsplit. More importantly, for the notable object identification task and the planning task, our \namemethod~outperforms all other baseline methods in both \namevsplit~and \namexsplit.

\begin{table*}[t!]
\caption{
\namemethod's testing performance in \namedataset's \namevsplit~in comparison with baseline methods. Q1: Grounding at a reference location. Q2: Grounding behind a reference object at a location. Q3: Grounding behind a reference object in a direction. Q\textsubscript{Gr}: Average of grounding (Q1, Q2, and Q3). Q4: Notable object identification. Q5: Planning. P: Precision. R: Recall. L2: L2 distance error. CR: Collision rate. Comm: Communication cost. In each column, the \textbf{best} results are in boldface, and the \underline{second-best} results are in underline.
\vspace{-10pt}
}
\small
\setlength{\tabcolsep}{2pt}
\begin{center}
\begin{tabular}{l ccc ccc ccc c ccc cc c}
  \hline
  \hline
  \multirow{2}{*}{Method} &
  \multicolumn{3}{c}{Q1} & \multicolumn{3}{c}{Q2} & \multicolumn{3}{c}{Q3} & \multicolumn{1}{c}{Q\textsubscript{Gr}} & \multicolumn{3}{c}{Q4} & \multicolumn{2}{c}{Q5} & \multirow{2}{*}{Comm(MB) $\downarrow$} \\
  \cmidrule(lr){2-4} \cmidrule(lr){5-7} \cmidrule(lr){8-10} \cmidrule(lr){11-11} \cmidrule(lr){12-14} \cmidrule(lr){15-16}
  &
  F1 $\uparrow$ & P $\uparrow$ & R $\uparrow$ &
  F1 $\uparrow$ & P $\uparrow$ & R $\uparrow$ &
  F1 $\uparrow$ & P $\uparrow$ & R $\uparrow$ &
  F1 $\uparrow$ & 
  F1 $\uparrow$ & P $\uparrow$ & R $\uparrow$ & 
  L2$_{avg}$ (m) $\downarrow$ & CR$_{avg}$ (\%) $\downarrow$ \\
  \hline
  \hline
  \textit{No Fusion}         & 66.6 & 77.9 & 58.2 & 22.6 & 29.4 & 18.4 & 17.2 & 17.4 & 16.9 & 35.5 & 47.3 & 49.2 & 45.6 & 6.55 & 4.57 & \textbf{0} \\
  \textit{Early Fusion}      & \textbf{73.5} & \textbf{82.2} & \underline{66.5} & 23.3 & 29.1 & 19.5 & 20.8 & \underline{22.7} & 19.3 & 39.2 & 53.9 & 55.4 & 52.6 & \underline{6.20} & \underline{3.55} & 1.9208 \\
  \hline
  \scriptsize{\textit{Intermediate Fusion}} \\ 
  AttFuse~\cite{xu2022opencood}         & 70.7 & 79.6 & 63.6 & 26.4 & 31.6 & 22.7 & 18.4 & 19.6 & 17.4 & 38.5 & 56.9 & \underline{57.2} & 56.6 & 6.83 & 4.12 & \underline{0.4008} \\
  V2X-ViT~\cite{xu2022v2xvit}           & 70.8 & \underline{81.1} & 62.8 & 28.0 & 33.9 & 23.9 & \textbf{22.6} & \textbf{25.2} & 20.5 & 40.5 & \underline{57.6} & 57.0 & \textbf{58.2} & 7.08 & 4.33 & \underline{0.4008} \\
  CoBEVT~\cite{xu2022cobevt}            & \underline{72.2} & 76.8 & \textbf{68.1} & \underline{29.3} & \underline{34.7} & \underline{25.3} & \underline{21.3} & 22.1 & \underline{20.6} & \textbf{40.9} & \underline{57.6} & \underline{57.2} & \underline{58.1} & 6.72 & 3.88 & \underline{0.4008} \\
  \hline
  \scriptsize{\textit{LLM Fusion}} \\
  \namemethod~(Ours)     & 70.0 & 80.1 & 62.2 & \textbf{30.8} & \textbf{36.3} & \textbf{26.7} & 21.2 & 21.5 & \textbf{20.8} & \underline{40.7} & \textbf{59.7} & \textbf{61.9} & 57.6  & \textbf{4.99} & \textbf{3.00} & 0.4068 \\
  \hline
\end{tabular}
\label{tab:all_v2v_result_supp}
\end{center}
\vspace{-10pt}
\end{table*}

\begin{table*}[t!]
\caption{
\namemethod's testing performance in \namedataset's \namexsplit~in comparison with baseline methods. 
Q1: Grounding at a reference location. Q2: Grounding behind a reference object at a location. Q3: Grounding behind a reference object in a direction. Q\textsubscript{Gr}: Average of grounding (Q1, Q2, and Q3). Q4: Notable object identification. Q5: Planning. P: Precision. R: Recall. L2: L2 distance error. CR: Collision rate. Comm: Communication cost. In each column, the \textbf{best} results are in boldface, and the \underline{second-best} results are in underline.
\vspace{-10pt}
}
\small
\setlength{\tabcolsep}{2pt}
\begin{center}
\begin{tabular}{l ccc ccc ccc c ccc cc c}
  \hline
  \hline
  \multirow{2}{*}{Method} &
  \multicolumn{3}{c}{Q1} & \multicolumn{3}{c}{Q2} & \multicolumn{3}{c}{Q3} & \multicolumn{1}{c}{Q\textsubscript{Gr}} & \multicolumn{3}{c}{Q4} & \multicolumn{2}{c}{Q5} & \multirow{2}{*}{Comm(MB) $\downarrow$} \\
  \cmidrule(lr){2-4} \cmidrule(lr){5-7} \cmidrule(lr){8-10} \cmidrule(lr){11-11} \cmidrule(lr){12-14} \cmidrule(lr){15-16}
  &
  F1 $\uparrow$ & P $\uparrow$ & R $\uparrow$ &
  F1 $\uparrow$ & P $\uparrow$ & R $\uparrow$ &
  F1 $\uparrow$ & P $\uparrow$ & R $\uparrow$ &
  F1 $\uparrow$ & 
  F1 $\uparrow$ & P $\uparrow$ & R $\uparrow$ & 
  L2$_{avg}$ (m) $\downarrow$ & CR$_{avg}$ (\%) $\downarrow$ \\
  \hline
  \hline
  \textit{No Fusion}         & 55.7 & \textbf{71.6} & 45.5 & 21.4 & 33.2 & 15.8 & 25.2 & 26.2 & 24.2 & 34.1 & 64.4 & 66.1 & 62.7 & 2.31 & 9.21 & \textbf{0} \\
  \textit{Early Fusion}      & \underline{59.7} & 70.6 & 51.8 & 23.3 & 34.0 & 17.7 & 26.1 & 28.0 & 24.5 & 36.4 & \underline{67.6} & \underline{69.3} & \underline{66.0} & \underline{2.12} & 8.61 & 1.9208 \\
  \hline
  \scriptsize{\textit{Intermediate Fusion}} \\ 
  AttFuse~\cite{xu2022opencood}         & 58.9 & \underline{71.1} & 50.3 & 23.9 & \underline{34.3} & 18.4 & \underline{26.3} & \textbf{28.3} & \underline{24.6} & 36.4 & 65.9 & 67.0 & 64.9 & 2.19 & \underline{8.39} & \underline{0.4008} \\
  V2X-ViT~\cite{xu2022v2xvit}           & 59.6 & 69.6 & \underline{52.1} & \underline{24.2} & 33.2 & \underline{19.1} & 26.1 & \underline{28.2} & 24.3 & \underline{36.6} & 65.0 & 64.8 & 65.3 & 2.29 & 8.86 & \underline{0.4008} \\
  \hline
  \scriptsize{\textit{LLM Fusion}} \\
  \namemethod~(Ours)     & \textbf{60.5} & 69.5 & \textbf{53.6} & \textbf{25.3} & \textbf{34.9} & \textbf{19.8} & \textbf{26.7} & 27.0 & \textbf{26.4} & \textbf{37.5} & \textbf{69.3} & \textbf{71.9} & \textbf{66.8} & \textbf{1.71} & \textbf{6.89} & 0.4068 \\
  \hline
\end{tabular}
\label{tab:all_v2x_result_supp}
\end{center}
\vspace{-5pt}
\end{table*}

\begin{table*}[t!]
\caption{
\namemethod's planning performance in \namedataset's \namevsplit~in comparison with baseline methods. L2: L2 distance error. CR: Collision rate. In each column, the \textbf{best} results are in boldface, and the \underline{second-best} results are in underline.
\vspace{-10pt}
}
\small
\begin{center}
\begin{tabular}{l cccc cccc }
  \hline
  \hline
  \multirow{2}{*}{Method} &
  \multicolumn{4}{c}{L2 (m)} & \multicolumn{4}{c}{CR (\%)} \\
  \cmidrule(lr){2-5} \cmidrule(lr){6-9}
  &
  1s $\downarrow$ & 2s $\downarrow$ & 3s $\downarrow$ & average $\downarrow$ &
  1s $\downarrow$ & 2s $\downarrow$ & 3s $\downarrow$ & average $\downarrow$ \\
  \hline
  \hline
  \textit{No Fusion}         & 3.84 & 6.52 & 9.30 & 6.55 & 1.31 & 4.76 & 7.63 & 4.57  \\
  \textit{Early Fusion}      & \underline{3.68} & \underline{6.19} & \underline{8.74} & \underline{6.20} & 0.96 & 3.86 & \underline{5.83} & \underline{3.55} \\
  \hline
  \scriptsize{\textit{Intermediate Fusion}} \\
  AttFuse~\cite{xu2022opencood}         & 4.06 & 6.78	& 9.64 & 6.83 & 1.42 & 4.41 & 6.53 & 4.12  \\
  V2X-ViT~\cite{xu2022v2xvit}           & 4.21 & 7.05 & 9.99 & 7.08 & 1.33 & 4.82 & 6.85 & 4.33  \\
  CoBEVT~\cite{xu2022cobevt}            & 3.97 & 6.71 & 9.47 & 6.72 & \underline{0.93} & \underline{3.74} & 6.96 & 3.88  \\
  \hline
  \scriptsize{\textit{LLM Fusion}} \\
  \namemethod~(ours)     & \textbf{2.96} & \textbf{4.97} & \textbf{7.05} & \textbf{4.99} & \textbf{0.55} & \textbf{3.19} & \textbf{5.25} &  \textbf{3.00} \\
  \hline
\end{tabular}
\label{tab:v2v_planning}
\end{center}
\end{table*}

\begin{table*}[t!]
\caption{
\namemethod's planning performance in \namedataset's \namexsplit~in comparison with baseline methods. L2: L2 distance error. CR: Collision rate. In each column, the \textbf{best} results are in boldface, and the \underline{second-best} results are in underline.
\vspace{-10pt}
}
\small
\begin{center}
\begin{tabular}{l cccc cccc }
  \hline
  \hline
  \multirow{2}{*}{Method} &
  \multicolumn{4}{c}{L2 (m)} & \multicolumn{4}{c}{CR (\%)} \\
  \cmidrule(lr){2-5} \cmidrule(lr){6-9}
  &
  1s $\downarrow$ & 2s $\downarrow$ & 3s $\downarrow$ & average $\downarrow$ &
  1s $\downarrow$ & 2s $\downarrow$ & 3s $\downarrow$ & average $\downarrow$ \\
  \hline
  \hline
  \textit{No Fusion}         & 1.33 & 2.28 & 3.31 & 2.31 & 2.52 & 9.54 & 15.57 & 9.21  \\
  \textit{Early Fusion}      & \underline{1.24} & \underline{2.10} & \underline{3.00} & \underline{2.12} & 3.51 & \underline{8.37} & 13.93 & 8.61  \\
  \hline
  \scriptsize{\textit{Intermediate Fusion}} \\
  AttFuse~\cite{xu2022opencood}         & 1.27 & 2.17 & 3.11 & 2.19 & 2.40 & 9.07 & \underline{13.70} & \underline{8.39} \\
  V2X-ViT~\cite{xu2022v2xvit}           & 1.34 & 2.27 & 3.25 & 2.29 & \textbf{1.41} & 9.89 & 15.28 & 8.86  \\
  \hline
  \scriptsize{\textit{LLM Fusion}} \\
  \namemethod~(ours)     & \textbf{0.99} & \textbf{1.70} & \textbf{2.45} & \textbf{1.71} & \underline{2.17} & \textbf{6.79} & \textbf{11.71} & \textbf{6.89}  \\
  \hline
\end{tabular}
\label{tab:v2x_planning}
\end{center}
\end{table*}

\section{Detailed Communication Cost and Scaling Analysis}
In our centralized setting, assume that there is one centralized LLM computing node, $N_v$ CAVs, and each CAV asks $N_q$ questions at each timestep. Each CAV sends one scene-level feature map ($\leq 0.2$MB), one set of individual object detection result parameters ($\leq 0.003$MB), $N_q$ questions (each $\leq 0.0002$MB) to the LLM and receives $N_q$ answers (each $\leq 0.0002$MB) at each timestep.
Note that each CAV only needs to send the same features to the LLM  once at each timestep because the LLM node can save and reuse them to answer multiple questions from the same or different CAVs at the same timestep. The communication cost of each CAV is: $0.2 + 0.003 + (0.0002 + 0.0002)N_q = (0.203 + 0.0004N_q)$ MB. The LLM receives $N_v$ scene-level feature maps, $N_v$ set of individual object detection result parameters, $N_qN_v$ questions and returns $N_qN_v$ answers. The communication cost of the centralized LLM is $(0.2 + 0.003 + (0.0002 + 0.0002)N_q) N_v = (0.203N_v + 0.0004N_qN_v)$ MB.


Alternatively, one can also consider a decentralized setting that deploys one LLM in each CAV. In this setting, each CAV receives the features from all other CAVs and does not need to send or receive any questions or answers. The communication cost of each CAV is $(0.2 + 0.003) (N_v - 1) = 0.203(N_v - 1)$ MB. Table \ref{tab:communication_cost_supp} summarizes the communication cost and scaling analysis in the aforementioned settings. There could be more different decentralized settings. Which setting works best in terms of communication costs is beyond the current focus of our work.

\begin{table}[!t]
\caption{
Communication cost (MB) and scaling analysis. $N_v$: number of CAVs. $N_q$: number of questions asked by each CAV at each timestep.
\vspace{-10pt}
}
\small
\setlength{\tabcolsep}{3pt}
\begin{center}
\begin{tabular}{l c c }
  \hline
  \hline
  Setting  & Each CAV & Centralized LLM \\
  \hline
  \hline
  Centralized  & $0.203 + 0.0004N_q$ & $0.203N_v + 0.0004N_qN_v$ \\
  Decentralized      & $0.203(N_v - 1)$ & -  \\
  \hline
\end{tabular}
\label{tab:communication_cost_supp}
\end{center}
\vspace{-10pt}
\end{table}

\section{Planning Results with Temporal Inputs}
In the main paper, all experiments use point clouds at a single frame from each CAV as the visual input to the models. In this section, we experiment with feeding visual features from $3$ consecutive frames, the current one and the previous two, as the visual input to the models. Table \ref{tab:planning_3} shows the planning results of the new setting together with the original setting from the main paper. In general, using visual inputs from multiple frames improves planning performance. 

\begin{table}[t!]
\caption{
\namemethod's planning performance in \namedataset's \namevsplit~in comparison with baseline methods. L2: L2 distance error. CR: Collision rate. In each column, the \textbf{best} results are in boldface. and the \underline{second-best} results are in underline.
\vspace{-10pt}
}
\small
\setlength{\tabcolsep}{4pt}
\begin{center}
\begin{tabular}{l cc cc }
  \hline
  \hline
  \multirow{2}{*}{Method} &
  \multicolumn{2}{c}{1 input frame} & \multicolumn{2}{c}{3 input frames} \\
  \cmidrule(lr){2-3} \cmidrule(lr){4-5}
  &
  L2 (m) $\downarrow$ & CR (\%) $\downarrow$ &
  L2 (m) $\downarrow$ & CR (\%) $\downarrow$ \\
  \hline
  \hline
  \textit{No Fusion}          & 6.55 &  4.57 & 5.94 & 3.77  \\
  \textit{Early Fusion}      & \underline{6.20} &  \underline{3.55} & \underline{5.13} & \underline{3.04} \\
  \hline
  \scriptsize{\textit{Intermediate Fusion}} \\
  AttFuse~\cite{xu2022opencood}         & 6.83 & 4.12 & 6.46 & 3.50 \\
  V2X-ViT~\cite{xu2022v2xvit}           & 7.08 & 4.33 & 5.52 & 3.84 \\
  CoBEVT~\cite{xu2022cobevt}            & 6.72 & 3.88 & 6.02 & 3.40 \\
  \hline
  \scriptsize{\textit{LLM Fusion}} \\
  \namemethod~(ours)     & \textbf{4.99} &  \textbf{3.00} & \textbf{4.82} & \textbf{2.93} \\
  \hline
\end{tabular}
\vspace{-5pt}
\label{tab:planning_3}
\end{center}
\end{table}

\section{Detailed Ablation Results}
Table \ref{tab:detailed_ablation_v2v} shows the detailed ablation results when using only the scene-level features or only the object-level features as input to our \namemethod. Both types of input features contribute to the final performance, and object-level features are easier for LLM to digest. Training from scratch achieves worse performance, meaning that pre-training with LLaVA's VQA tasks improves our \namemethod's performance in \namedataset.

\begin{table*}[!ht]
\caption{
Ablation study in \namedataset's \namevsplit. Q1: Grounding at a reference location. Q2: Grounding behind a reference object at a location. Q3: Grounding behind a reference object in a direction. Q\textsubscript{Gr}: Average of grounding (Q1, Q2, and Q3). Q4: Notable object identification. Q5: Planning. P: Precision. R: Recall. L2: L2 distance error. CR: Collision rate. Comm: Communication cost.
\vspace{-10pt}
}
\small
\setlength{\tabcolsep}{2pt}
\begin{center}
\begin{tabular}{l ccc ccc ccc c ccc cc c}
  \hline
  \hline
  \multirow{2}{*}{Method} &
  \multicolumn{3}{c}{Q1} & \multicolumn{3}{c}{Q2} & \multicolumn{3}{c}{Q3} & \multicolumn{1}{c}{Q\textsubscript{Gr}} & \multicolumn{3}{c}{Q4} & \multicolumn{2}{c}{Q5} & \multirow{2}{*}{Comm (MB) $\downarrow$} \\
  \cmidrule(lr){2-4} \cmidrule(lr){5-7} \cmidrule(lr){8-10} \cmidrule(lr){11-11} \cmidrule(lr){12-14} \cmidrule(lr){15-16}
  &
  F1 $\uparrow$ & P $\uparrow$ & R $\uparrow$ &
  F1 $\uparrow$ & P $\uparrow$ & R $\uparrow$ &
  F1 $\uparrow$ & P $\uparrow$ & R $\uparrow$ &
  F1 $\uparrow$ &
  F1 $\uparrow$ & P $\uparrow$ & R $\uparrow$ & 
  L2$_{avg}$ (m) $\downarrow$ & CR$_{avg}$ (\%) $\downarrow$ \\
  \hline
  \hline
  Scene-level only         & 69.9 & 74.9 & \textbf{65.5} & 15.4 & 19.9 & 12.6 & 17.9 & \textbf{26.9} & 13.5 & 34.4 & 43.2 & 40.2 & 46.7 & 7.21 & 15.55 & 0.4008 \\
  Object-level only      & 69.0 & \textbf{80.9} & 60.1 & 26.9 & 34.7 & 21.9 & 17.6 & 18.3 & 16.9 & 37.8 & 52.6 & 57.3 & 48.6 & 5.24 & 7.78 & \textbf{0.0068} \\
  \hline
  Scratch & 67.6 & 77.6 & 60.0 & 26.5 & 26.4 & 26.5 & 17.2 & 16.4 & 18.2 & 37.1 & 49.3 & 52.7 & 46.3 & 6.30 & 5.01 & 0.4068 \\
  \hline
  \namemethod~(ours)     & \textbf{70.0} & 80.1 & 62.2 & \textbf{30.8} & \textbf{36.3} & \textbf{26.7} & \textbf{21.2} & 21.5 & \textbf{20.8} & \textbf{40.7} & \textbf{59.7} & \textbf{61.9} & \textbf{57.6} & \textbf{4.99} & \textbf{3.00} & 0.4068 \\
  \hline
\end{tabular}
\vspace{-5pt}
\label{tab:detailed_ablation_v2v}
\end{center}
\vspace{-10pt}
\end{table*} 
 

\section{Additional Dataset Statistics}

For the grounding questions (Q1, Q2, Q3) and the notable object identification question (Q4), a QA pair can be categorized into either a positive case or a negative case. If at least one object exists that satisfies the condition specified in the questions, the corresponding QA pair is a positive case. Otherwise, it is a negative case. Tables \ref{tab:stats_v2v_pos_neg} and \ref{tab:stats_v2x_pos_neg} summarize the numbers of QA pairs in each category, for \namedataset's \namevsplit~and \namexsplit~respectively. This table shows that \namedataset~has sufficient positive and negative data samples in both training and testing sets for each of these QA pairs. The planning task (Q5) is excluded from Tables \ref{tab:stats_v2v_pos_neg} and \ref{tab:stats_v2x_pos_neg}, as each planning QA pair inherently includes a ground-truth future trajectory in its corresponding answer.

\begin{table}[!t]
\caption{
Dataset statistics of our \namedataset's \namevsplit~on positive and negative samples. 
\vspace{-10pt}
}
\small
\setlength{\tabcolsep}{3.5pt}
\begin{center}
\begin{tabular}{c | rr | rr | r}
  \hline
  \hline
  QA type & Train-Pos & Train-Neg & Test-Pos & Test-Neg & Total \\
  \hline
  \hline
  Q1 & 217403 & 137417 & 76522 & 44861 & 476203 \\
  Q2 &  17859 &  17841 &  8391 &  5491 &  49582 \\
  Q3 &   7197 &   7142 &  3082 &  2015 &  19436 \\
  Q4 &   9911 &   2379 &  2517 &   929 &  15736 \\
  \hline
  Total & 252370 & 164779 & 90512 & 53296 & 560957 
\end{tabular}
\label{tab:stats_v2v_pos_neg}
\end{center}
\vspace{-10pt}
\end{table}

\begin{table}[!t]
\caption{
Dataset statistics of our \namedataset's \namexsplit~on positive and negative samples. 
\vspace{-10pt}
}
\small
\setlength{\tabcolsep}{3.5pt}
\begin{center}
\begin{tabular}{c | rr | rr | r}
  \hline
  \hline
  QA type & Train-Pos & Train-Neg & Test-Pos & Test-Neg & Total \\
  \hline
  \hline
  Q1 & 247447 & 247843 & 62332 & 66379 & 624001 \\
  Q2 &  84005 &  83689 & 18297 & 16936 & 202927 \\
  Q3 &  14346 &  14394 &  3421 &  3044 &  35205 \\
  Q4 &   4624 &   1650 &  1172 &   536 &  7982 \\
  \hline
  Total & 350422 & 347576 & 85222 & 86895 & 870115 
\end{tabular}
\label{tab:stats_v2x_pos_neg}
\end{center}
\vspace{-10pt}
\end{table}

\begin{figure}[!t]
        \centering
        \begin{subfigure}[t]{0.23\textwidth}
            \centering 
            \includegraphics[width=\textwidth]{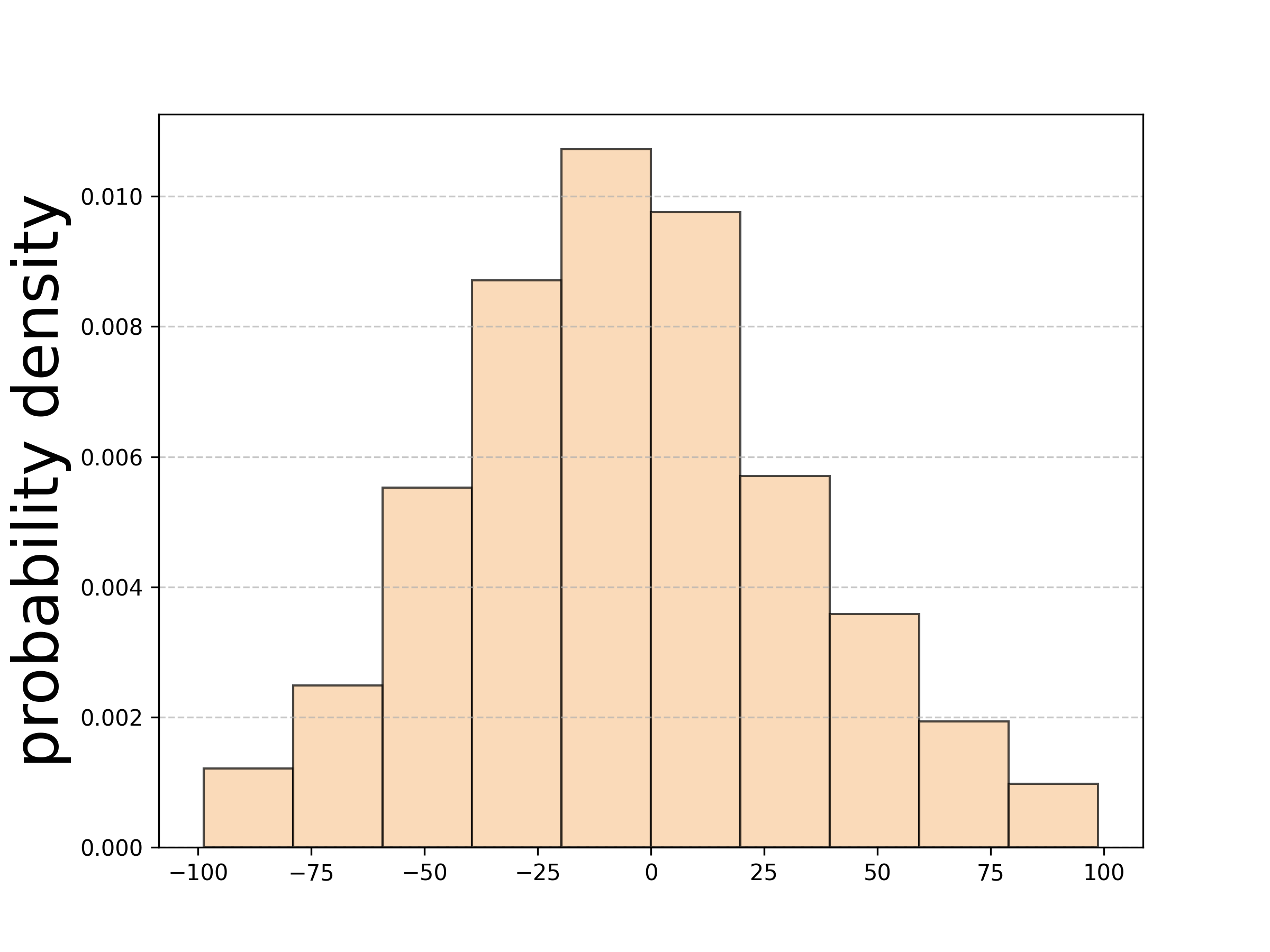}
            \vspace{-20pt}
            \caption[]%
            {{x (meters)}}    
        \end{subfigure}
        \hfill
        \begin{subfigure}[t]{0.23\textwidth}  
            \centering 
            \includegraphics[width=\textwidth]{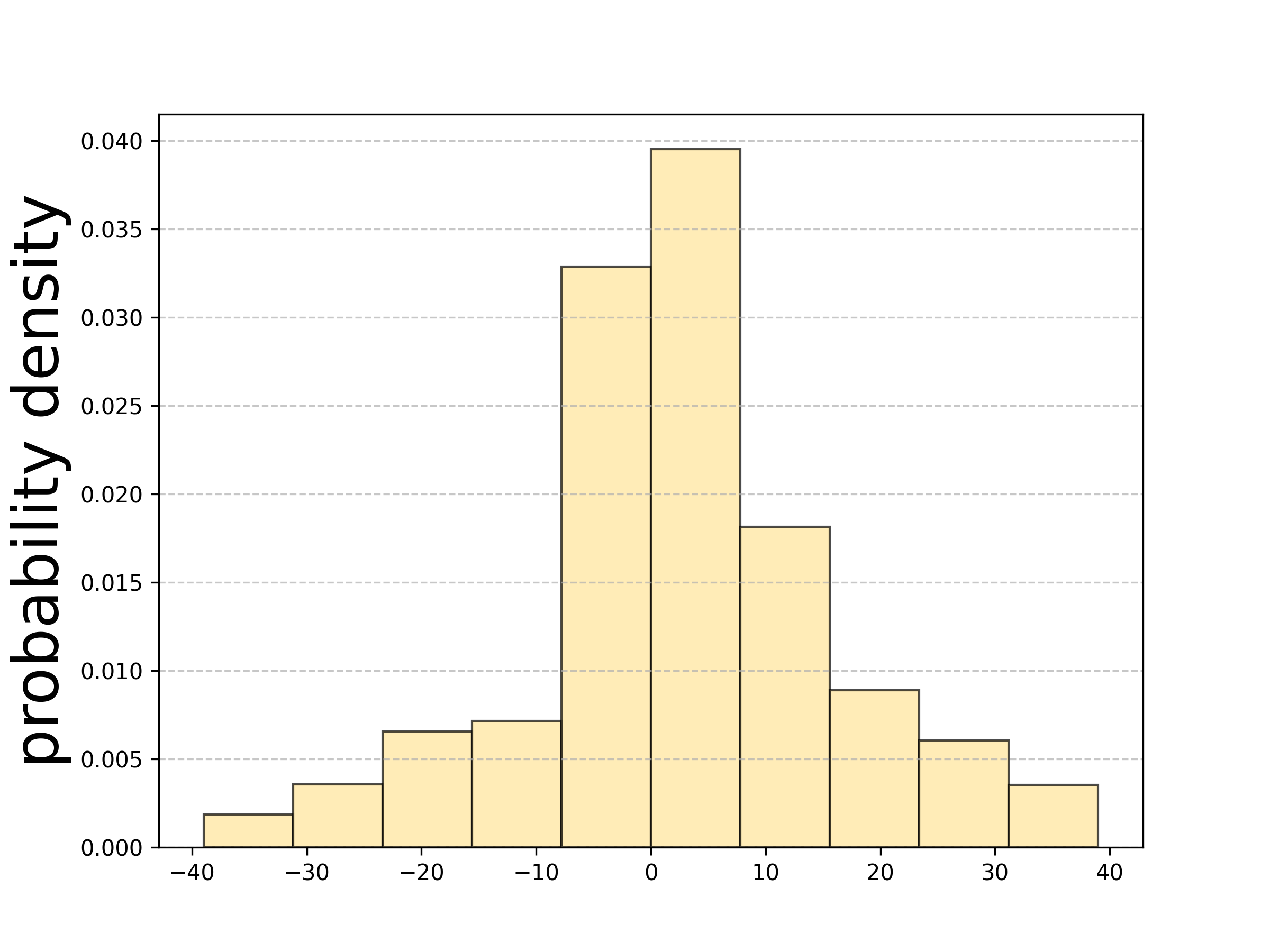}
            \vspace{-20pt}
            \caption[]%
            {{y (meters)}}
        \end{subfigure}

        \begin{subfigure}[t]{0.23\textwidth}
            \centering 
            \includegraphics[width=\textwidth]{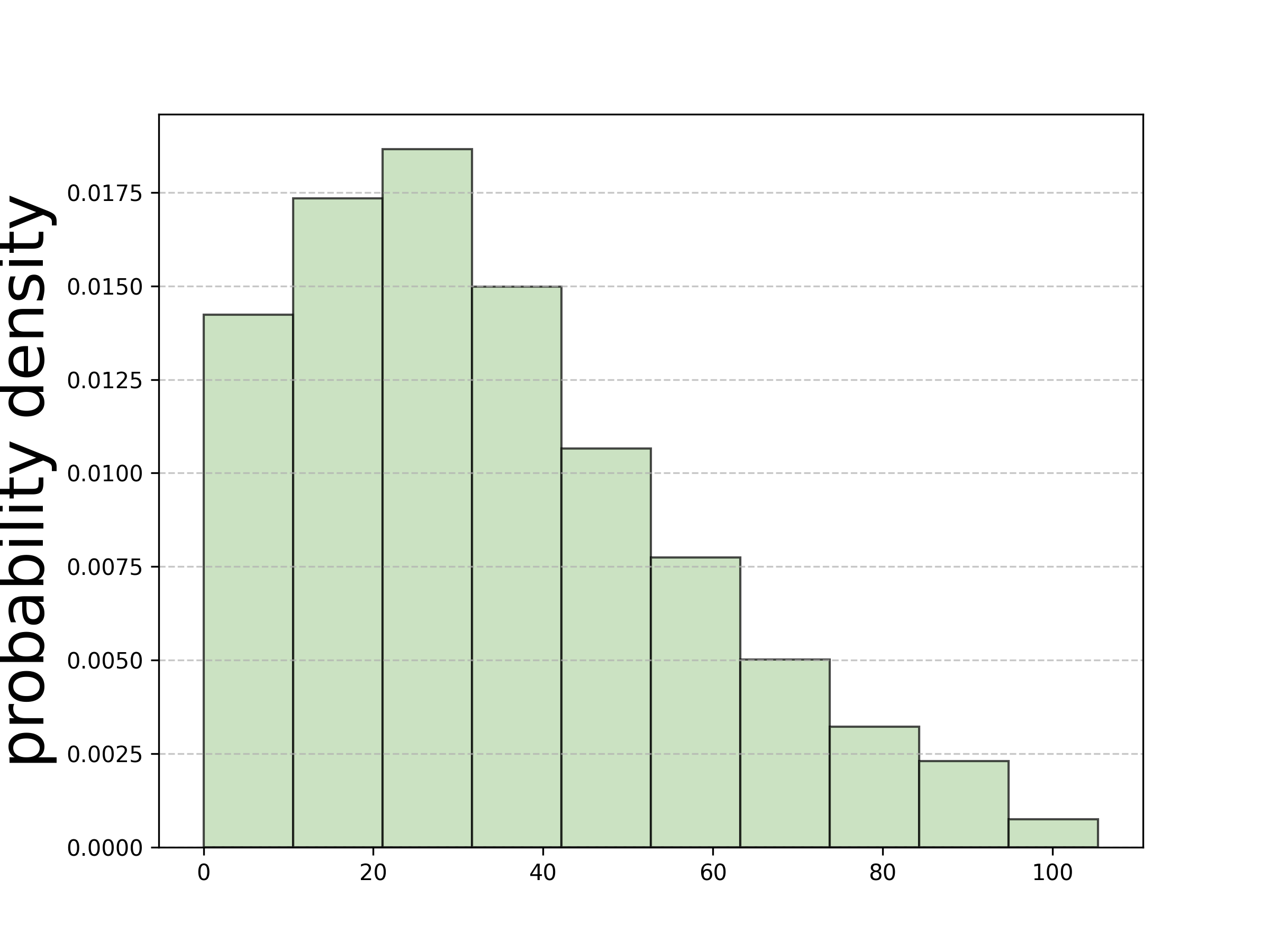}
            \vspace{-20pt}
            \caption[]%
            {{distance (meters)}}
        \end{subfigure}
        \hfill
        \begin{subfigure}[t]{0.23\textwidth}
            \centering 
            \includegraphics[width=\textwidth]{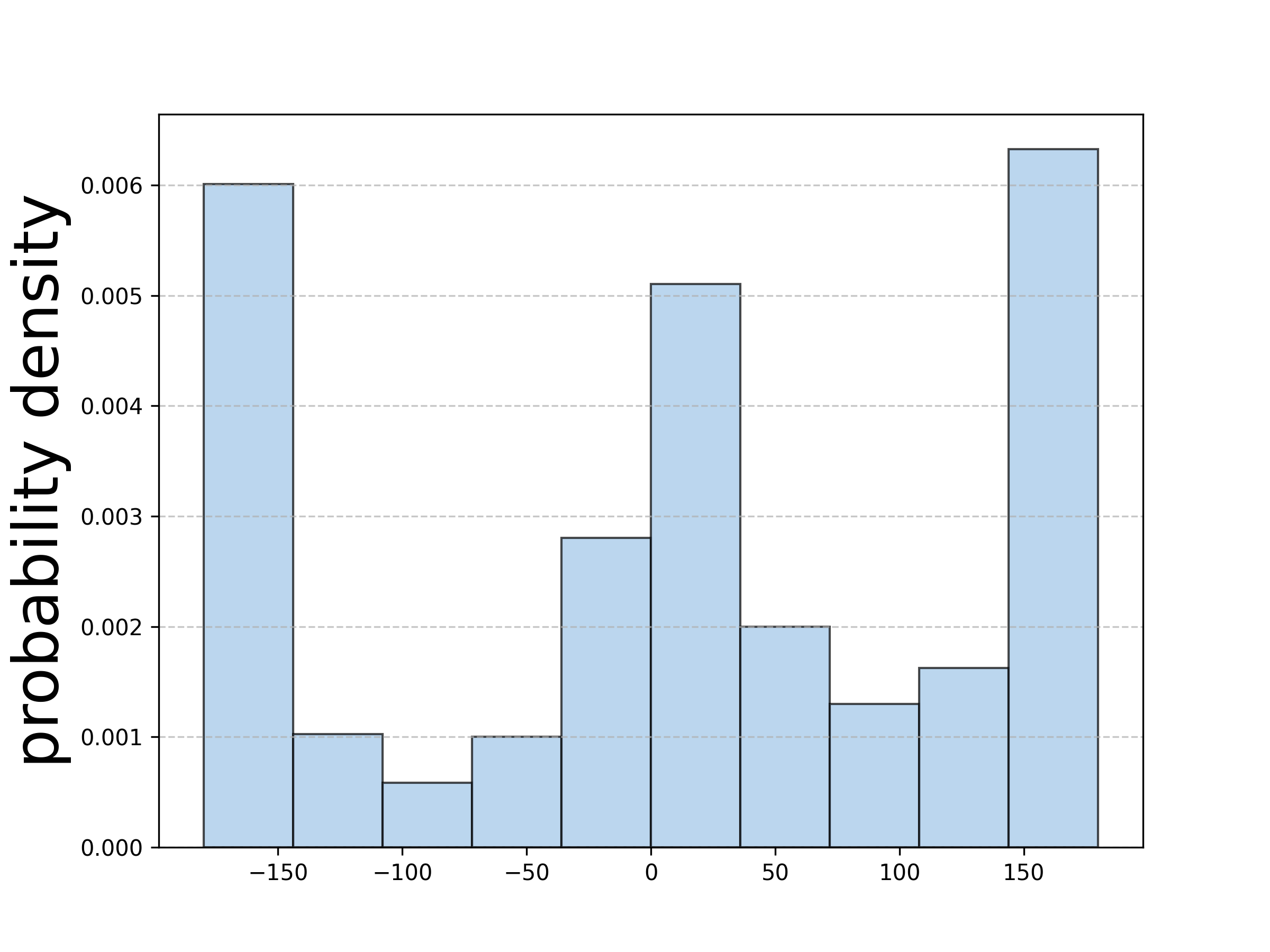}
            \vspace{-20pt}
            \caption[]%
            {{angle (degrees)}}
        \end{subfigure}
        \hfill
        
        \vspace{-5pt}
        \caption[]
        {
        The distribution of ground-truth answer locations relative to CAV in \namedataset's \namevsplit~Q1: Grounding at a reference location. 
        } 
        \label{fig:stats_v2v_q1}
        \vspace{-10pt}
\end{figure}

\begin{figure}[!t]
        \centering
        \begin{subfigure}[t]{0.23\textwidth}
            \centering 
            \includegraphics[width=\textwidth]{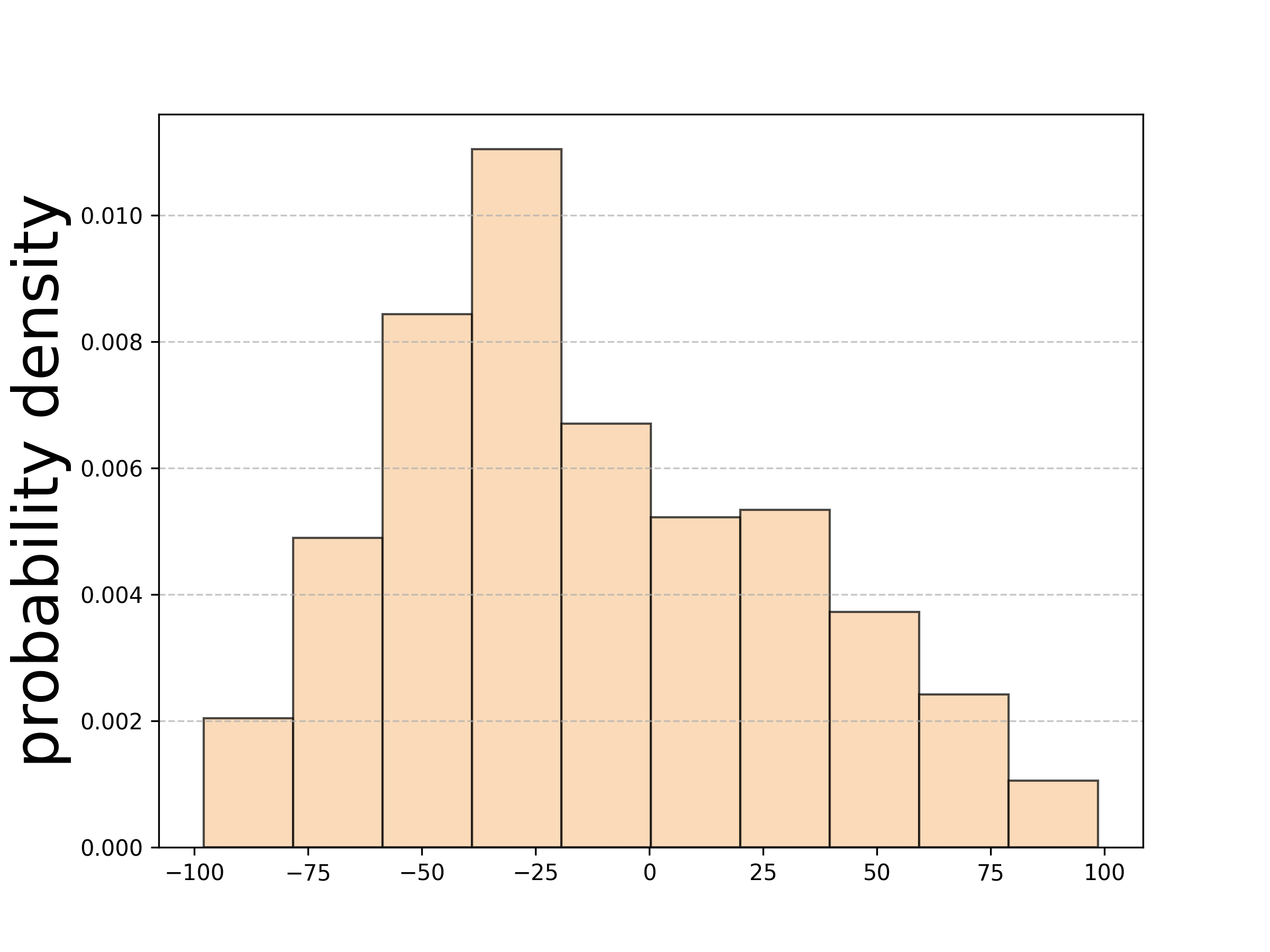}
            \vspace{-20pt}
            \caption[]%
            {{x (meters)}}    
        \end{subfigure}
        \hfill
        \begin{subfigure}[t]{0.23\textwidth}  
            \centering 
            \includegraphics[width=\textwidth]{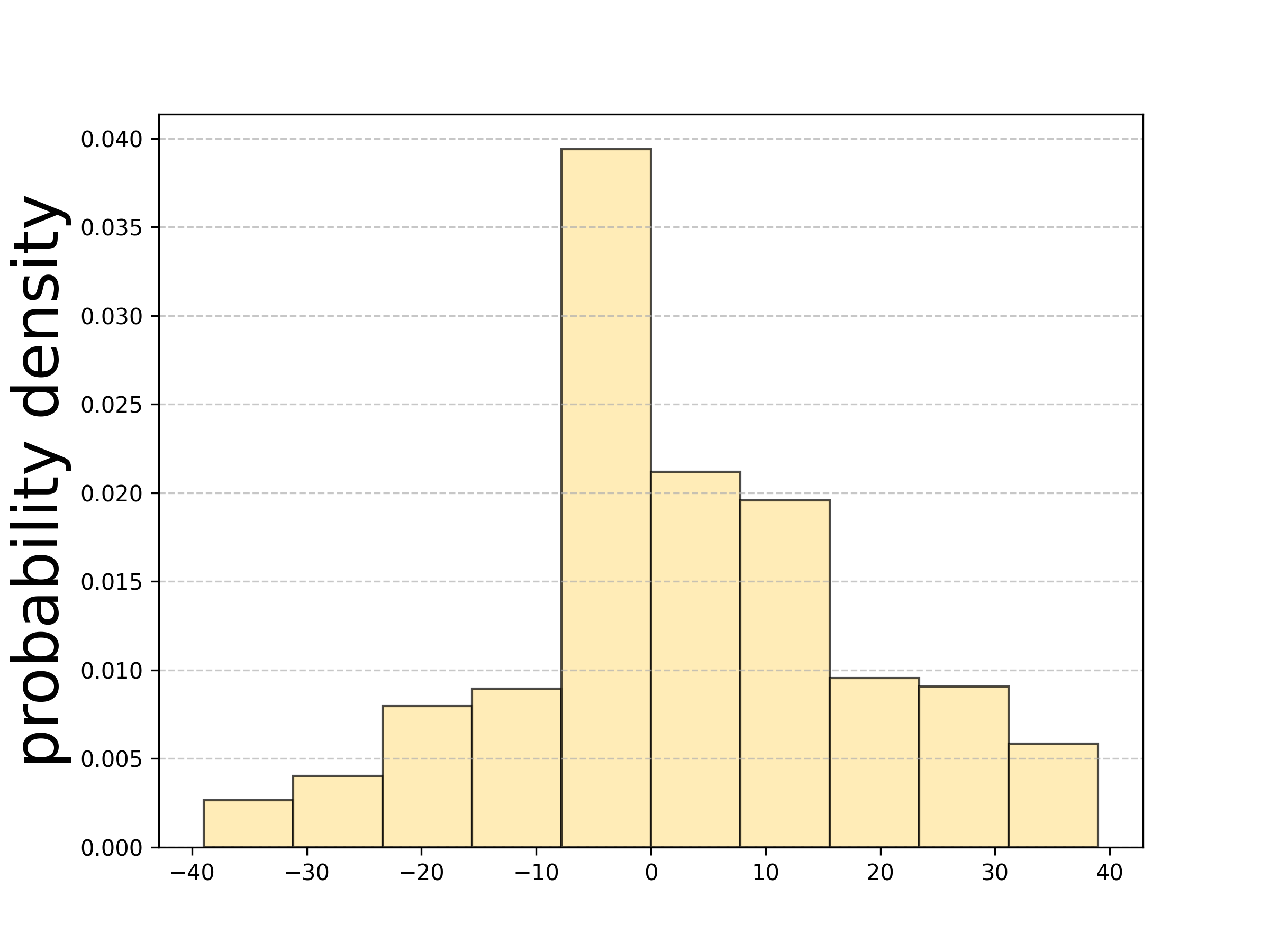}
            \vspace{-20pt}
            \caption[]%
            {{y (meters)}}
        \end{subfigure}

        \begin{subfigure}[t]{0.23\textwidth}
            \centering 
            \includegraphics[width=\textwidth]{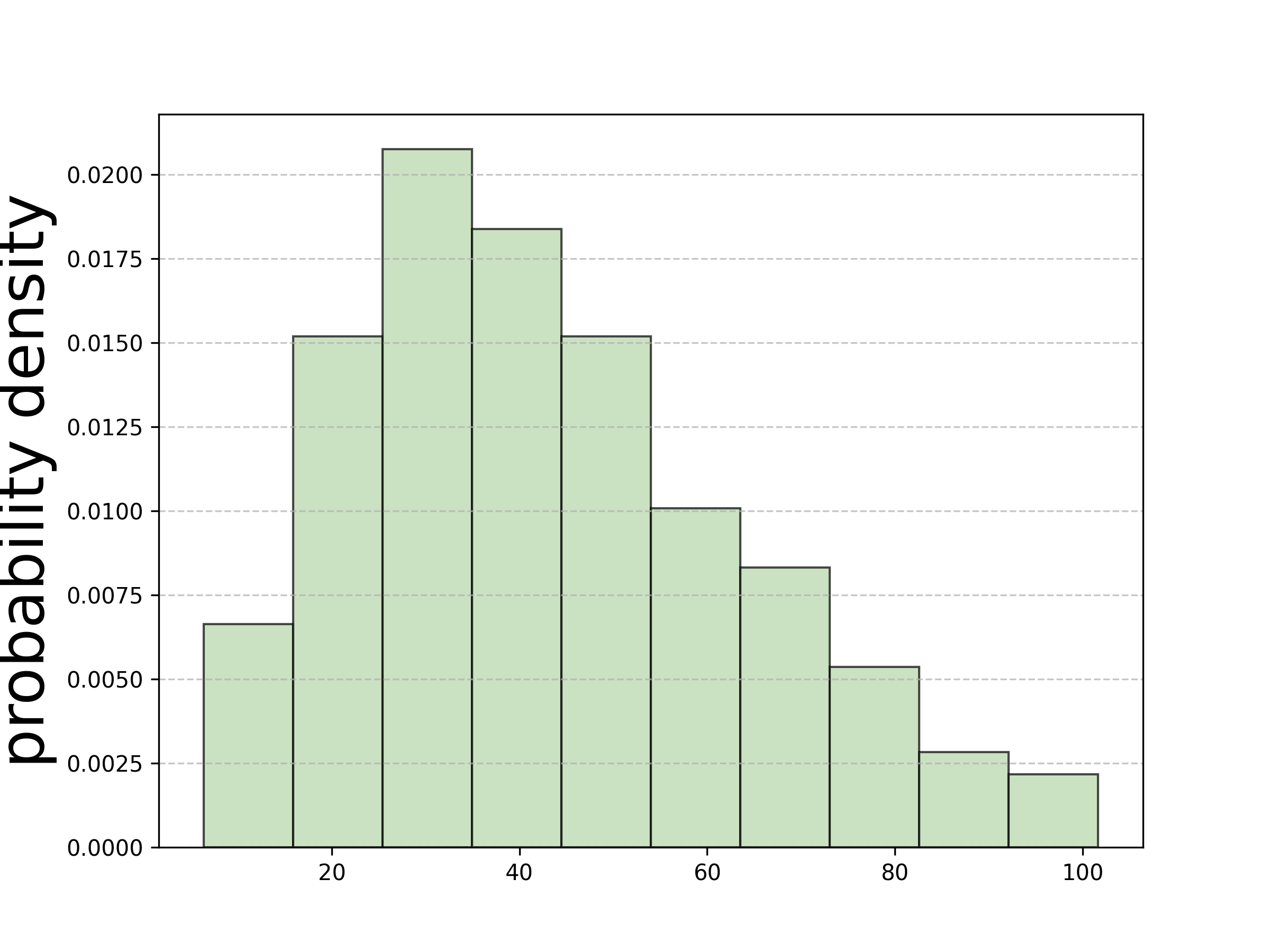}
            \vspace{-20pt}
            \caption[]%
            {{distance (meters)}}
        \end{subfigure}
        \hfill
        \begin{subfigure}[t]{0.23\textwidth}
            \centering 
            \includegraphics[width=\textwidth]{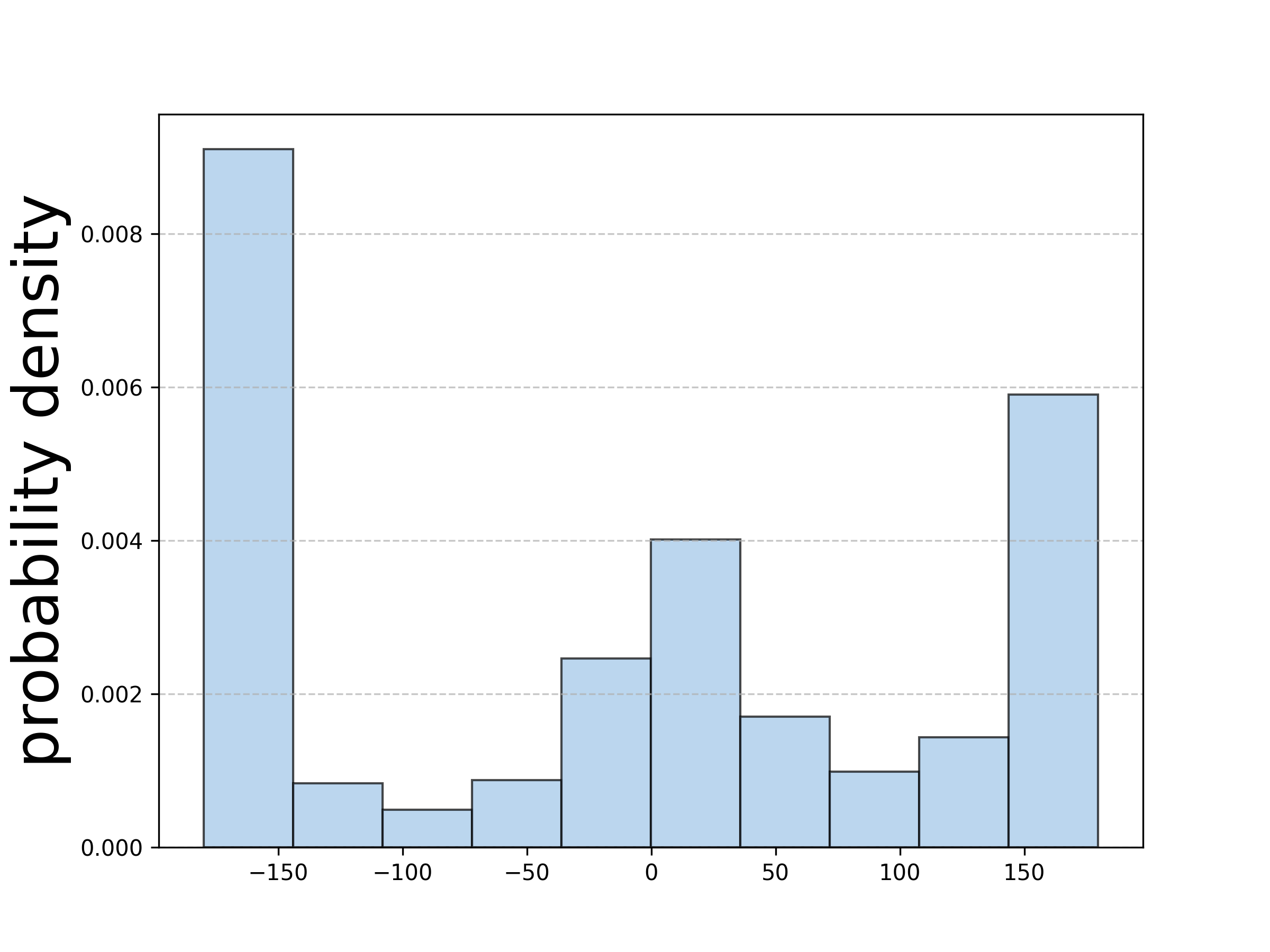}
            \vspace{-20pt}
            \caption[]%
            {{angle (degrees)}}
        \end{subfigure}
        \hfill
        
        \vspace{-5pt}
        \caption[]
        {
        The distribution of ground-truth answer locations relative to CAV in \namedataset's \namevsplit~Q2: Grounding behind a reference object at a location. 
        } 
        \label{fig:stats_v2v_q2}
        \vspace{-10pt}
\end{figure}

\begin{figure}[!t]
        \centering
        \begin{subfigure}[t]{0.23\textwidth}
            \centering 
            \includegraphics[width=\textwidth]{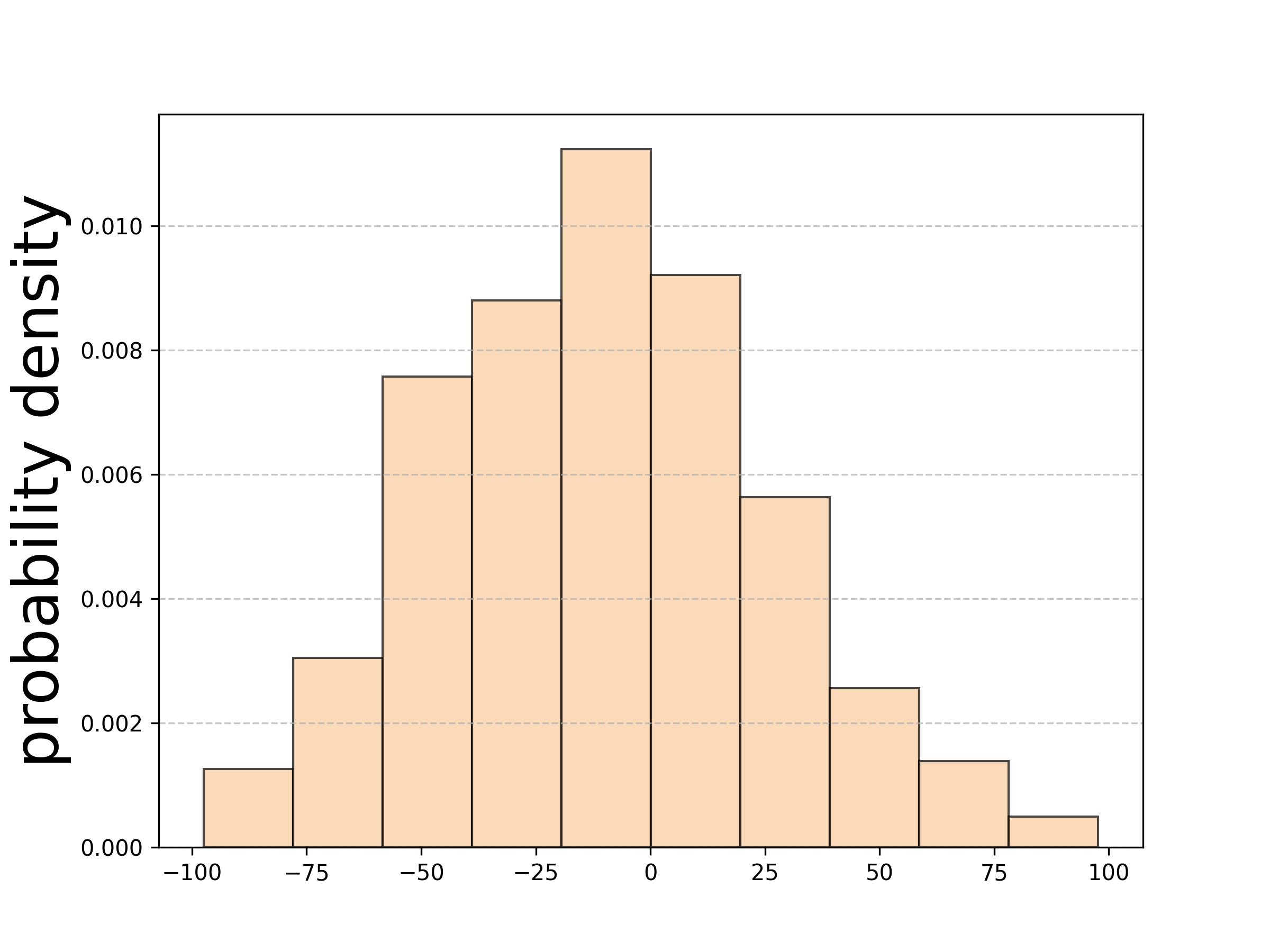}
            \vspace{-20pt}
            \caption[]%
            {{x (meters)}}    
        \end{subfigure}
        \hfill
        \begin{subfigure}[t]{0.23\textwidth}  
            \centering 
            \includegraphics[width=\textwidth]
            {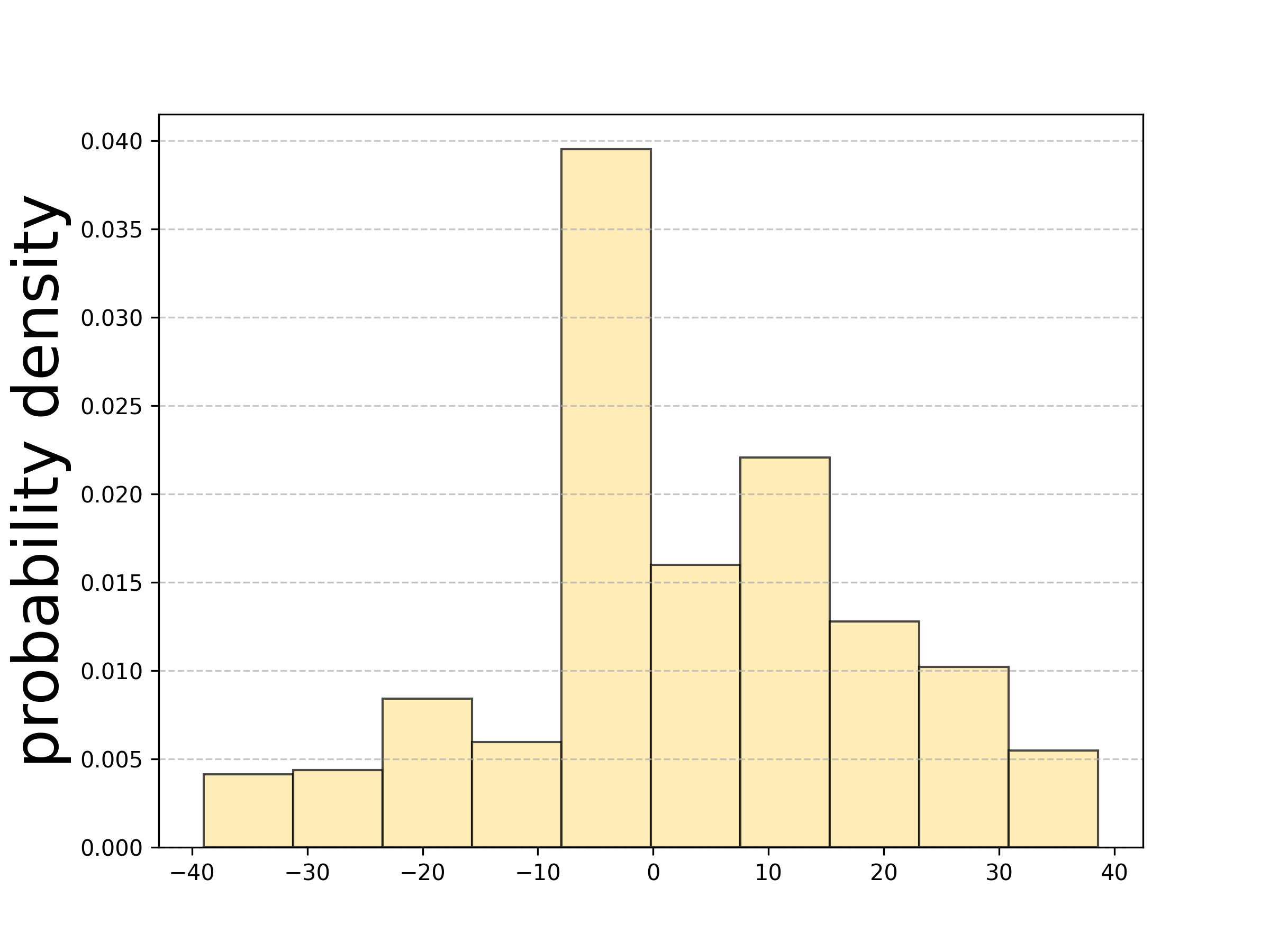}
            \vspace{-20pt}
            \caption[]%
            {{y (meters)}}
        \end{subfigure}

        \begin{subfigure}[t]{0.23\textwidth}
            \centering 
            \includegraphics[width=\textwidth]{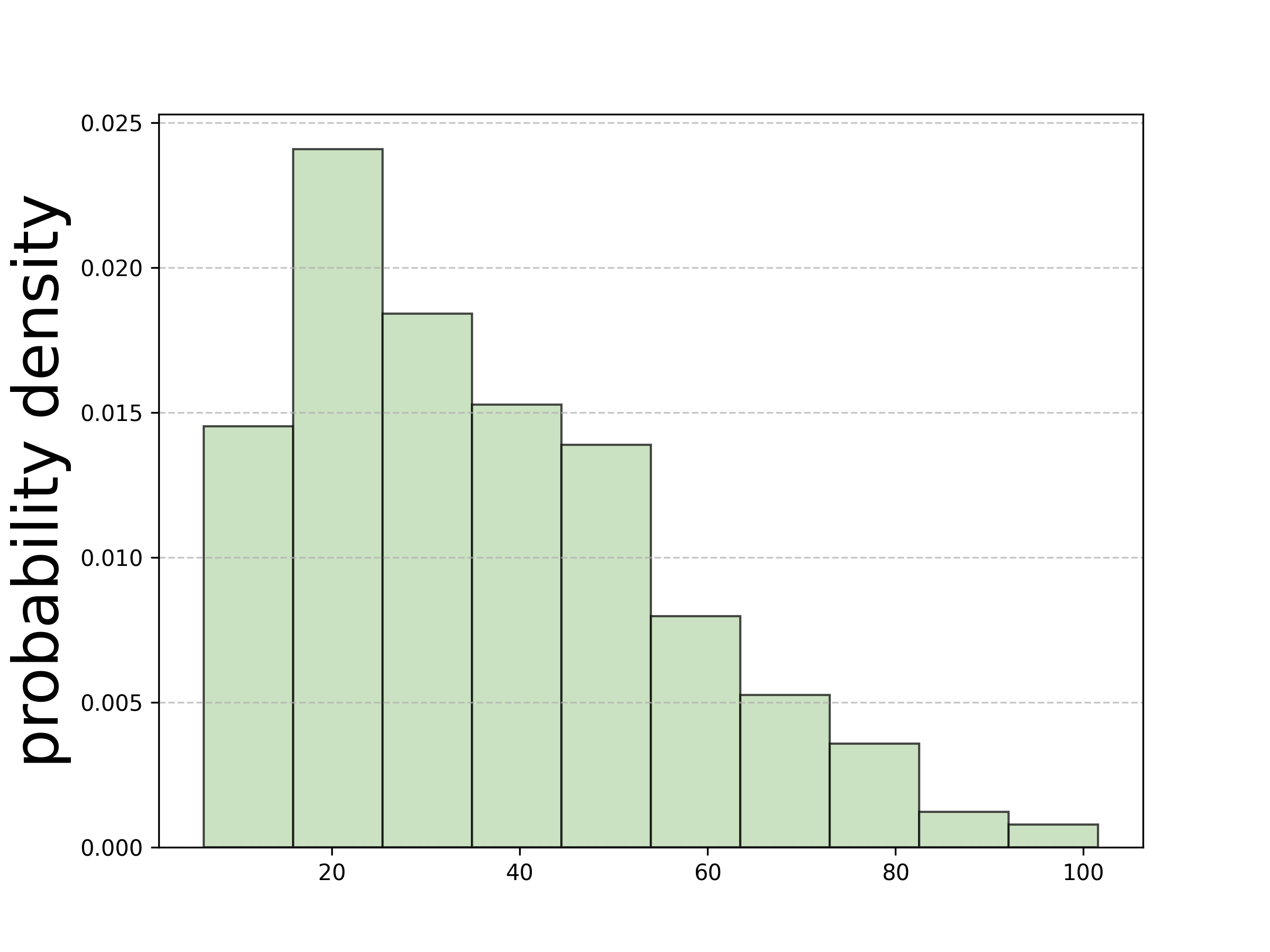}
            \vspace{-20pt}
            \caption[]%
            {{distance (meters)}}
        \end{subfigure}
        \hfill
        \begin{subfigure}[t]{0.23\textwidth}
            \centering 
            \includegraphics[width=\textwidth]{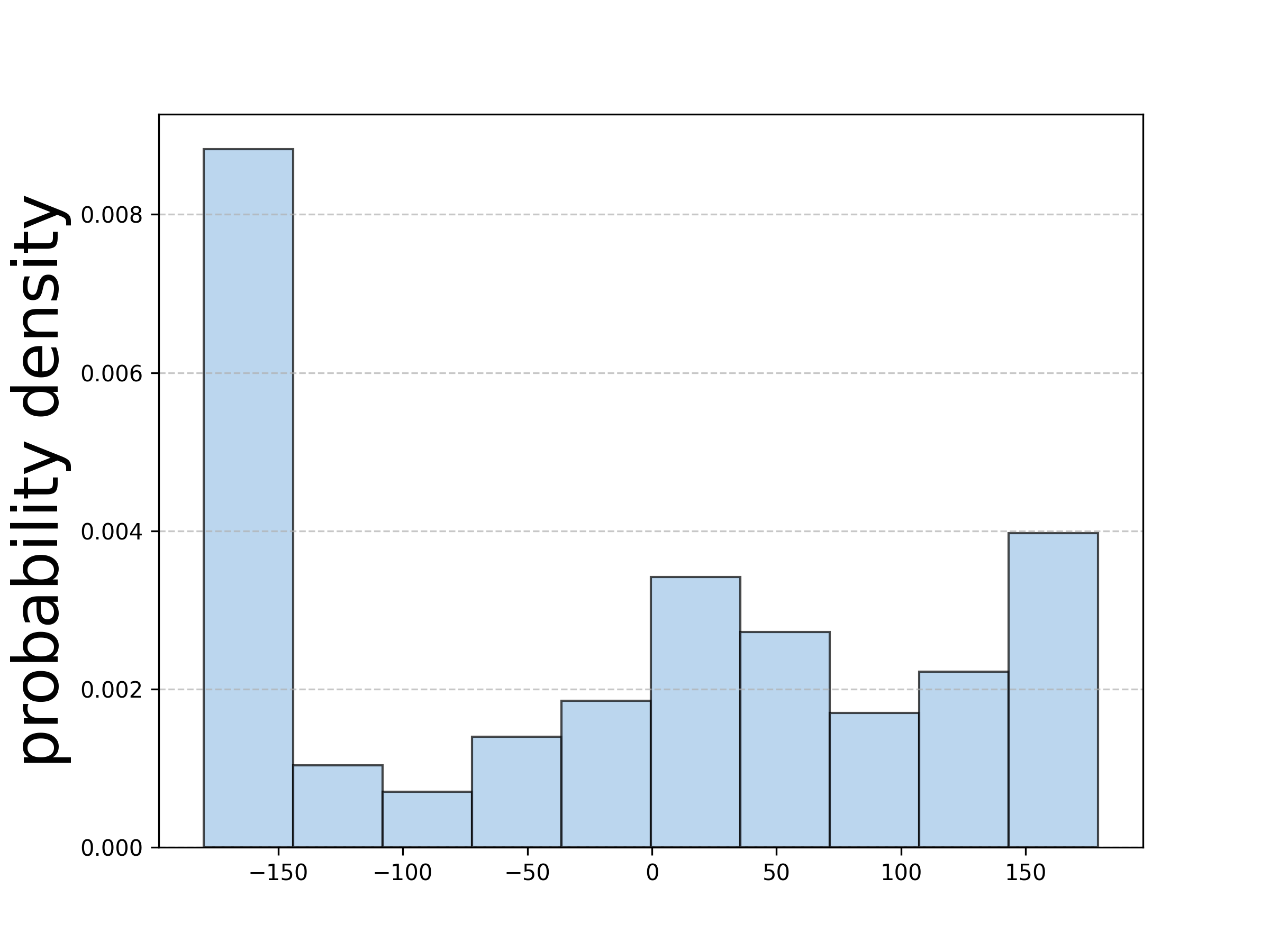}
            \vspace{-20pt}
            \caption[]%
            {{angle (degrees)}}
        \end{subfigure}
        \hfill
        
        \vspace{-5pt}
        \caption[]
        {
        The distribution of ground-truth answer locations relative to CAV in \namedataset's \namevsplit~Q3: Grounding behind a reference object in a direction. 
        } 
        \label{fig:stats_v2v_q3}
        \vspace{-10pt}
\end{figure}

\begin{figure}[!t]
        \centering
        \begin{subfigure}[t]{0.23\textwidth}
            \centering 
            \includegraphics[width=\textwidth]{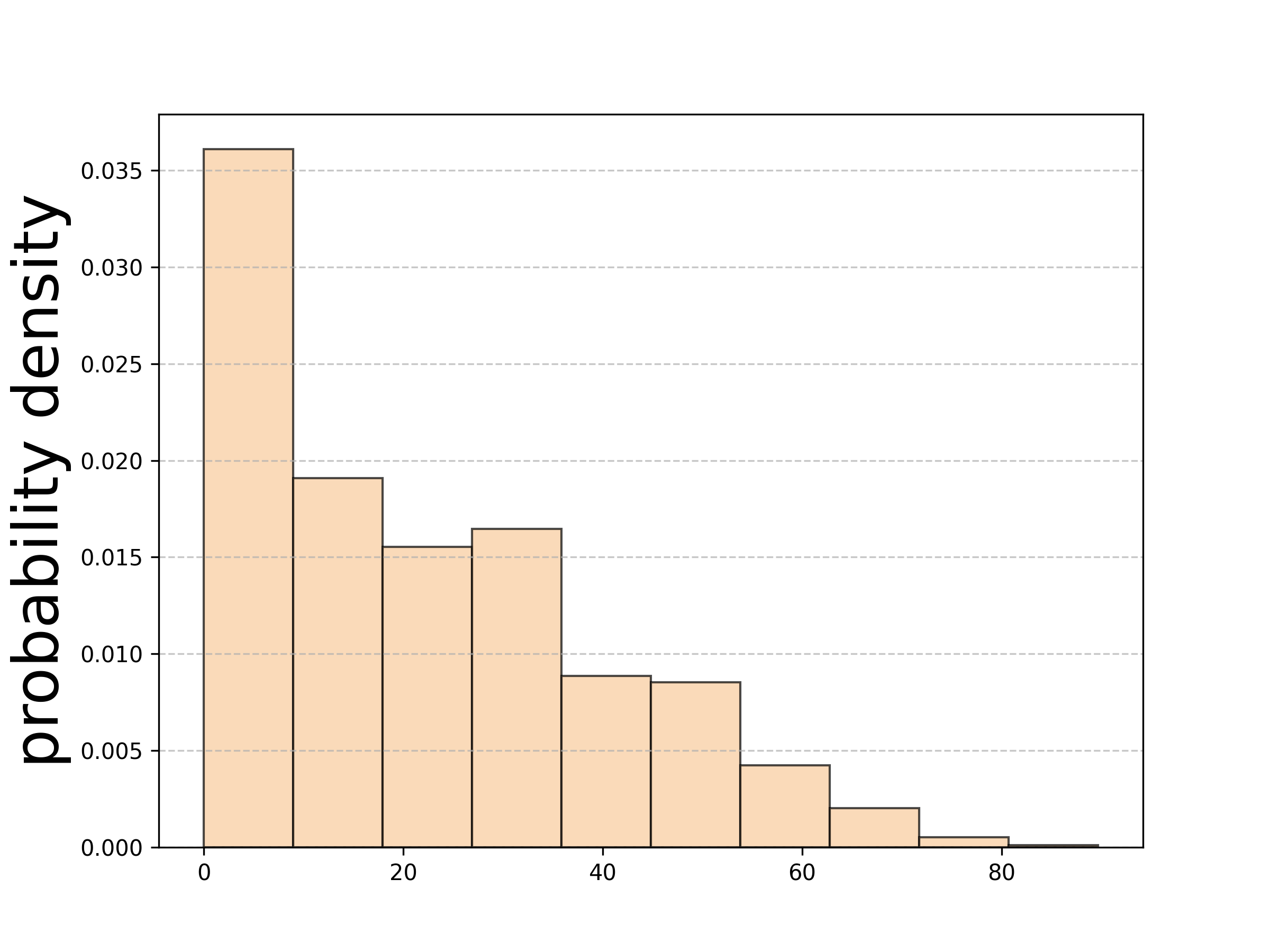}
            \vspace{-20pt}
            \caption[]%
            {{x (meters)}}    
        \end{subfigure}
        \hfill
        \begin{subfigure}[t]{0.23\textwidth}  
            \centering 
            \includegraphics[width=\textwidth]{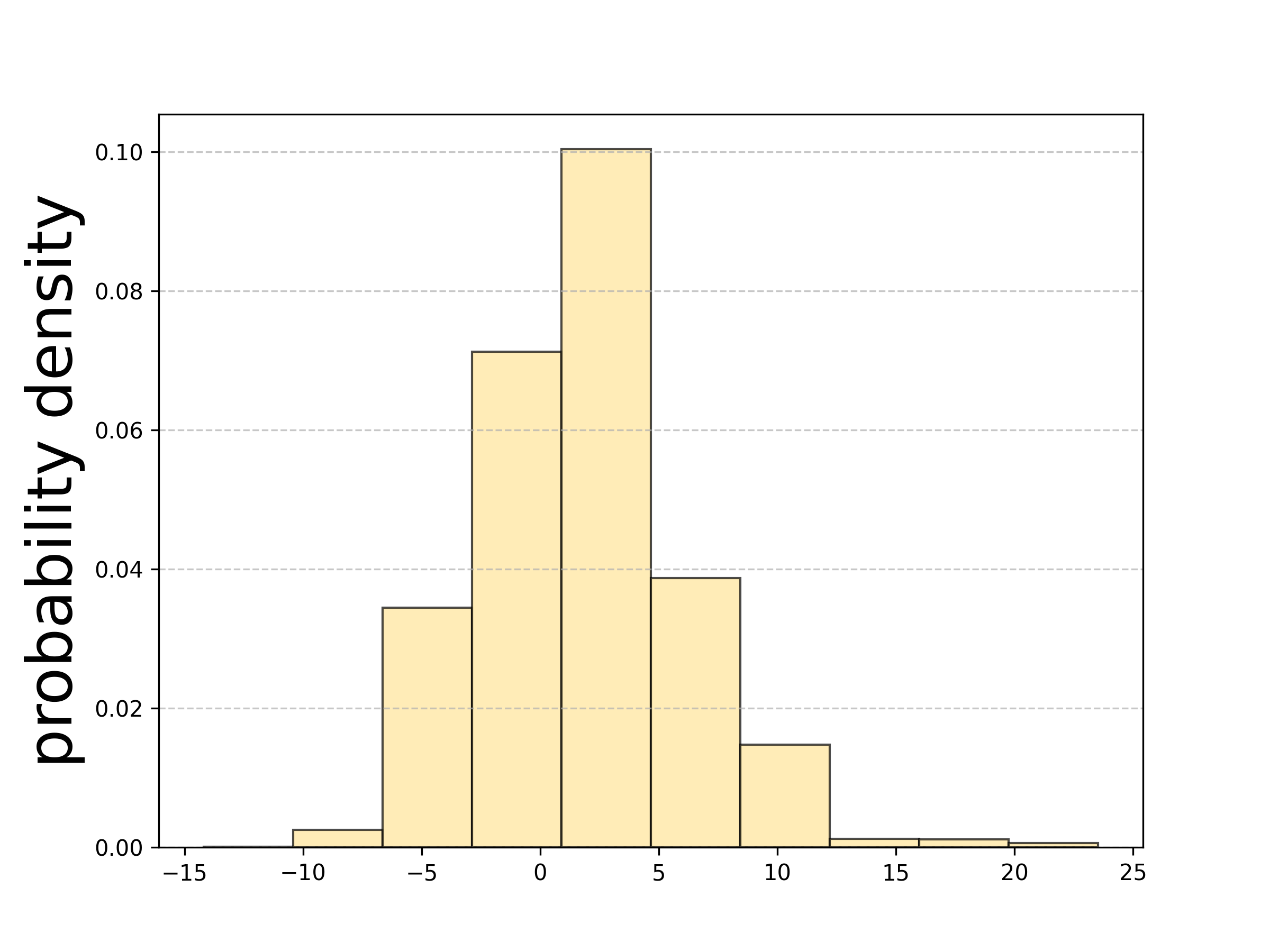}
            \vspace{-20pt}
            \caption[]%
            {{y (meters)}}
        \end{subfigure}

        \begin{subfigure}[t]{0.23\textwidth}
            \centering 
            \includegraphics[width=\textwidth]{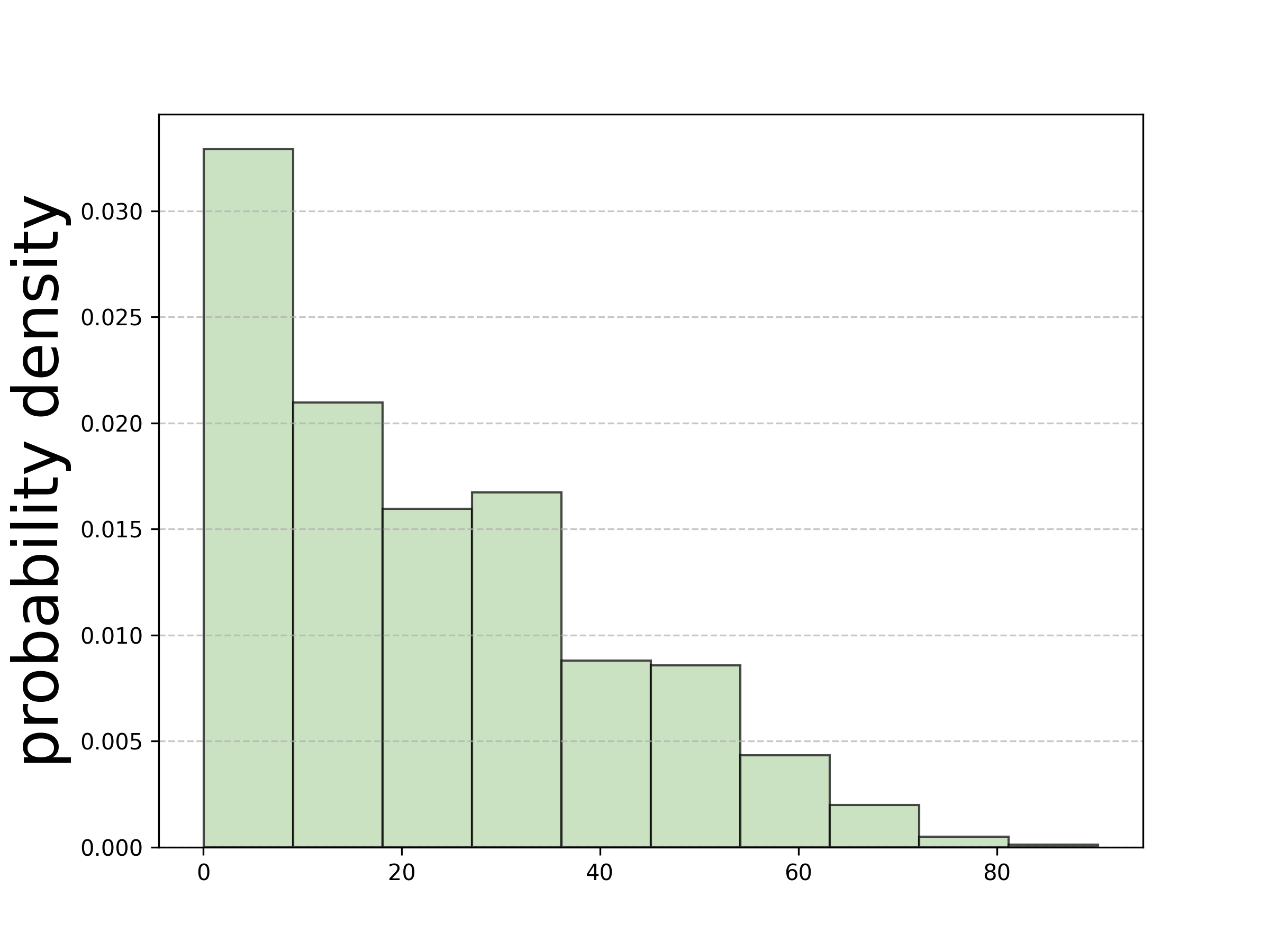}
            \vspace{-20pt}
            \caption[]%
            {{distance (meters)}}
        \end{subfigure}
        \hfill
        \begin{subfigure}[t]{0.23\textwidth}
            \centering 
            \includegraphics[width=\textwidth]{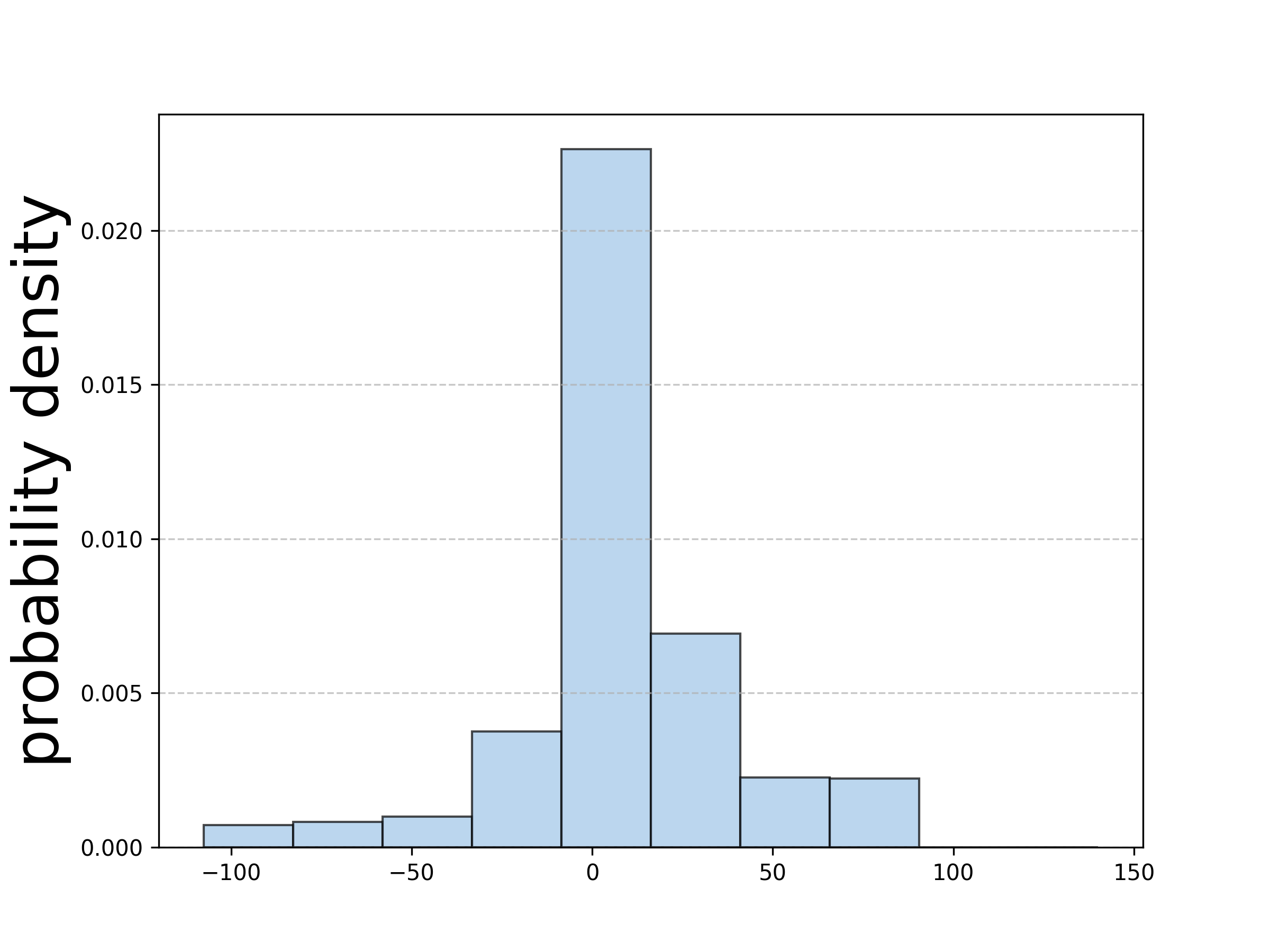}
            \vspace{-20pt}
            \caption[]%
            {{angle (degrees)}}
        \end{subfigure}
        \hfill
        
        \vspace{-5pt}
        \caption[]
        {
        The distribution of ground-truth answer locations relative to CAV in \namedataset's \namevsplit~Q4: Notable object identification. 
        } 
        \label{fig:stats_v2v_q4}
        \vspace{-10pt}
\end{figure}

\begin{figure}[!t]
        \centering
        \begin{subfigure}[t]{0.23\textwidth}
            \centering 
            \includegraphics[width=\textwidth]{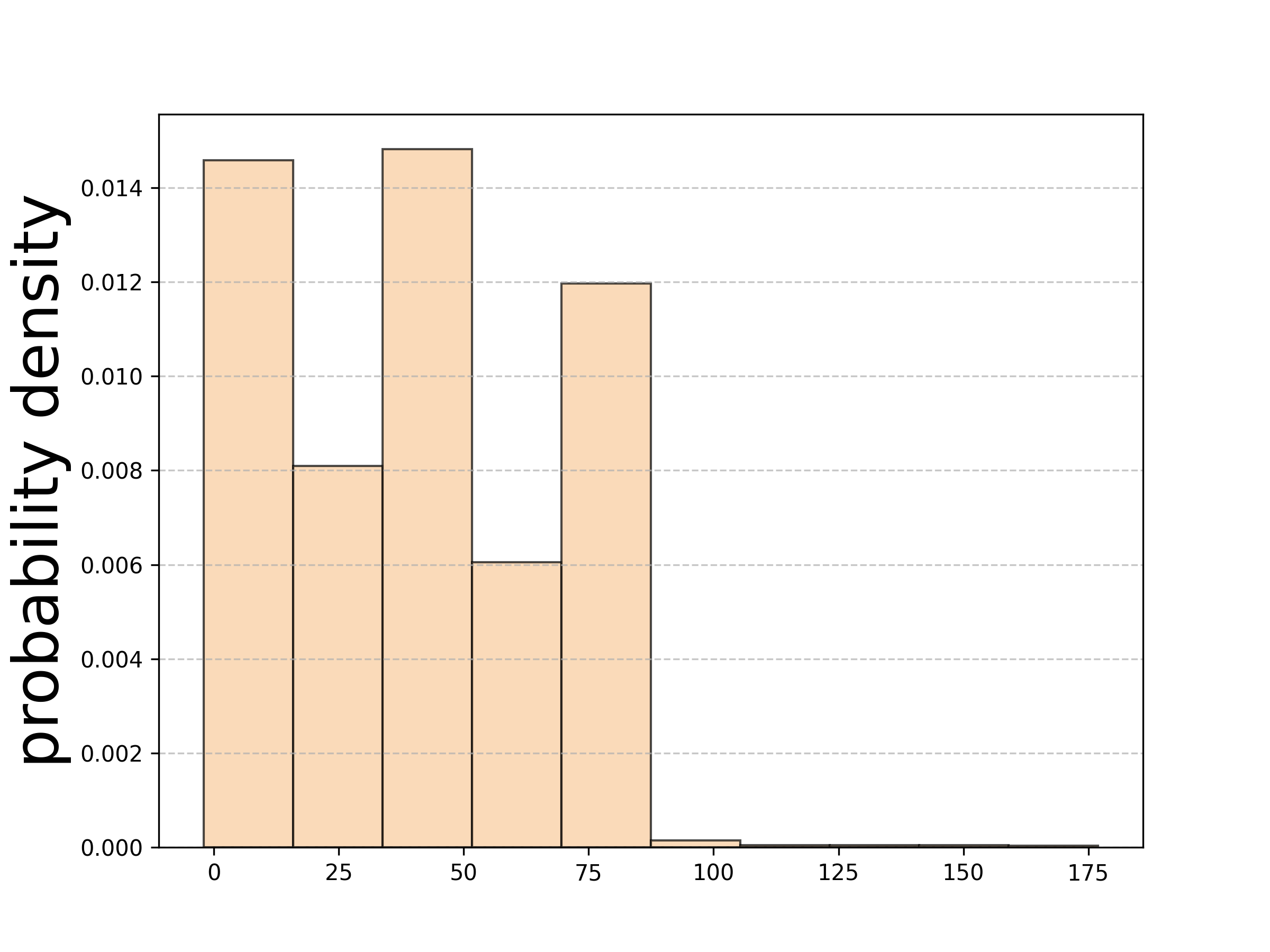}
            \vspace{-20pt}
            \caption[]%
            {{x (meters)}}    
        \end{subfigure}
        \hfill
        \begin{subfigure}[t]{0.23\textwidth}  
            \centering 
            \includegraphics[width=\textwidth]{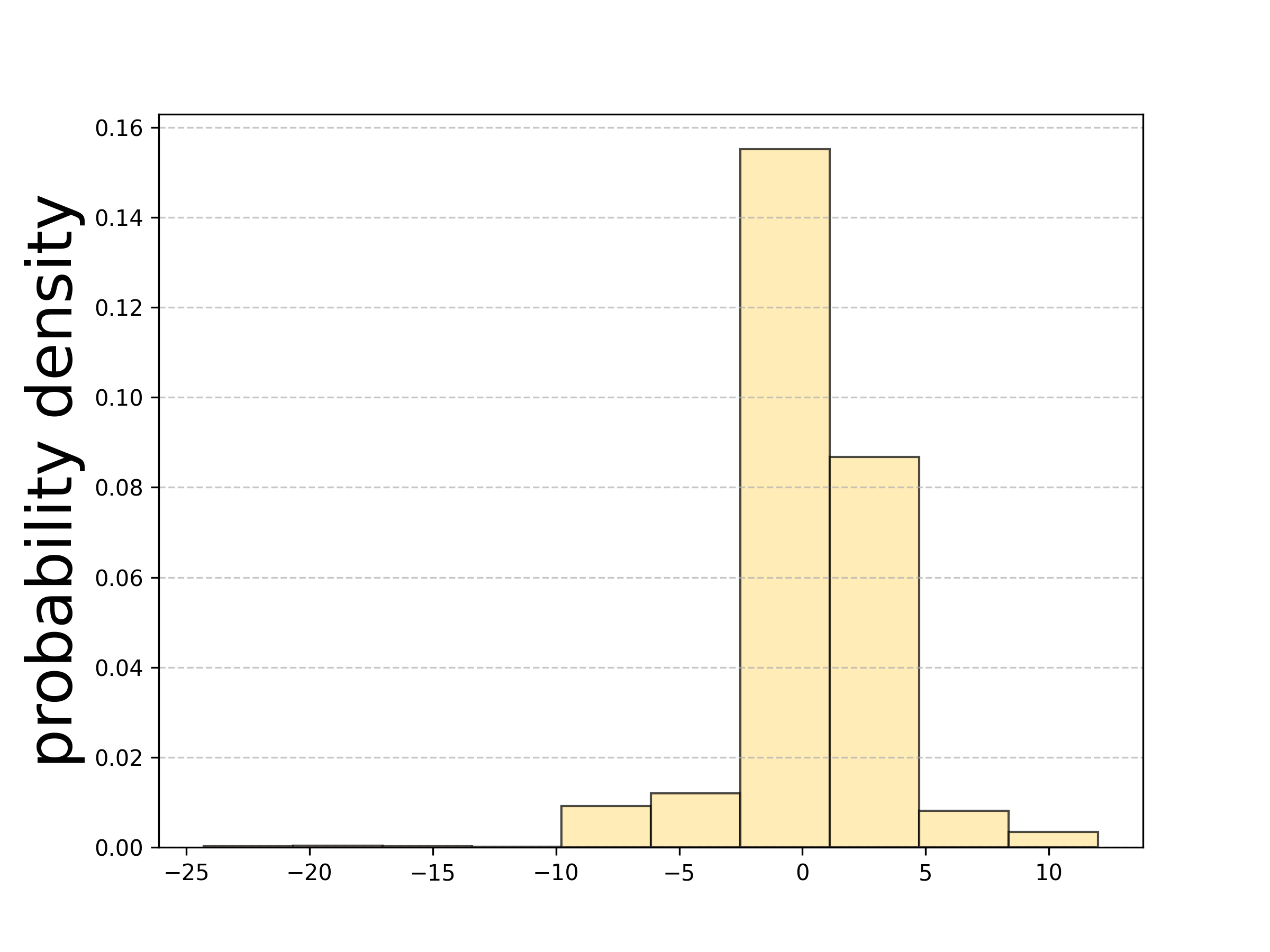}
            \vspace{-20pt}
            \caption[]%
            {{y (meters)}}
        \end{subfigure}

        \begin{subfigure}[t]{0.23\textwidth}
            \centering 
            \includegraphics[width=\textwidth]{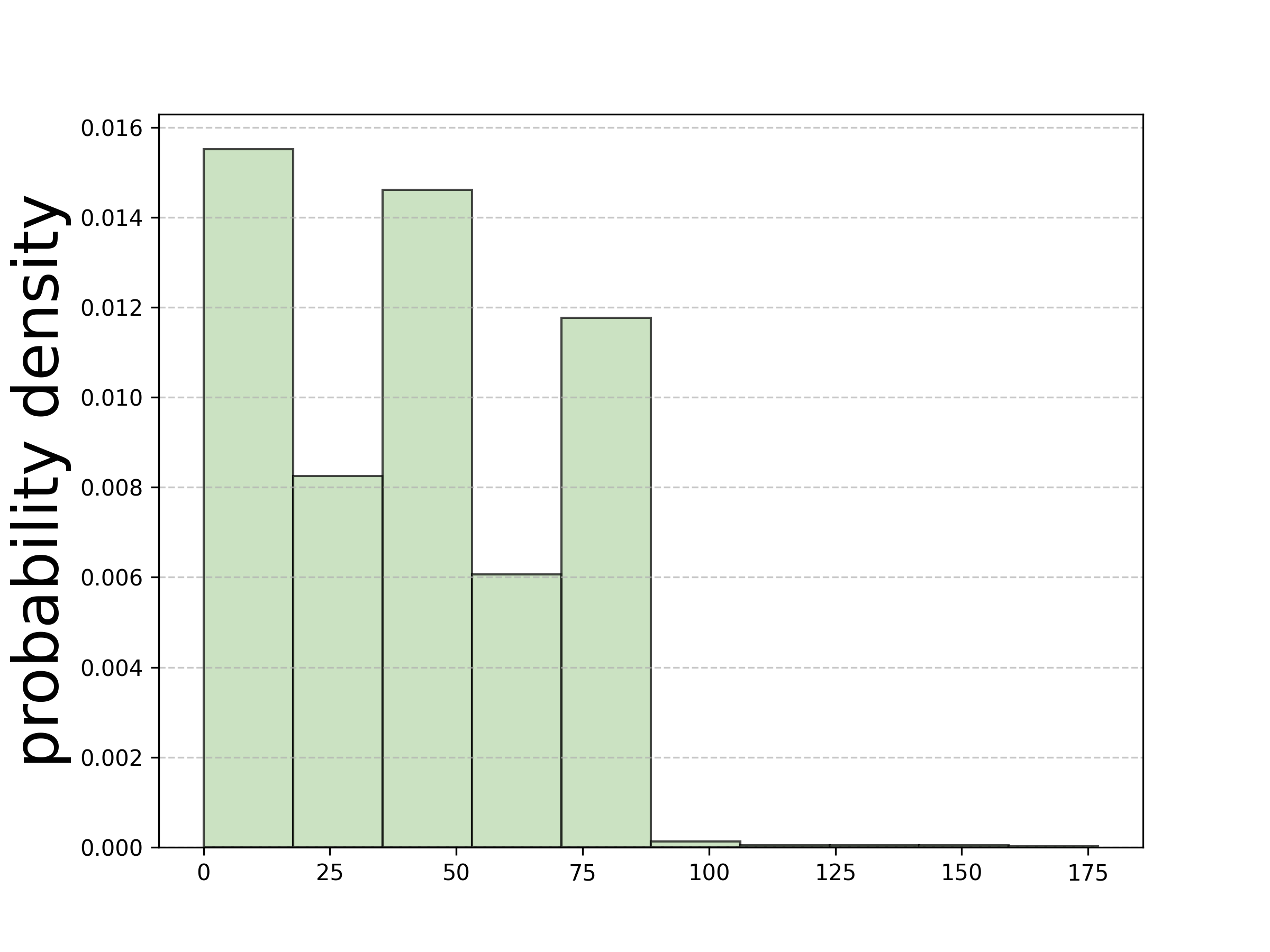}
            \vspace{-20pt}
            \caption[]%
            {{distance (meters)}}
        \end{subfigure}
        \hfill
        \begin{subfigure}[t]{0.23\textwidth}
            \centering 
            \includegraphics[width=\textwidth]{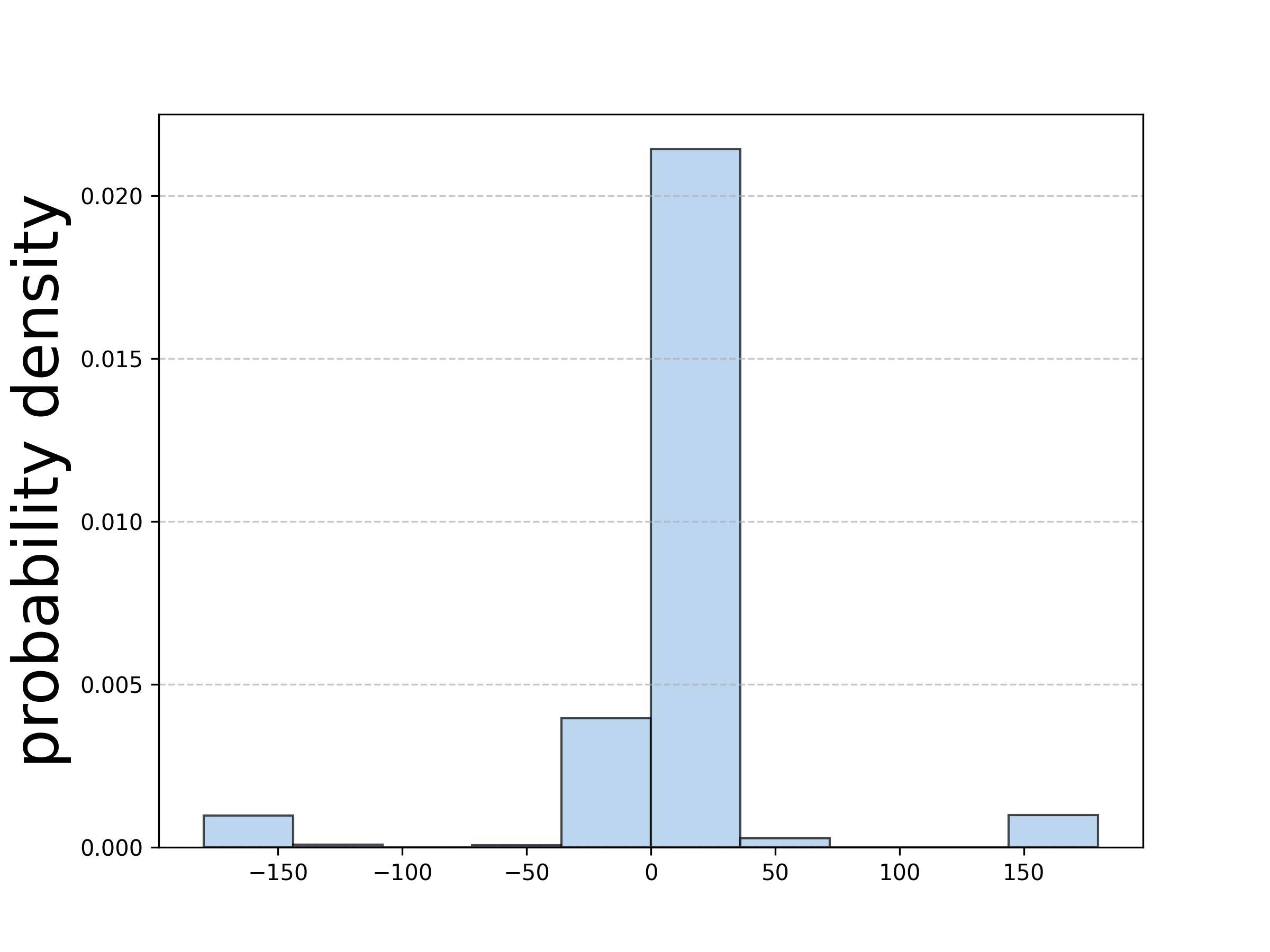}
            \vspace{-20pt}
            \caption[]%
            {{angle (degrees)}}
        \end{subfigure}
        \hfill
        
        \vspace{-5pt}
        \caption[]
        {
        The distribution of ground-truth answer locations relative to CAV in \namedataset's \namevsplit~Q5: Planning. 
        } 
        \label{fig:stats_v2v_q5}
        \vspace{-10pt}
\end{figure}

\begin{figure}[!t]
        \centering
        \begin{subfigure}[t]{0.23\textwidth}
            \centering 
            \includegraphics[width=\textwidth]{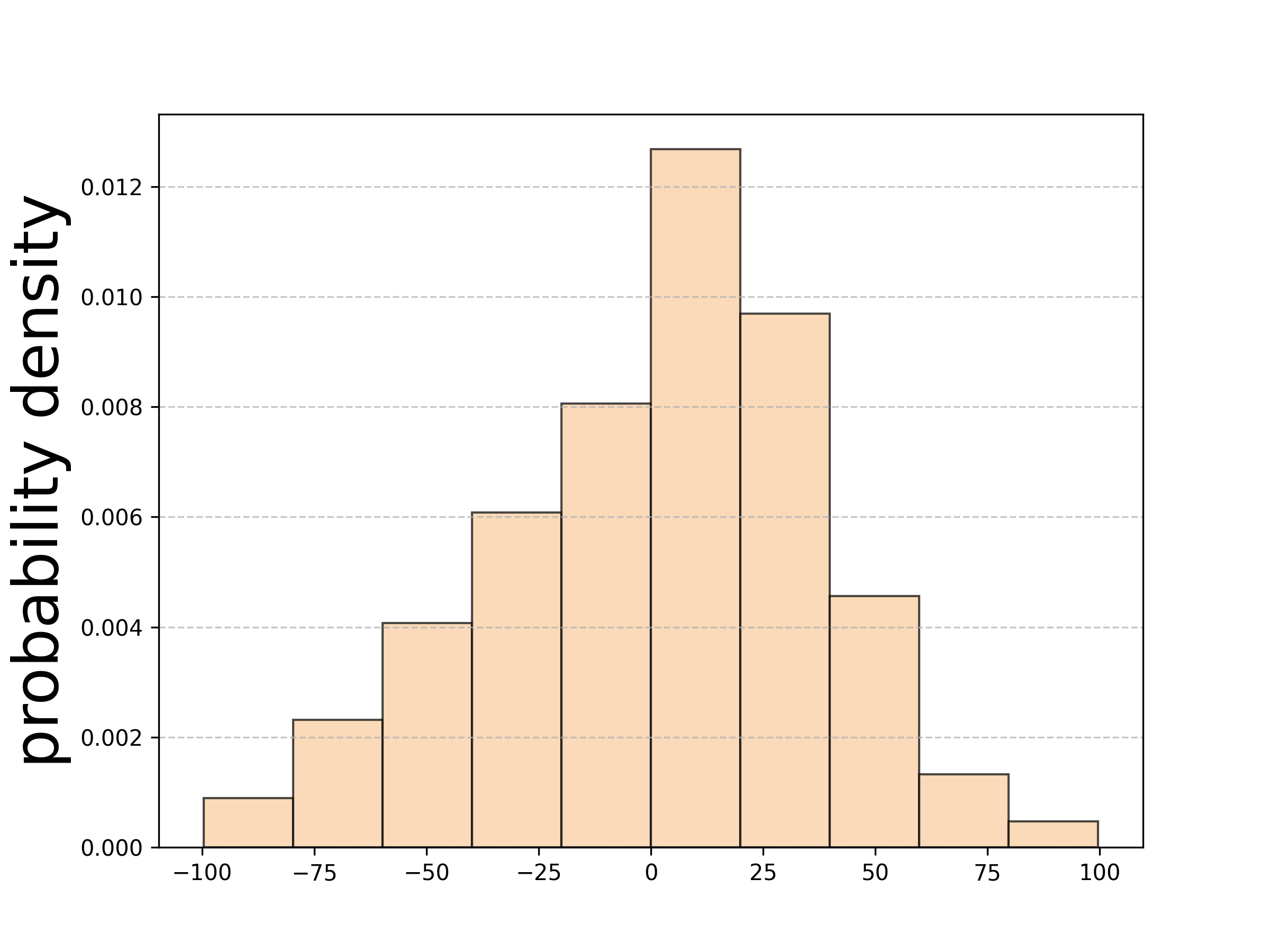}
            \vspace{-20pt}
            \caption[]%
            {{x (meters)}}    
        \end{subfigure}
        \hfill
        \begin{subfigure}[t]{0.23\textwidth}  
            \centering 
            \includegraphics[width=\textwidth]{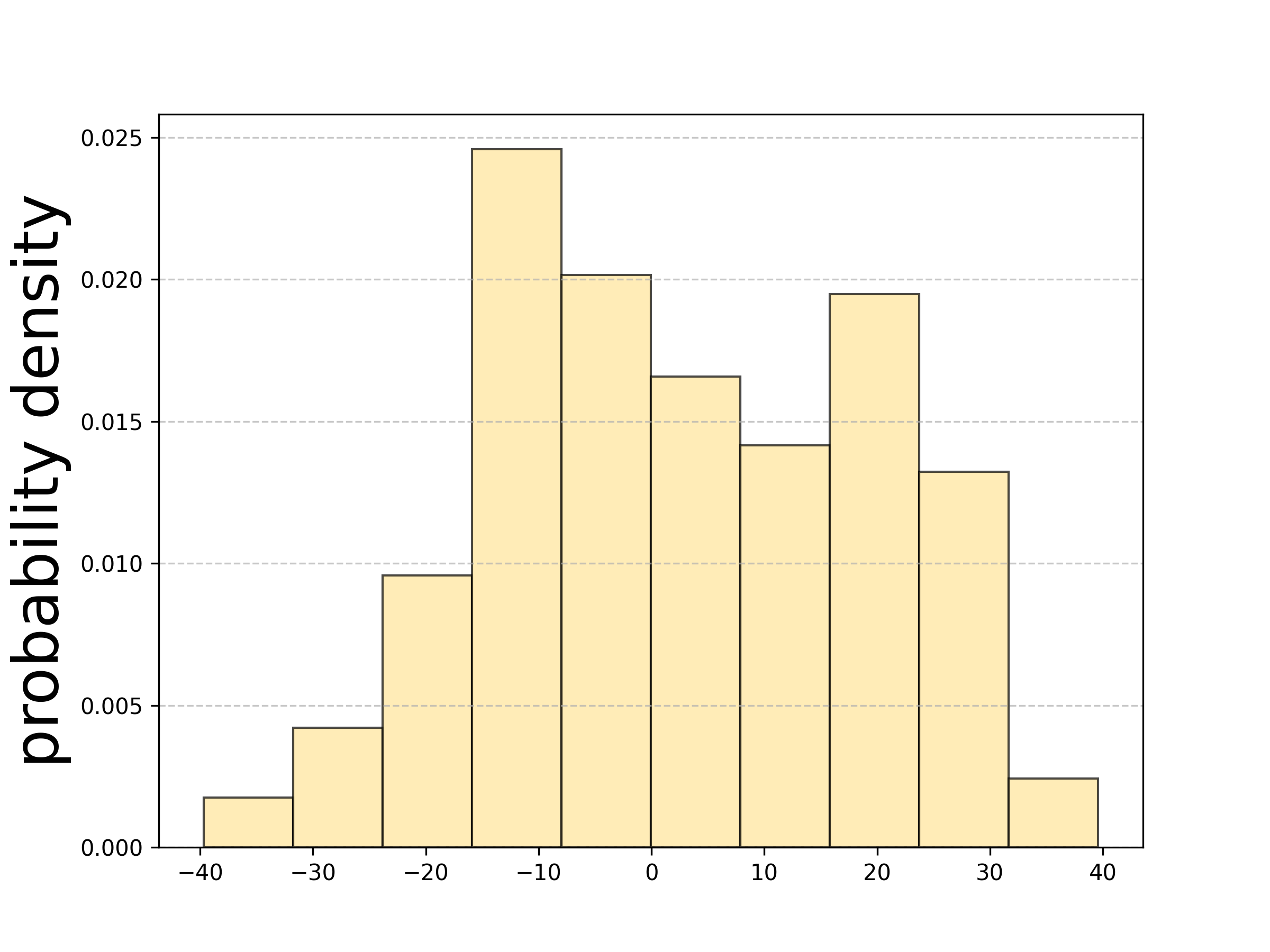}
            \vspace{-20pt}
            \caption[]%
            {{y (meters)}}
        \end{subfigure}

        \begin{subfigure}[t]{0.23\textwidth}
            \centering 
            \includegraphics[width=\textwidth]{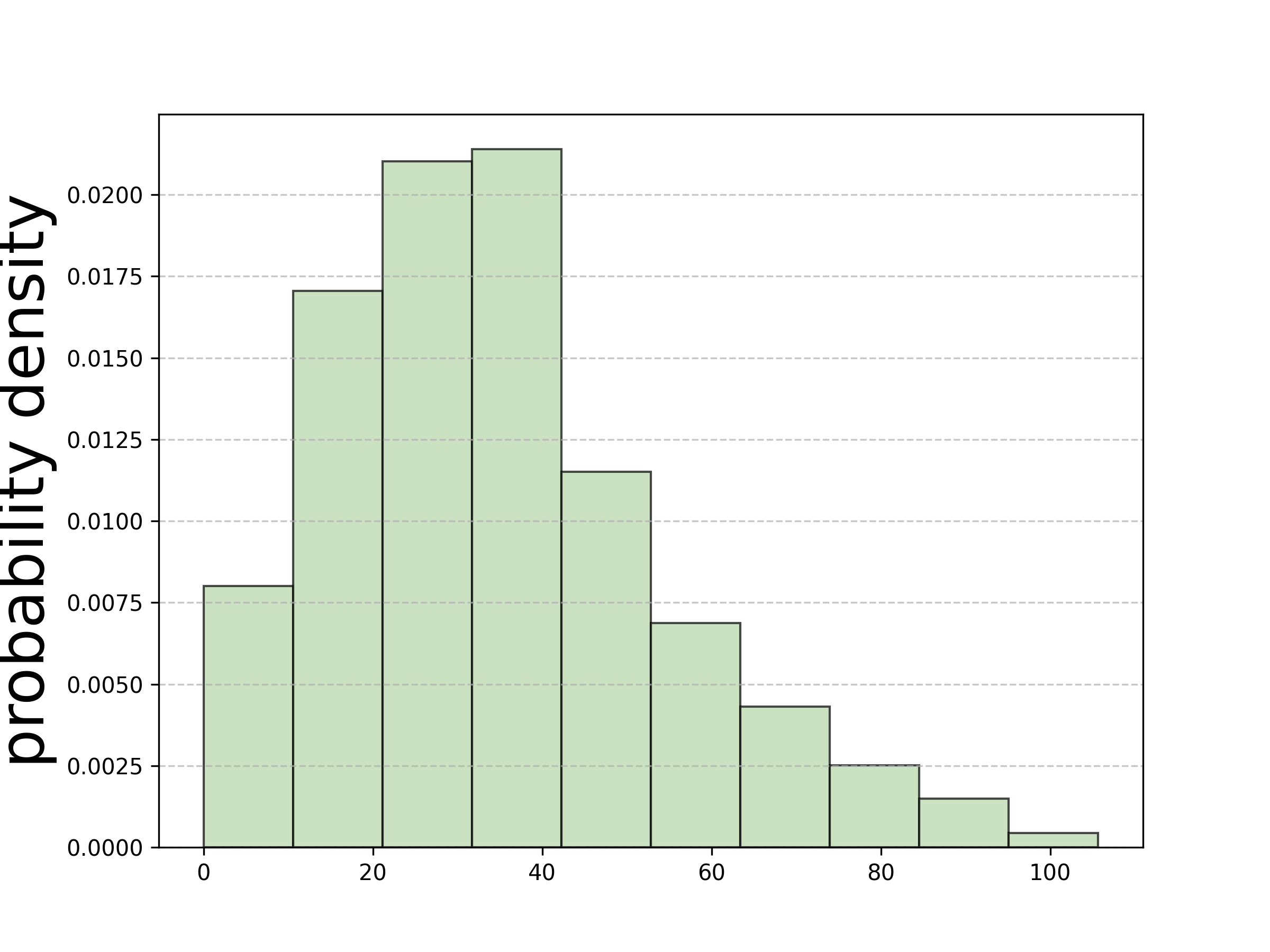}
            \vspace{-20pt}
            \caption[]%
            {{distance (meters)}}
        \end{subfigure}
        \hfill
        \begin{subfigure}[t]{0.23\textwidth}
            \centering 
            \includegraphics[width=\textwidth]{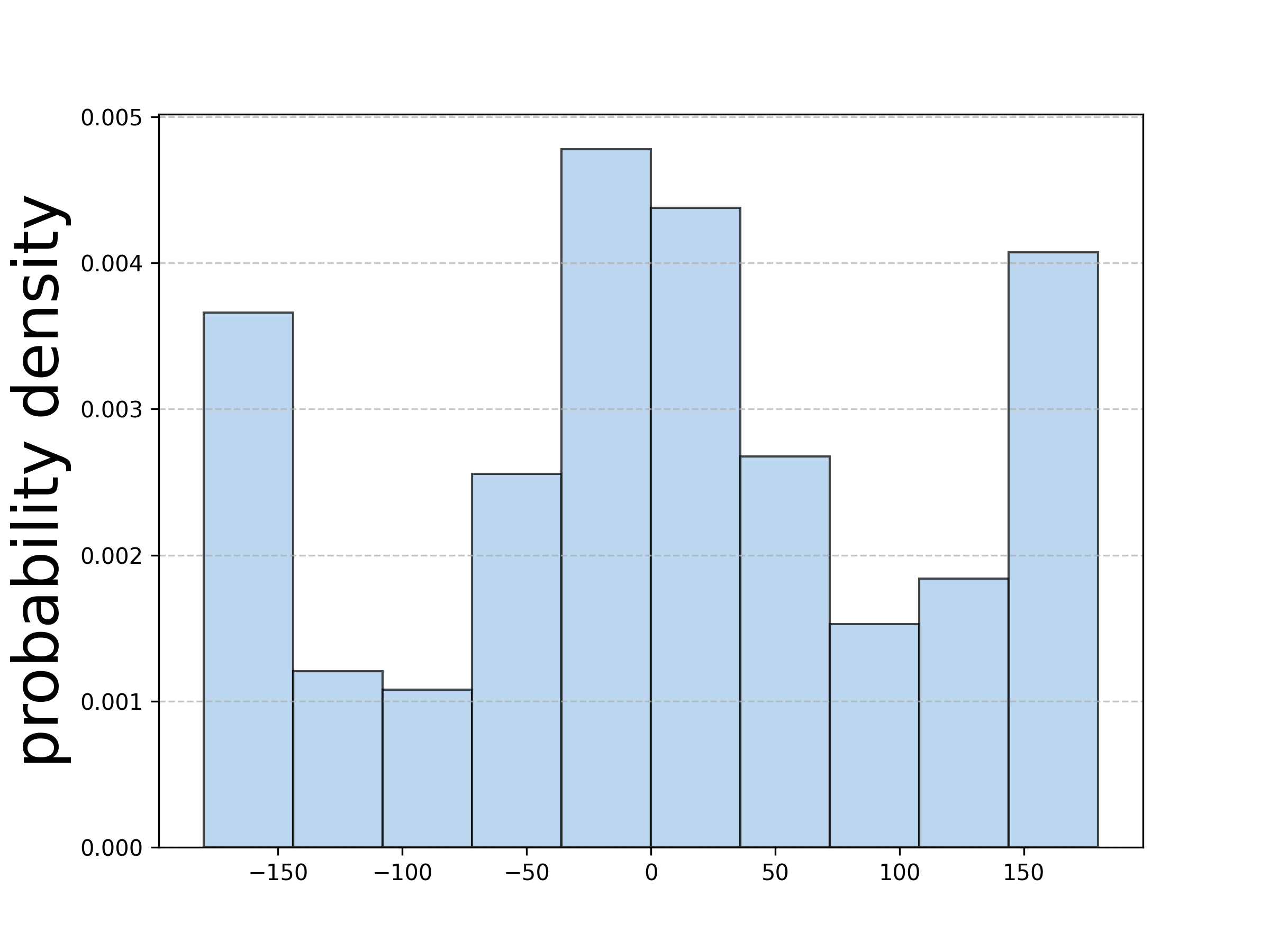}
            \vspace{-20pt}
            \caption[]%
            {{angle (degrees)}}
        \end{subfigure}
        \hfill
        
        \vspace{-5pt}
        \caption[]
        {
        The distribution of ground-truth answer locations relative to CAV in \namedataset's \namexsplit~Q1: Grounding at a reference location. 
        } 
        \label{fig:stats_v2x_q1}
        \vspace{-10pt}
\end{figure}

\begin{figure}[!t]
        \centering
        \begin{subfigure}[t]{0.23\textwidth}
            \centering 
            \includegraphics[width=\textwidth]{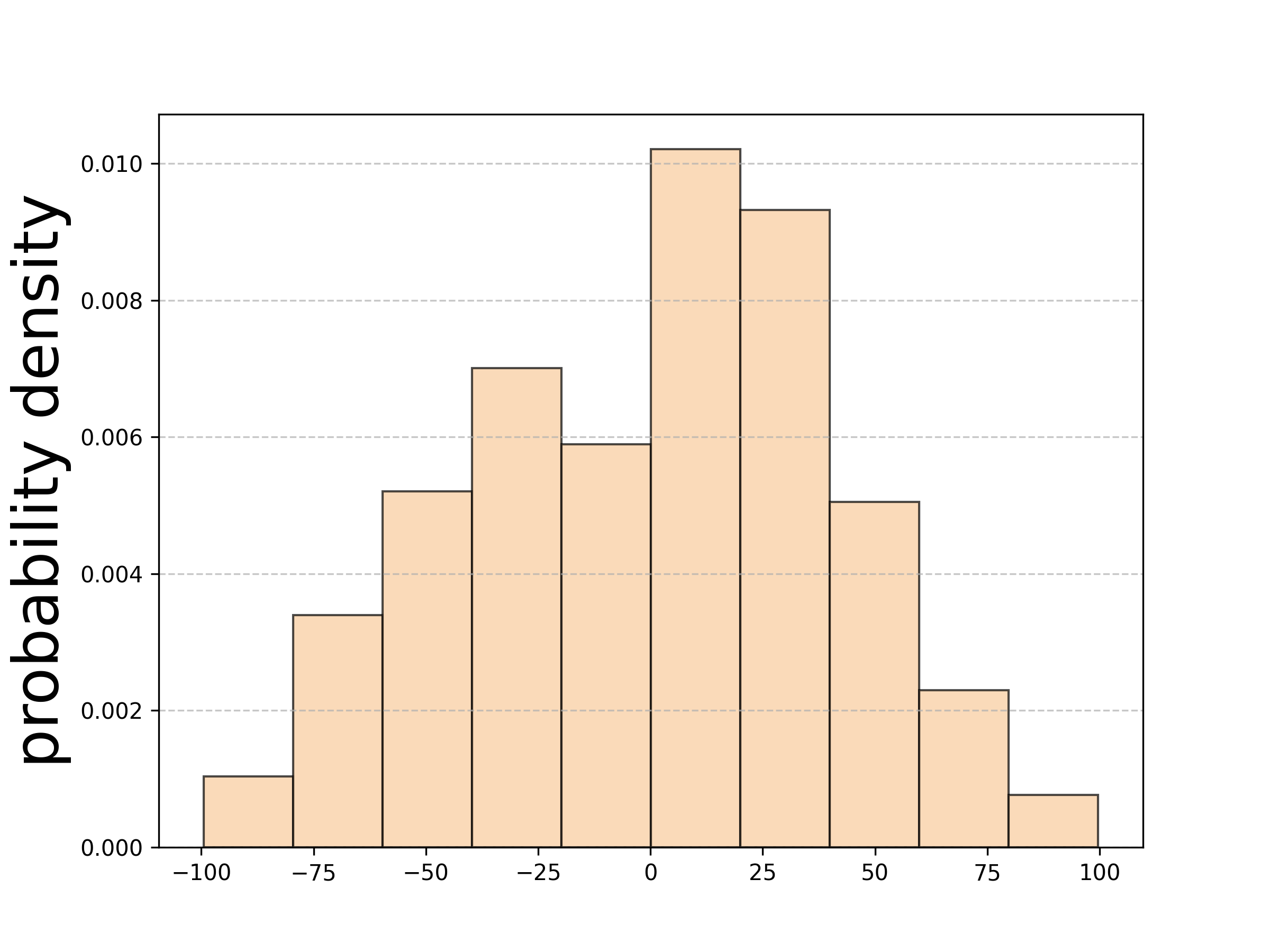}
            \vspace{-20pt}
            \caption[]%
            {{x (meters)}}    
        \end{subfigure}
        \hfill
        \begin{subfigure}[t]{0.23\textwidth}  
            \centering 
            \includegraphics[width=\textwidth]{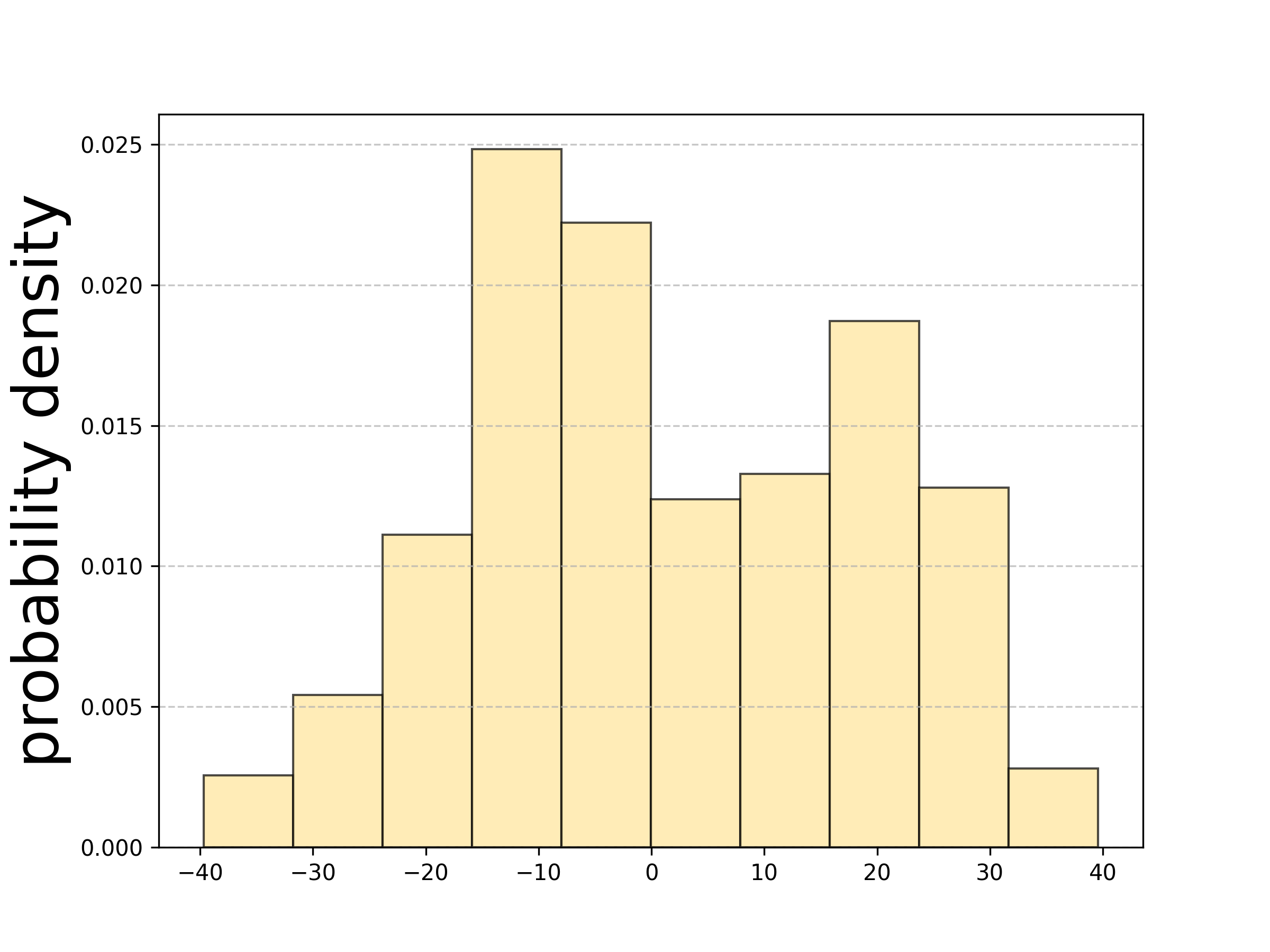}
            \vspace{-20pt}
            \caption[]%
            {{y (meters)}}
        \end{subfigure}

        \begin{subfigure}[t]{0.23\textwidth}
            \centering 
            \includegraphics[width=\textwidth]{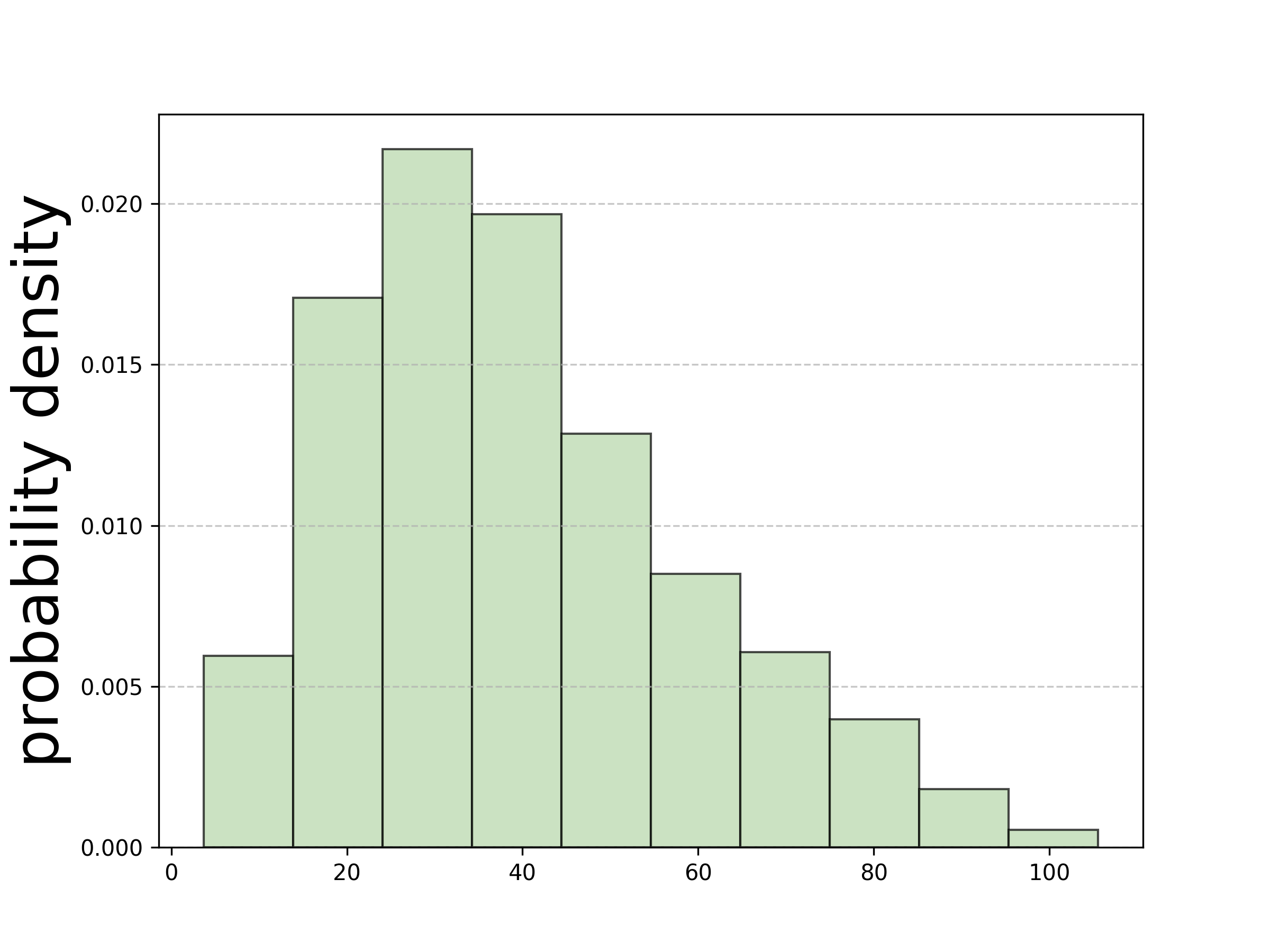}
            \vspace{-20pt}
            \caption[]%
            {{distance (meters)}}
        \end{subfigure}
        \hfill
        \begin{subfigure}[t]{0.23\textwidth}
            \centering 
            \includegraphics[width=\textwidth]{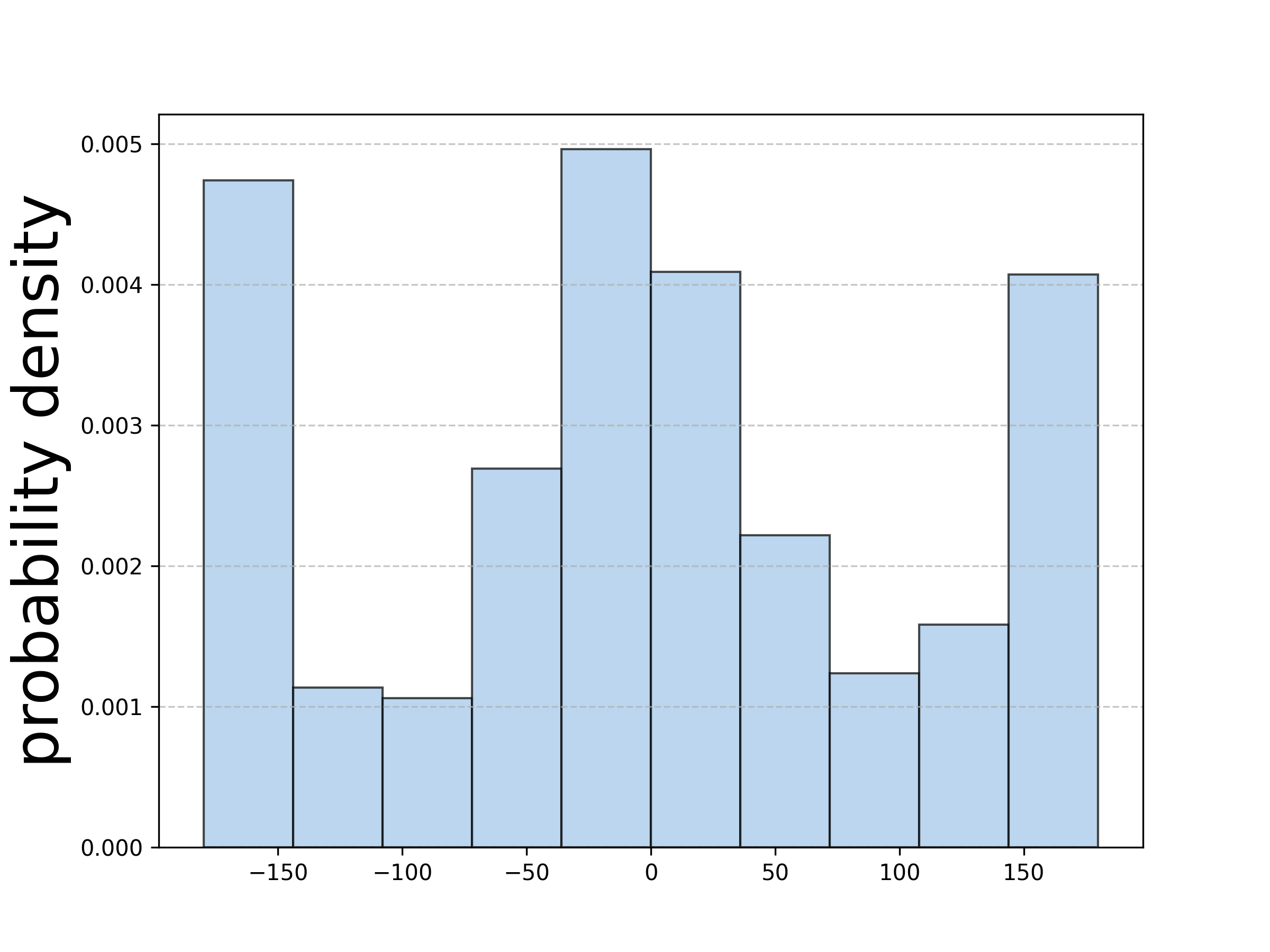}
            \vspace{-20pt}
            \caption[]%
            {{angle (degrees)}}
        \end{subfigure}
        \hfill
        
        \vspace{-5pt}
        \caption[]
        {
        The distribution of ground-truth answer locations relative to CAV in \namedataset's \namexsplit~Q2: Grounding behind a reference object at a location. 
        } 
        \label{fig:stats_v2x_q2}
        \vspace{-10pt}
\end{figure}

\begin{figure}[!t]
        \centering
        \begin{subfigure}[t]{0.23\textwidth}
            \centering 
            \includegraphics[width=\textwidth]{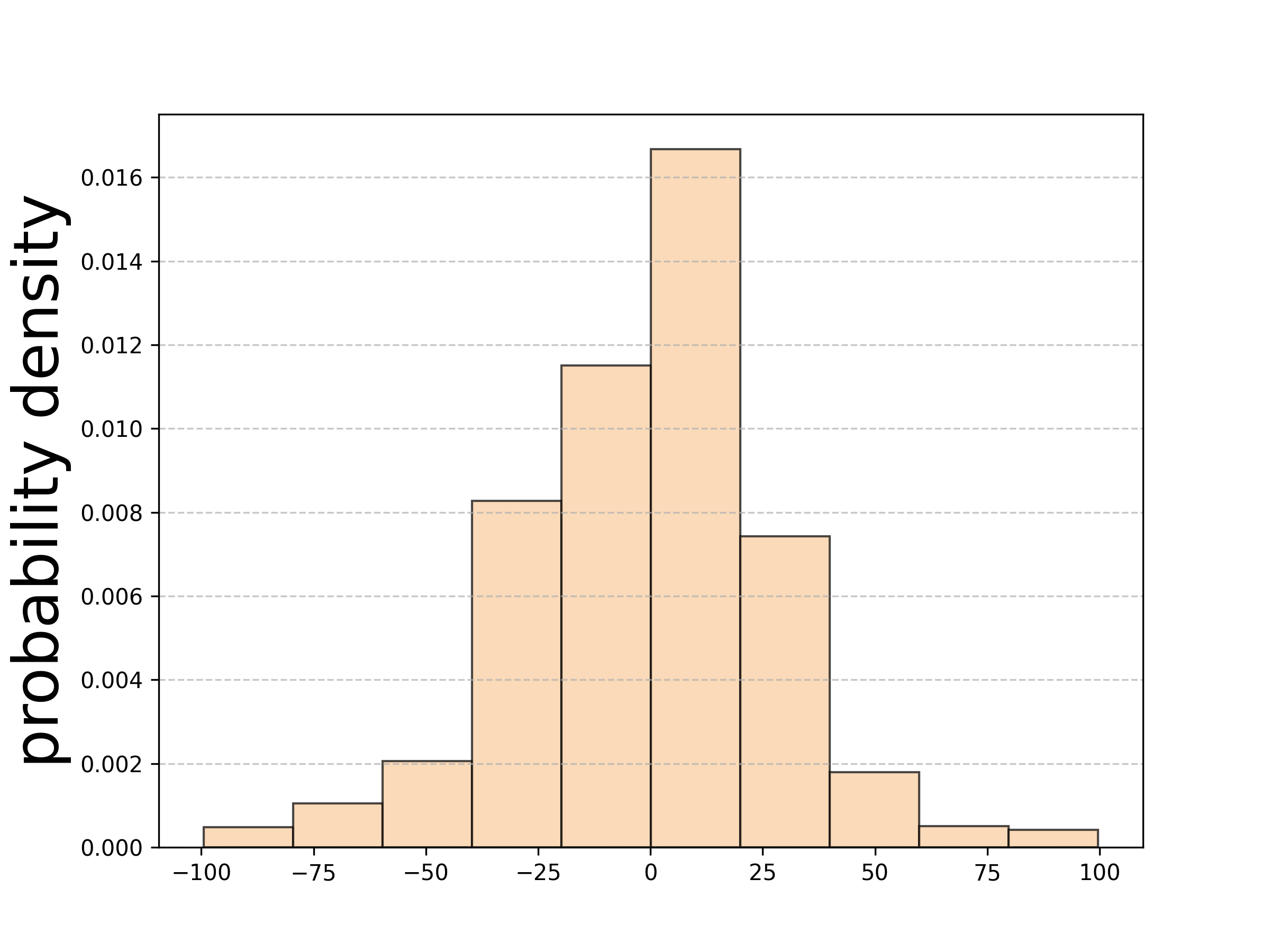}
            \vspace{-20pt}
            \caption[]%
            {{x (meters)}}    
        \end{subfigure}
        \hfill
        \begin{subfigure}[t]{0.23\textwidth}  
            \centering 
            \includegraphics[width=\textwidth]
            {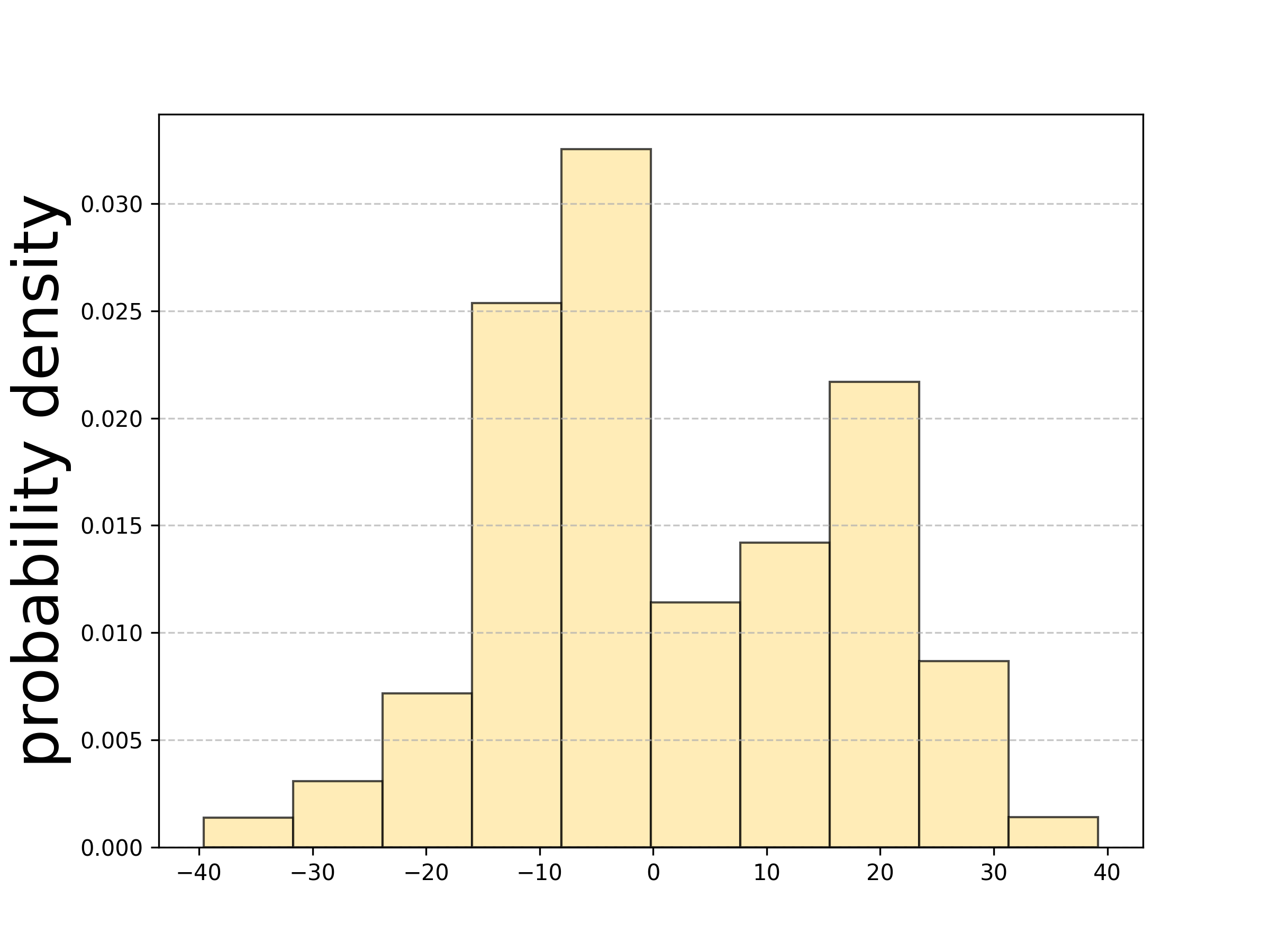}
            \vspace{-20pt}
            \caption[]%
            {{y (meters)}}
        \end{subfigure}

        \begin{subfigure}[t]{0.23\textwidth}
            \centering 
            \includegraphics[width=\textwidth]{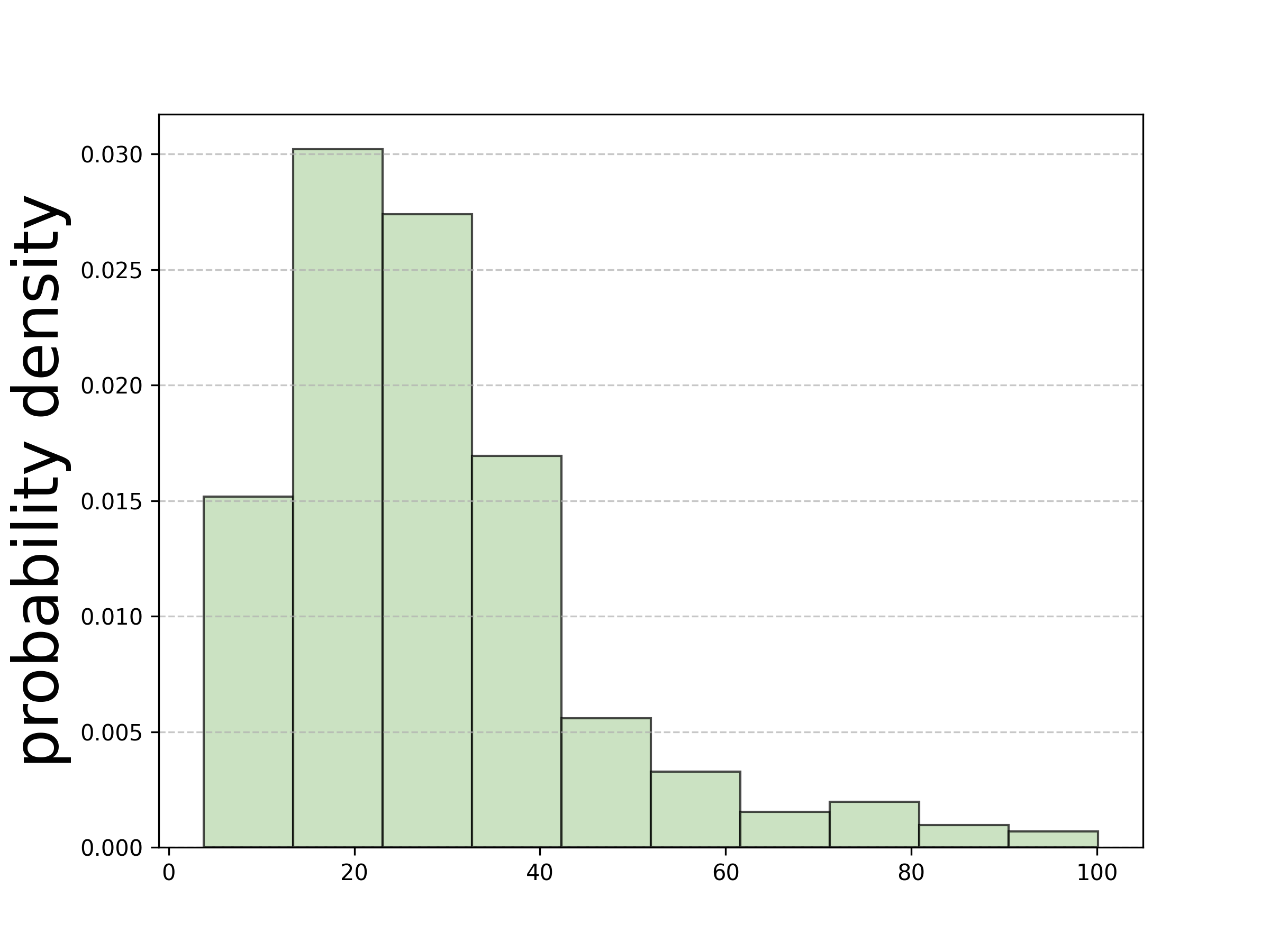}
            \vspace{-20pt}
            \caption[]%
            {{distance (meters)}}
        \end{subfigure}
        \hfill
        \begin{subfigure}[t]{0.23\textwidth}
            \centering 
            \includegraphics[width=\textwidth]{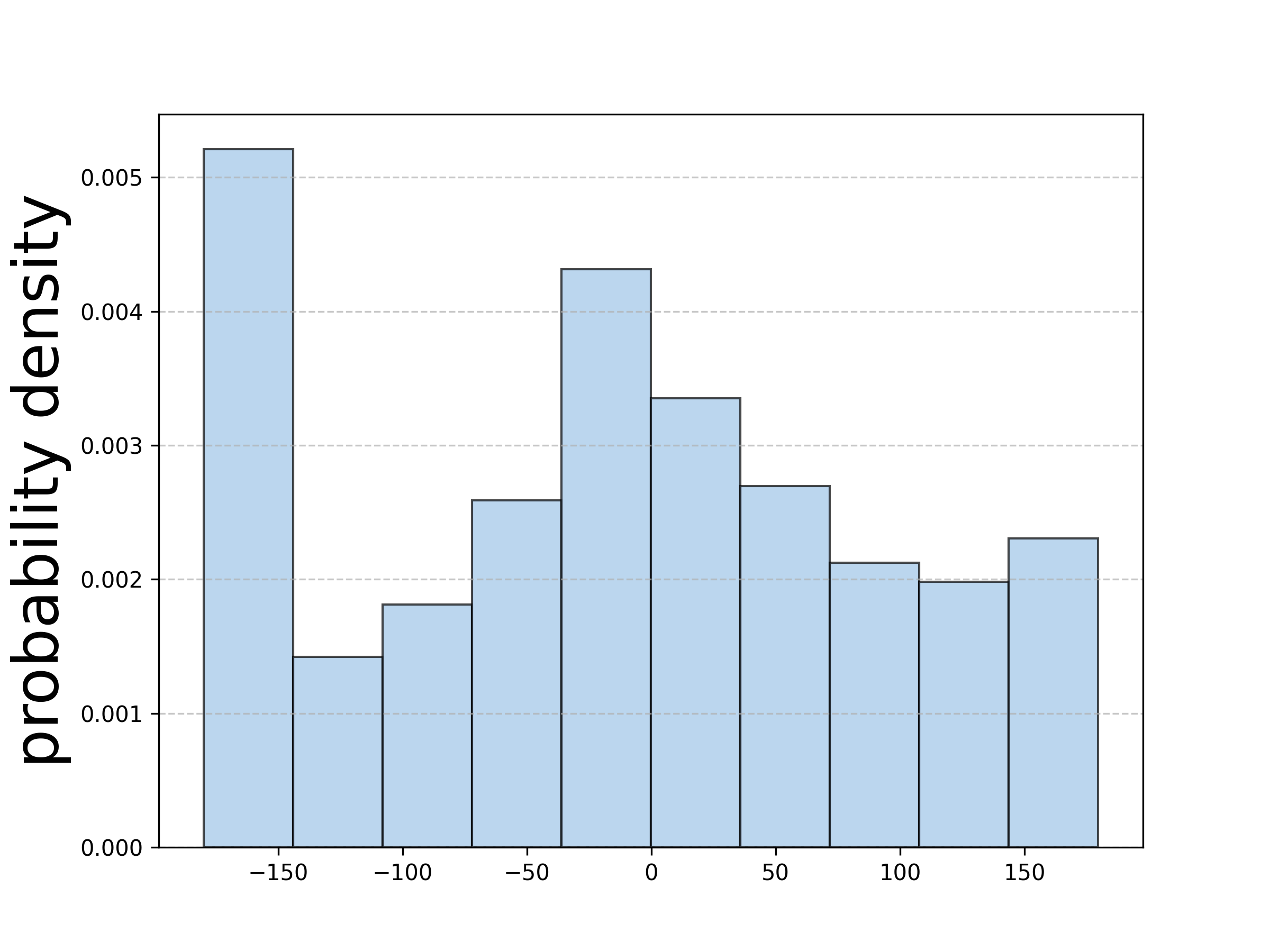}
            \vspace{-20pt}
            \caption[]%
            {{angle (degrees)}}
        \end{subfigure}
        \hfill
        
        \vspace{-5pt}
        \caption[]
        {
        The distribution of ground-truth answer locations relative to CAV in \namedataset's \namexsplit~Q3: Grounding behind a reference object in a direction. 
        } 
        \label{fig:stats_v2x_q3}
        \vspace{-10pt}
\end{figure}

\begin{figure}[!t]
        \centering
        \begin{subfigure}[t]{0.23\textwidth}
            \centering 
            \includegraphics[width=\textwidth]{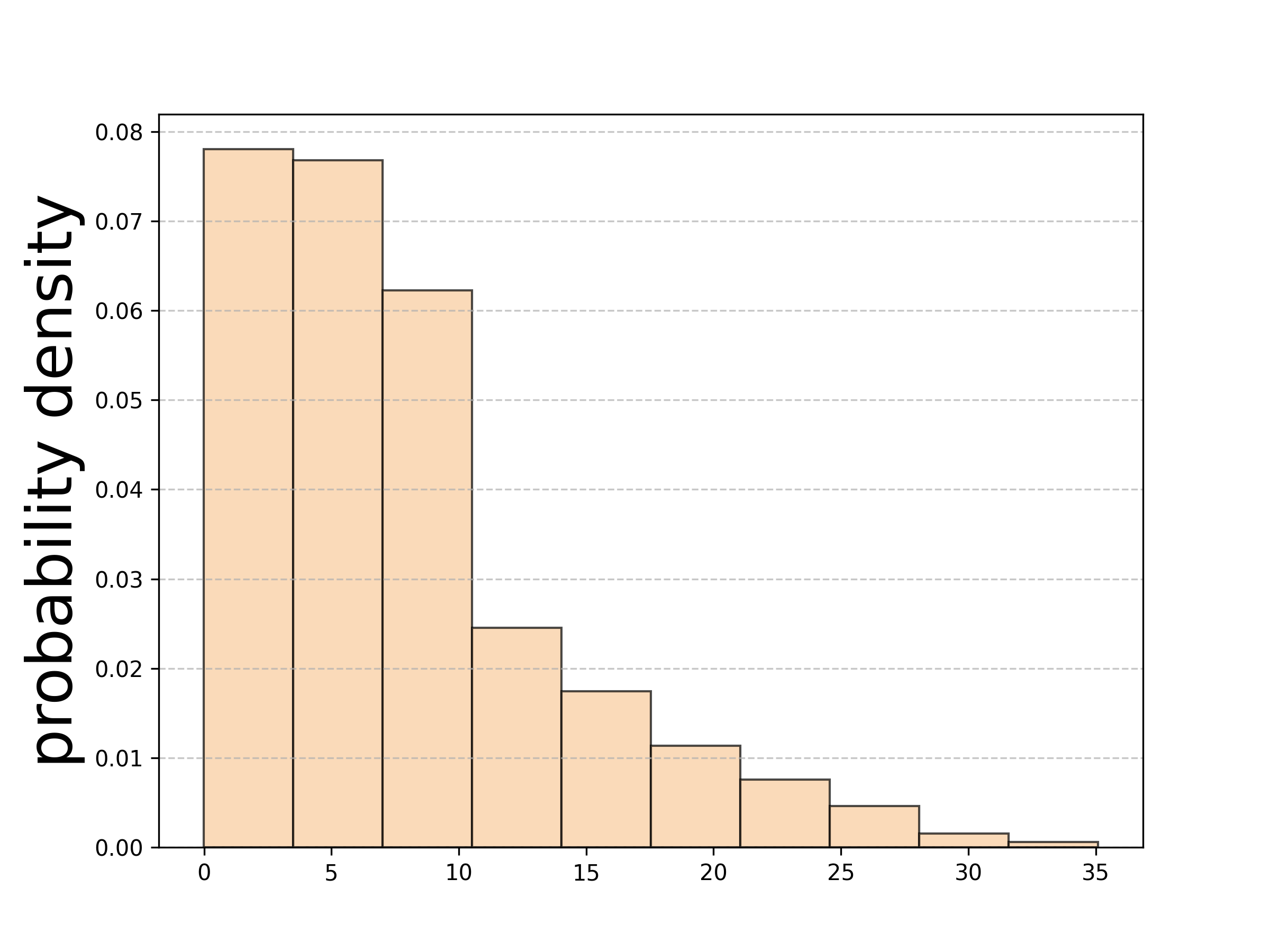}
            \vspace{-20pt}
            \caption[]%
            {{x (meters)}}    
        \end{subfigure}
        \hfill
        \begin{subfigure}[t]{0.23\textwidth}  
            \centering 
            \includegraphics[width=\textwidth]{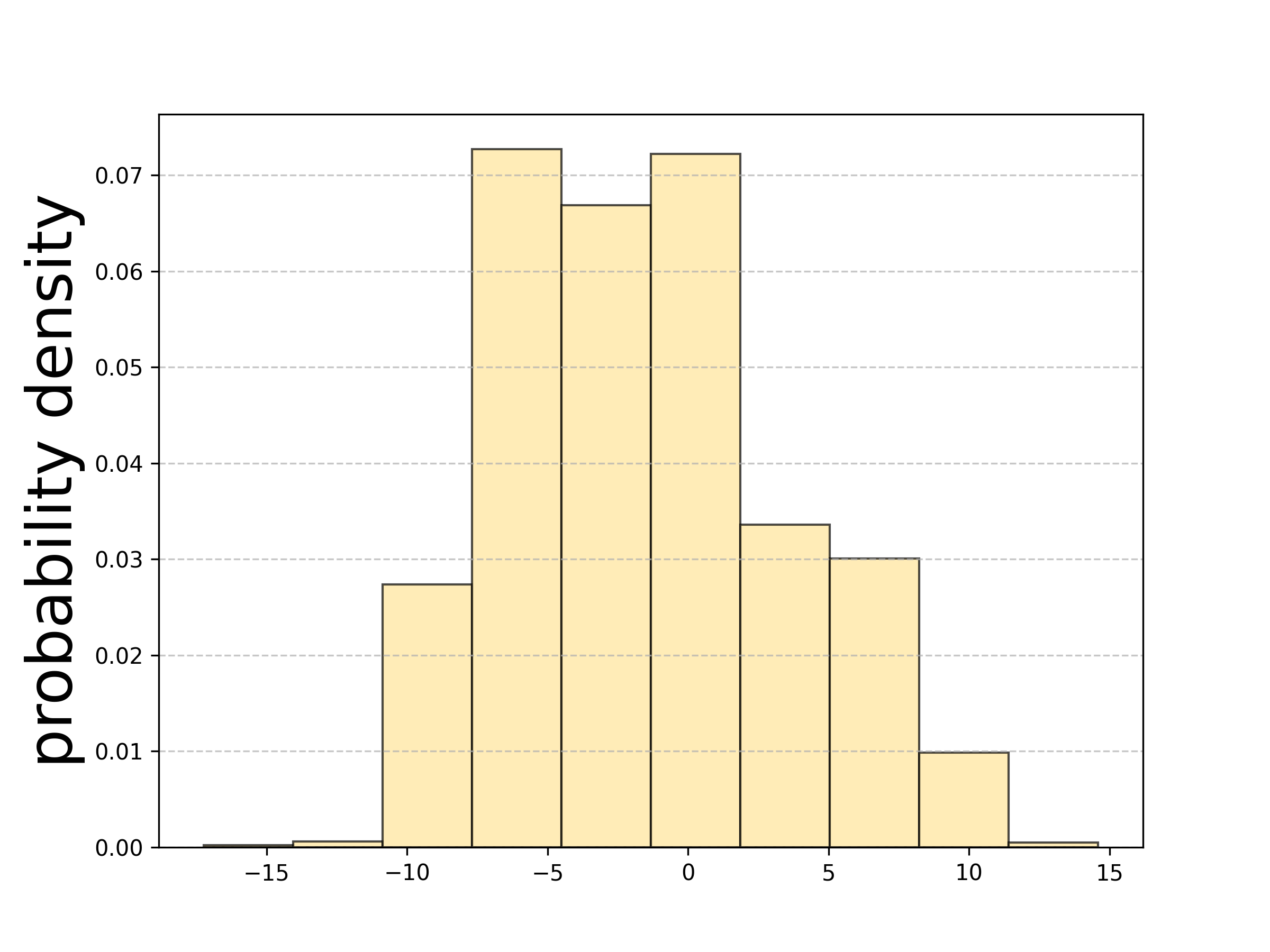}
            \vspace{-20pt}
            \caption[]%
            {{y (meters)}}
        \end{subfigure}

        \begin{subfigure}[t]{0.23\textwidth}
            \centering 
            \includegraphics[width=\textwidth]{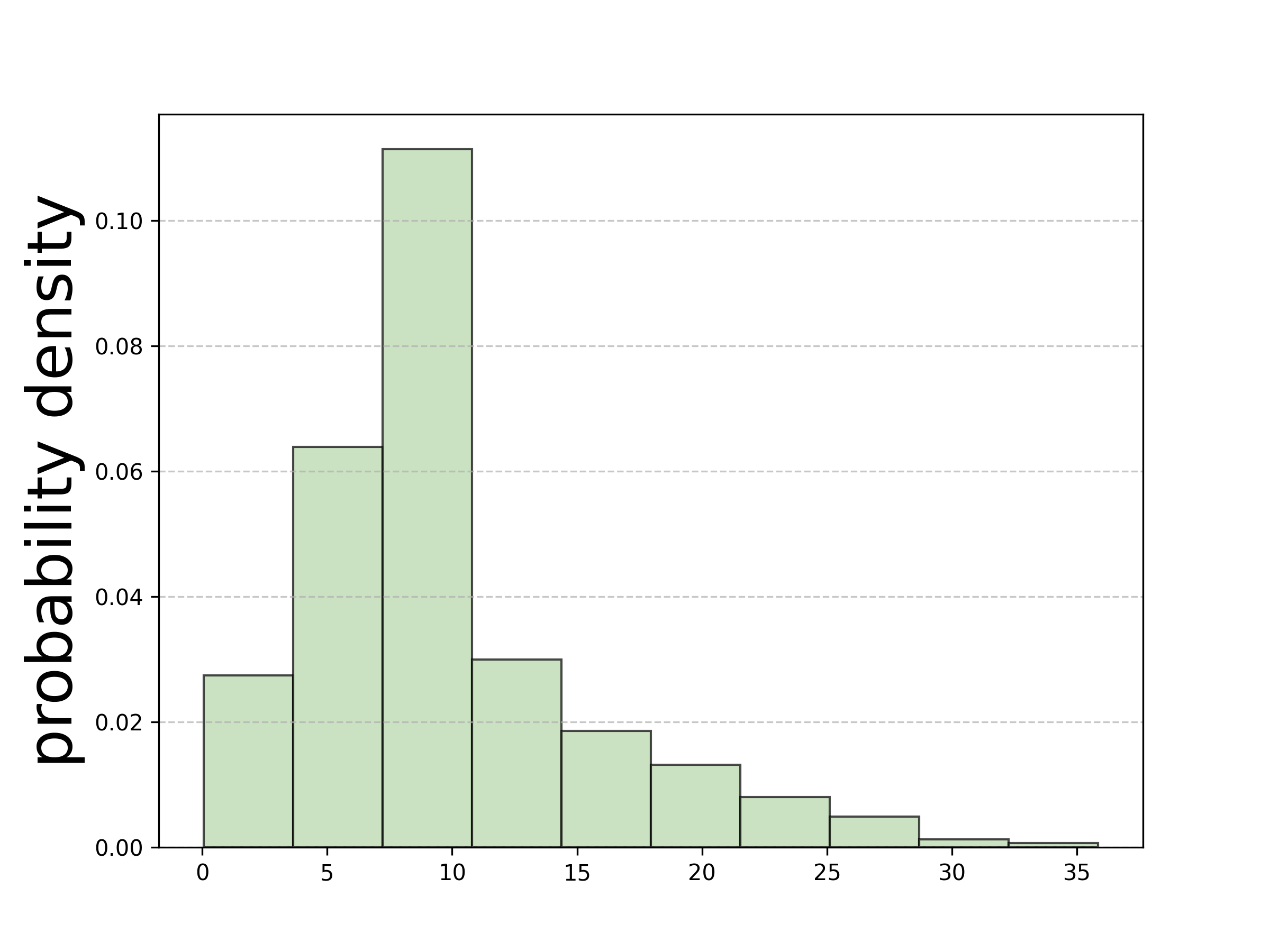}
            \vspace{-20pt}
            \caption[]%
            {{distance (meters)}}
        \end{subfigure}
        \hfill
        \begin{subfigure}[t]{0.23\textwidth}
            \centering 
            \includegraphics[width=\textwidth]{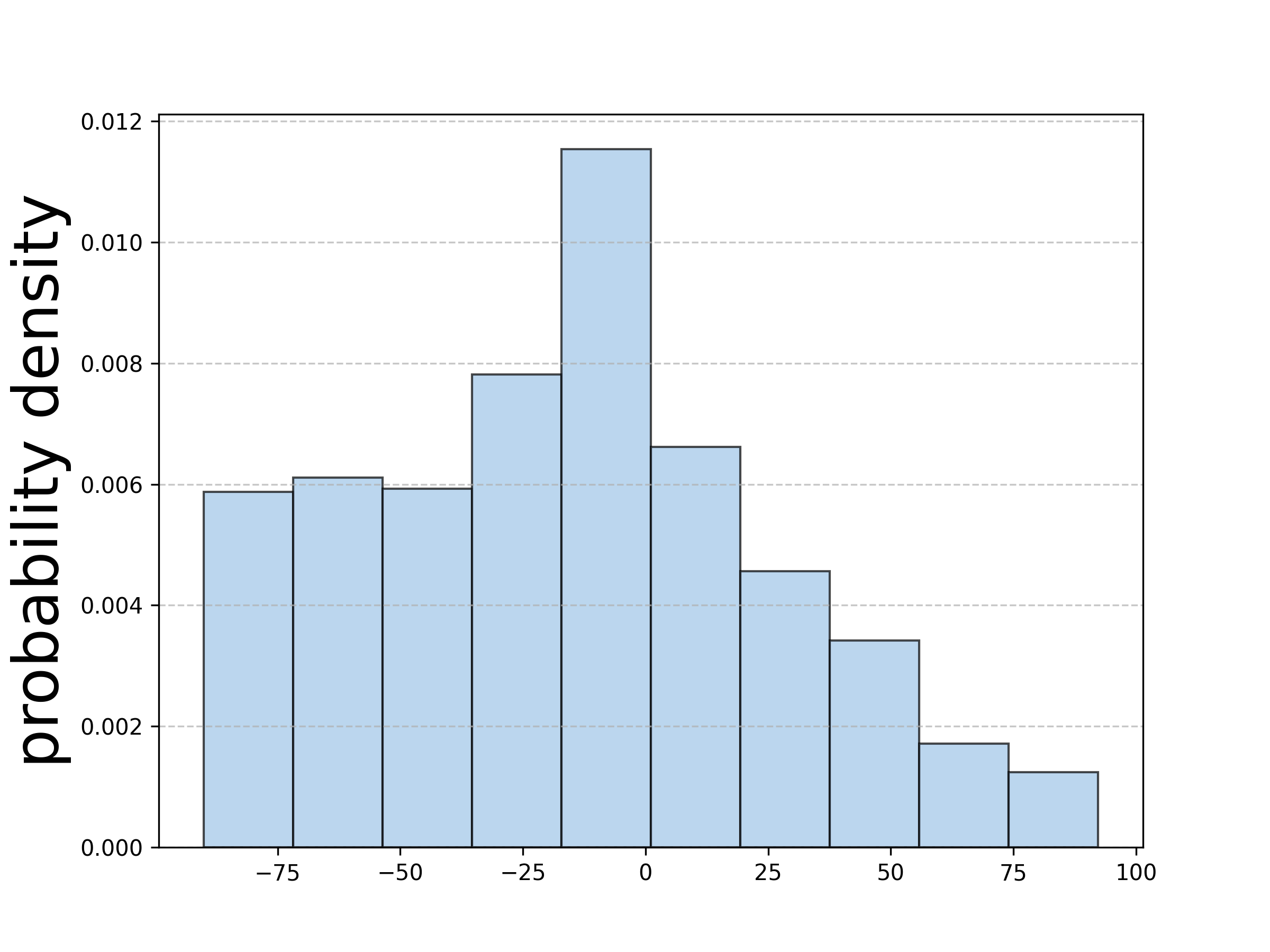}
            \vspace{-20pt}
            \caption[]%
            {{angle (degrees)}}
        \end{subfigure}
        \hfill
        
        \vspace{-5pt}
        \caption[]
        {
        The distribution of ground-truth answer locations relative to CAV in \namedataset's \namexsplit~Q4: Notable object identification. 
        } 
        \label{fig:stats_v2x_q4}
        \vspace{-10pt}
\end{figure}

\begin{figure}[!t]
        \centering
        \begin{subfigure}[t]{0.23\textwidth}
            \centering 
            \includegraphics[width=\textwidth]{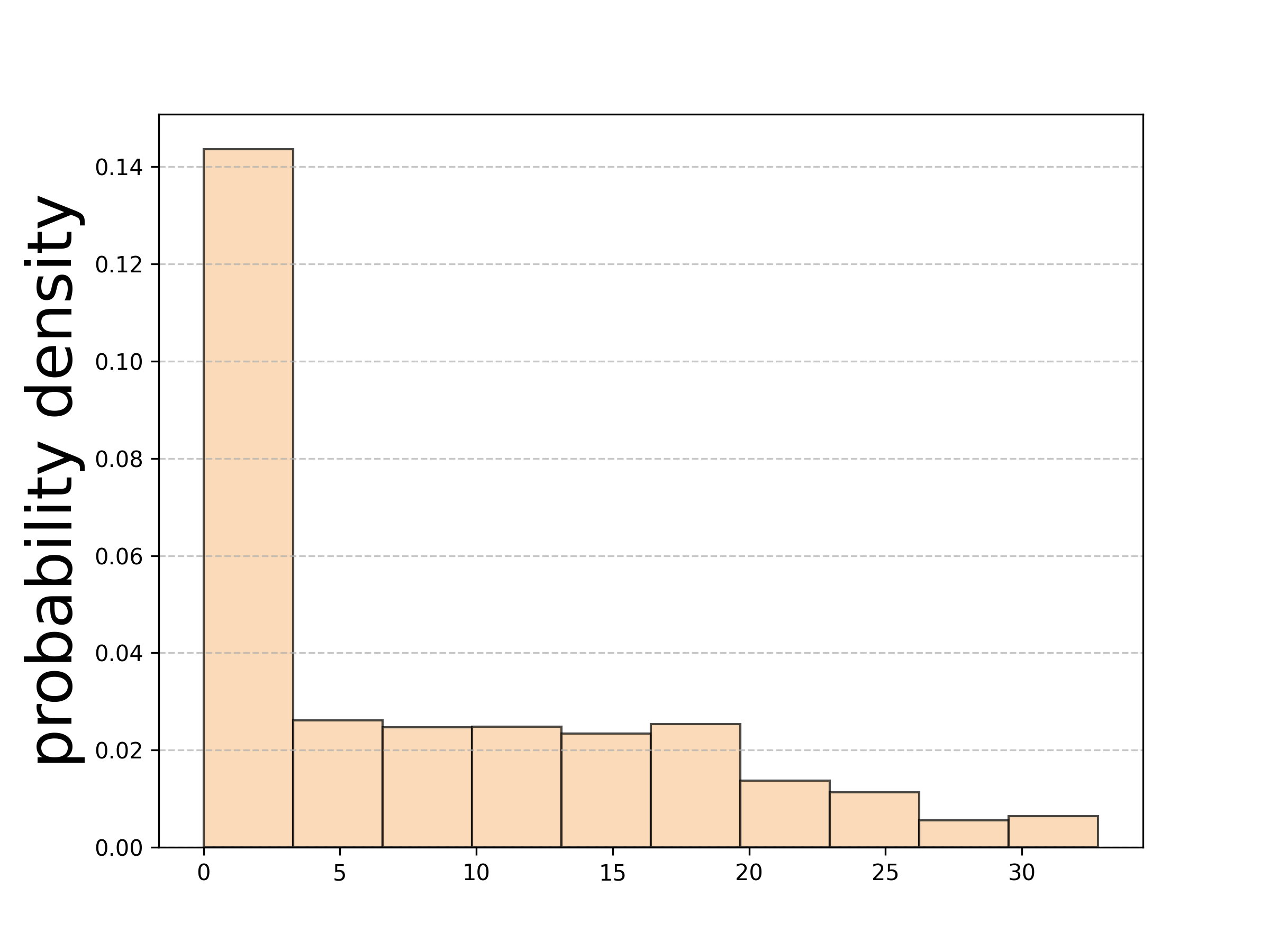}
            \vspace{-20pt}
            \caption[]%
            {{x (meters)}}    
        \end{subfigure}
        \hfill
        \begin{subfigure}[t]{0.23\textwidth}  
            \centering 
            \includegraphics[width=\textwidth]{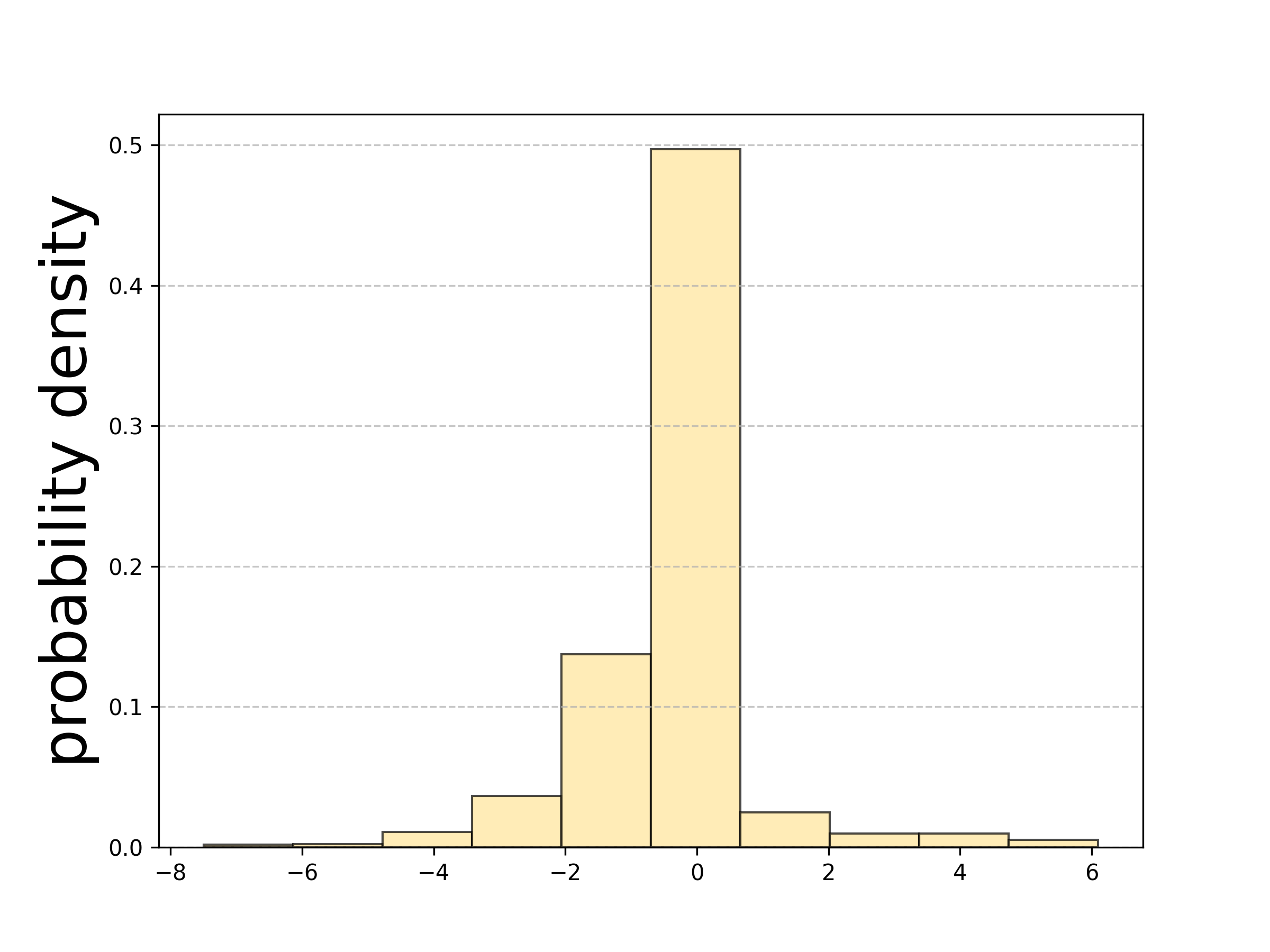}
            \vspace{-20pt}
            \caption[]%
            {{y (meters)}}
        \end{subfigure}

        \begin{subfigure}[t]{0.23\textwidth}
            \centering 
            \includegraphics[width=\textwidth]{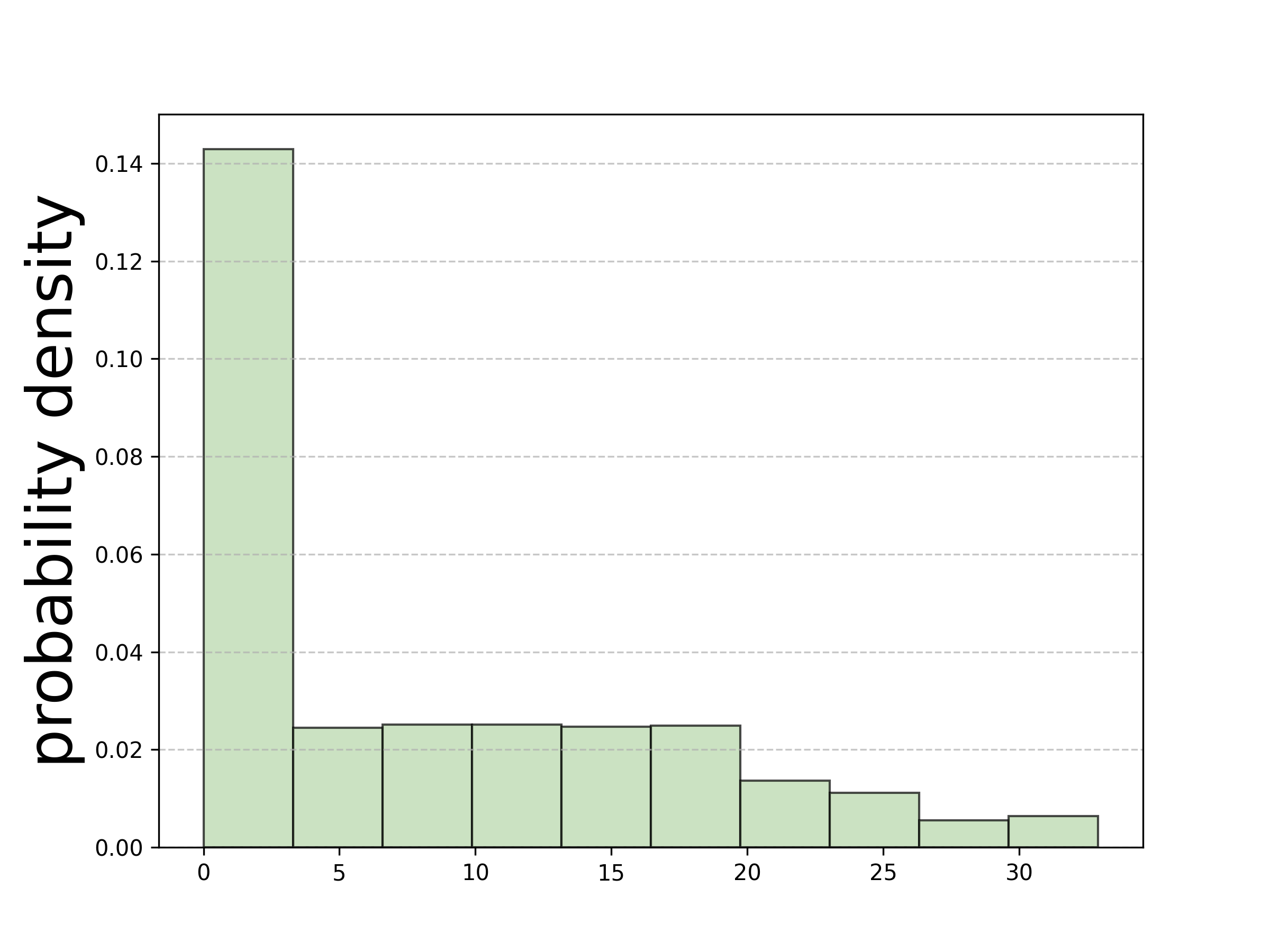}
            \vspace{-20pt}
            \caption[]%
            {{distance (meters)}}
        \end{subfigure}
        \hfill
        \begin{subfigure}[t]{0.23\textwidth}
            \centering 
            \includegraphics[width=\textwidth]{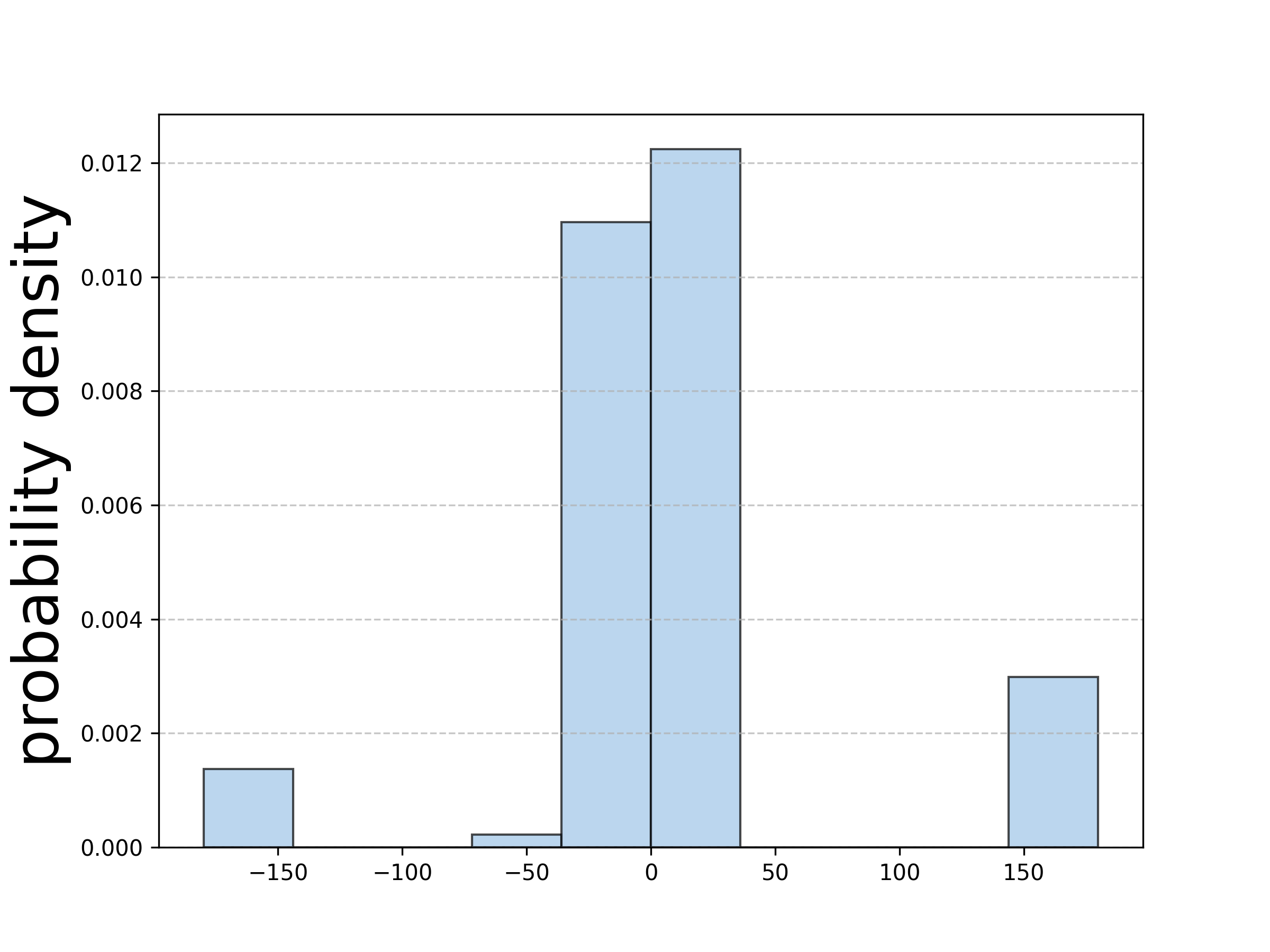}
            \vspace{-20pt}
            \caption[]%
            {{angle (degrees)}}
        \end{subfigure}
        \hfill
        
        \vspace{-5pt}
        \caption[]
        {
        The distribution of ground-truth answer locations relative to CAV in \namedataset's \namexsplit~Q5: Planning. 
        } 
        \label{fig:stats_v2x_q5}
        \vspace{-10pt}
\end{figure}

We also visualize our \namevsplit~distribution of ground truth answer locations relative to the asking CAV for the grounding questions (Q1, Q2, Q3) and the notable object identification question (Q4), as shown in Figures \ref{fig:stats_v2v_q1}, \ref{fig:stats_v2v_q2}, \ref{fig:stats_v2v_q3}, and \ref{fig:stats_v2v_q4}. In our coordinate system, $x$ is the CAV's front direction, and $y$ is the CAV's right direction. For the planning question (Q5), we show the distribution of the ending waypoints in the ground truth answer future trajectories, as shown in \ref{fig:stats_v2v_q5}. We visualize the location distribution of \namedataset's \namexsplit in Figures \ref{fig:stats_v2x_q1}, \ref{fig:stats_v2x_q2}, \ref{fig:stats_v2x_q3}, \ref{fig:stats_v2x_q4}, and \ref{fig:stats_v2x_q5}. These figures indicate that our \namedataset~has diverse spatial distributions in the driving scenes. Compared to NuScenes~\cite{caesar2019nuscenes},
our ~\namedataset~has larger ranges and standard deviations of the ground-truth ending waypoints, as shown in Table \ref{tab:stats_range_std_q5}. Therefore, the planning task in our ~\namedataset~is more challenging.

\begin{table}[!th]
\caption{
Ranges and standard deviations of ground-truth ending waypoints.
\vspace{-10pt}
}
\small
\begin{center}
{
\begin{tabular}{l ccc ccc }
  \hline
  \hline
  \multirow{2}{*}{Dataset} &
  \multicolumn{3}{c}{x: forward} & \multicolumn{3}{c}{y: right} \\
  \cmidrule(lr){2-4} \cmidrule(lr){5-7}
  &
  min & max & std & 
  min & max & std \\
  \hline
  \hline
  NuScenes & -0.9 & 39.7 & 10.4 & -11.0 & 11.1 & 1.9 \\
  \namedataset~(ours) & -2.1 & 177.0 & 28.1 & -24.3 & 12.0 & 2.4 \\
  \hline
\end{tabular}
}
\label{tab:stats_range_std_q5}
\end{center}
\vspace{-15pt}
\end{table}

\section{Additional Qualitative Results}

\begin{figure*}[!t]
\centering
\includegraphics[width=1\textwidth]{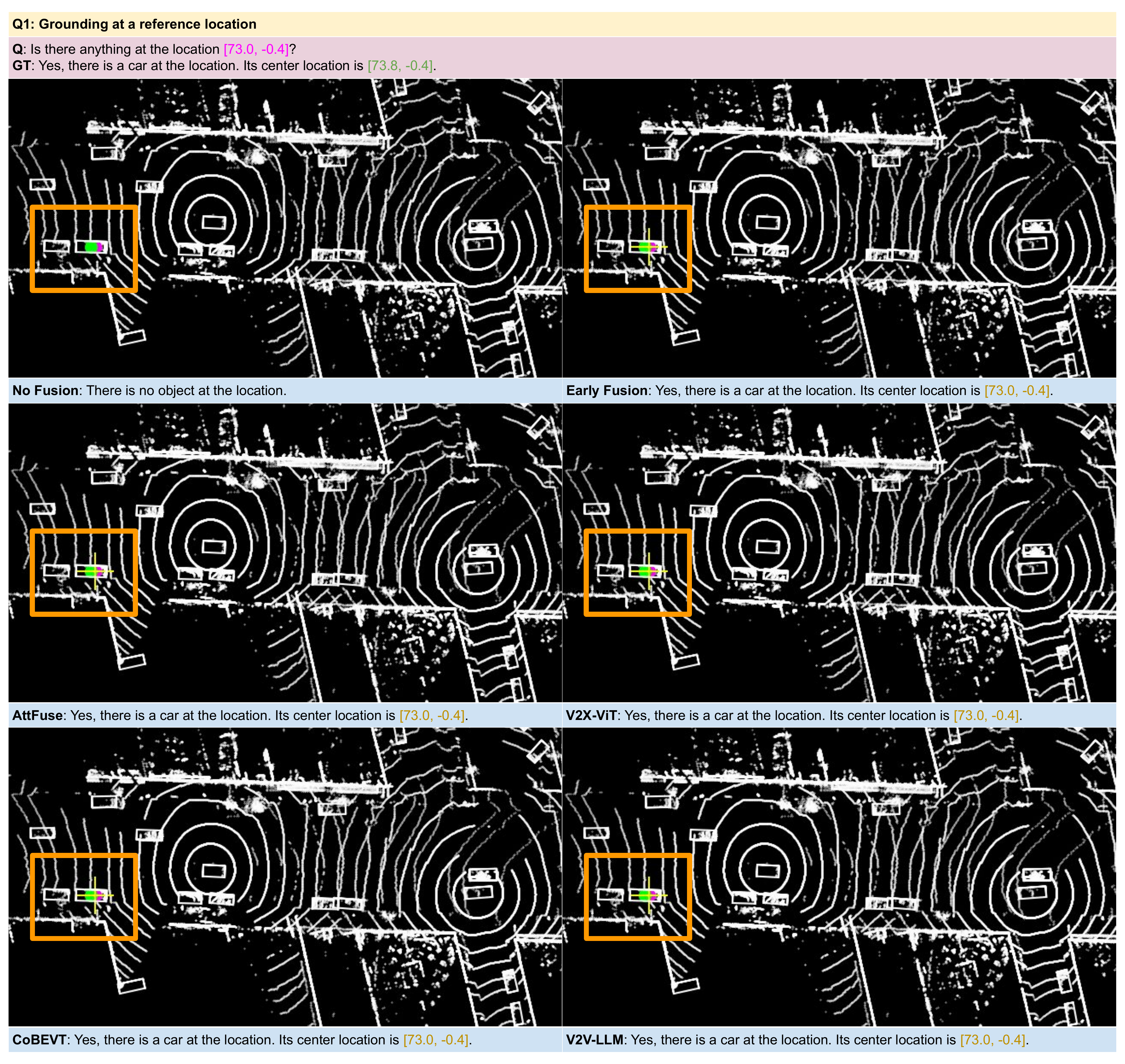}
\caption[]
        {\namemethod~and baseline methods' \textit{grounding} results on \namedataset's \namevsplit~testing set.~\textcolor{magenta}{Magenta $\circ$}: reference locations in questions. \textcolor{olive}{Yellow $+$}: model output locations. \textcolor{Green}{Green $\circ$}: ground-truth answers.} 
        \label{fig:supp_q1}
        \vspace{-5pt}
\end{figure*}
\begin{figure*}[!t]
\centering
\includegraphics[width=1\textwidth]{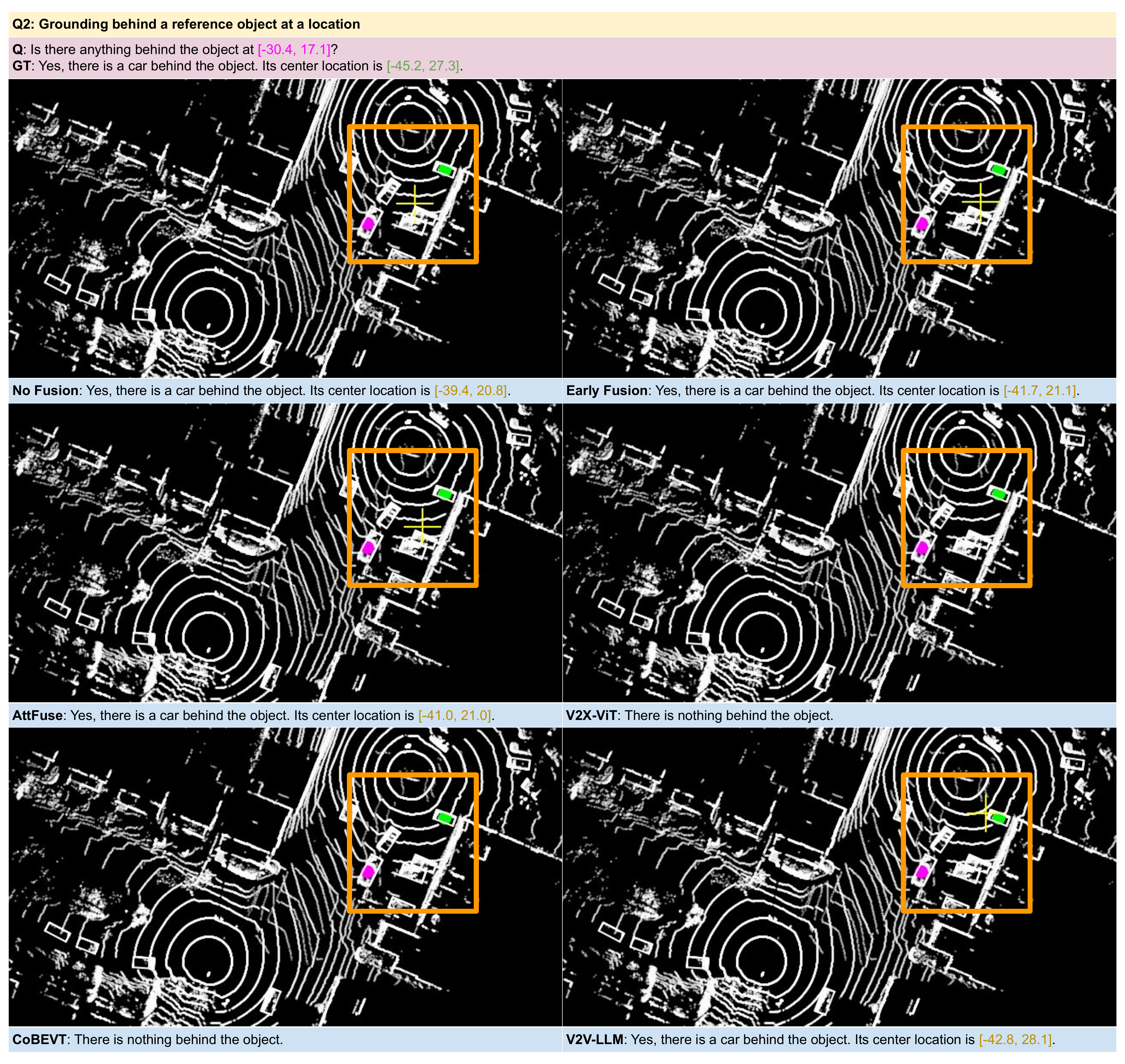}
\caption[]
        {\namemethod~and baseline methods' \textit{grounding} results on \namedataset's \namevsplit~testing set.~\textcolor{magenta}{Magenta $\circ$}: reference locations in questions. \textcolor{olive}{Yellow $+$}: model output locations. \textcolor{Green}{Green $\circ$}: ground-truth answers.} 
        \label{fig:supp_q2}
        \vspace{-5pt}
\end{figure*}
\begin{figure*}[!t]
\centering
\includegraphics[width=1\textwidth]{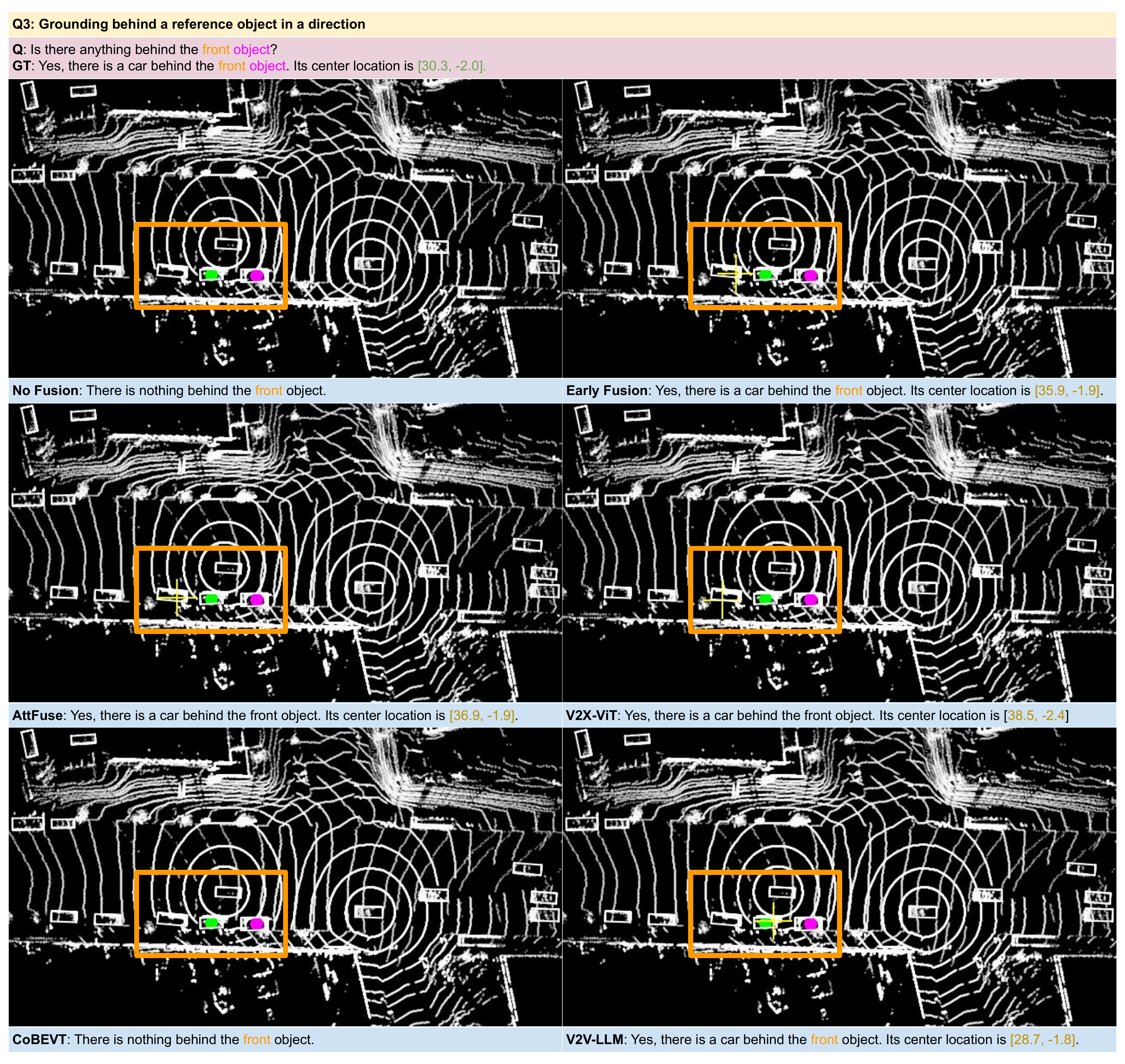}
\caption[]
        {\namemethod~and baseline methods' \textit{grounding} results on \namedataset's \namevsplit~testing set.~\textcolor{magenta}{Magenta $\circ$}: reference locations in questions. \textcolor{olive}{Yellow $+$}: model output locations. \textcolor{Green}{Green $\circ$}: ground-truth answers.} 
        \label{fig:supp_q3_1}
        \vspace{-5pt}
\end{figure*}
\begin{figure*}[!t]
\centering
\includegraphics[width=1\textwidth]{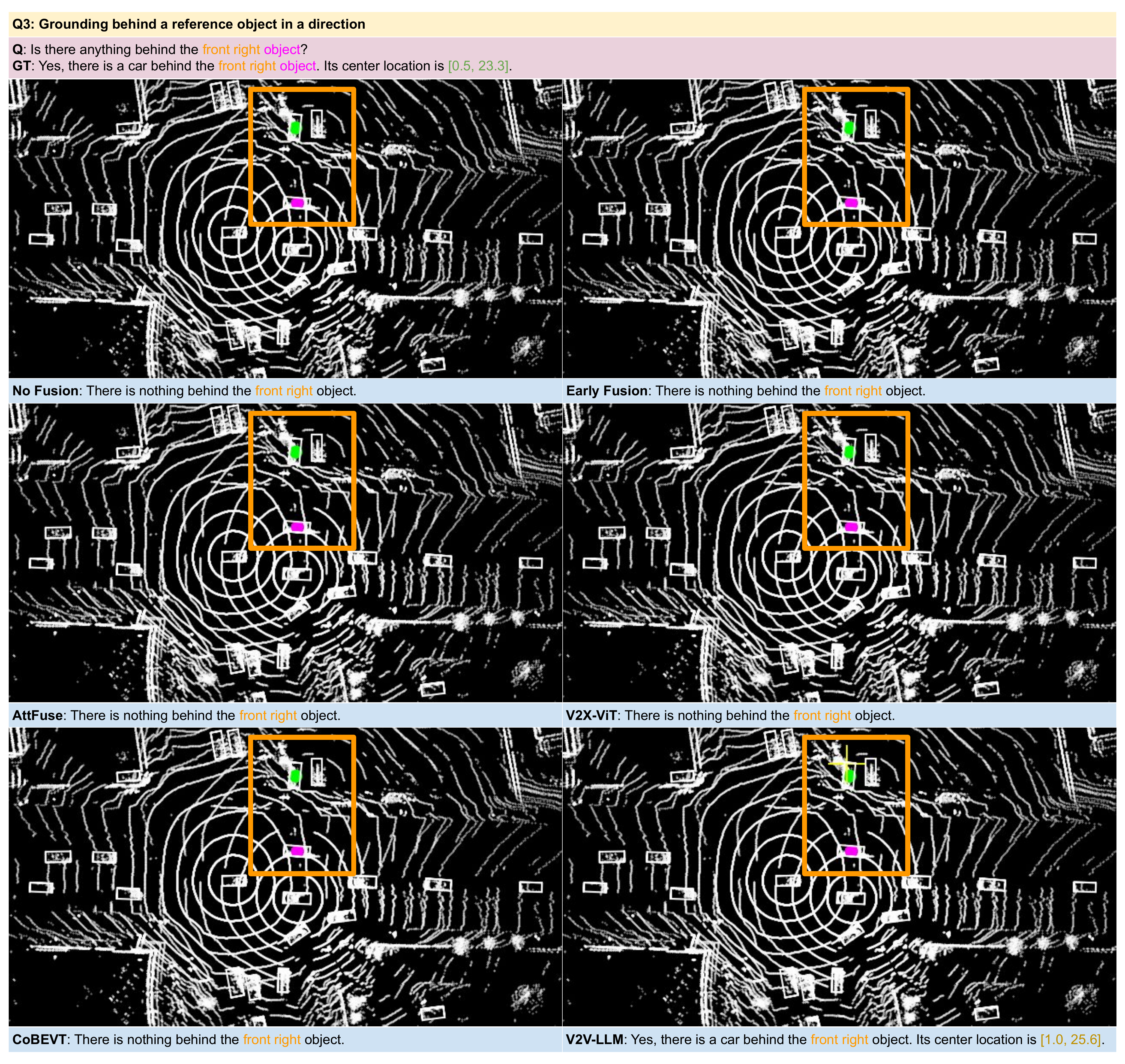}
\caption[]
        {\namemethod~and baseline methods' \textit{grounding} results on \namedataset's \namevsplit~testing set.~\textcolor{magenta}{Magenta $\circ$}: reference locations in questions. \textcolor{olive}{Yellow $+$}: model output locations. \textcolor{Green}{Green $\circ$}: ground-truth answers.} 
        \label{fig:supp_q3_2}
        \vspace{-5pt}
\end{figure*}
\begin{figure*}[!t]
\centering
\includegraphics[width=1\textwidth]{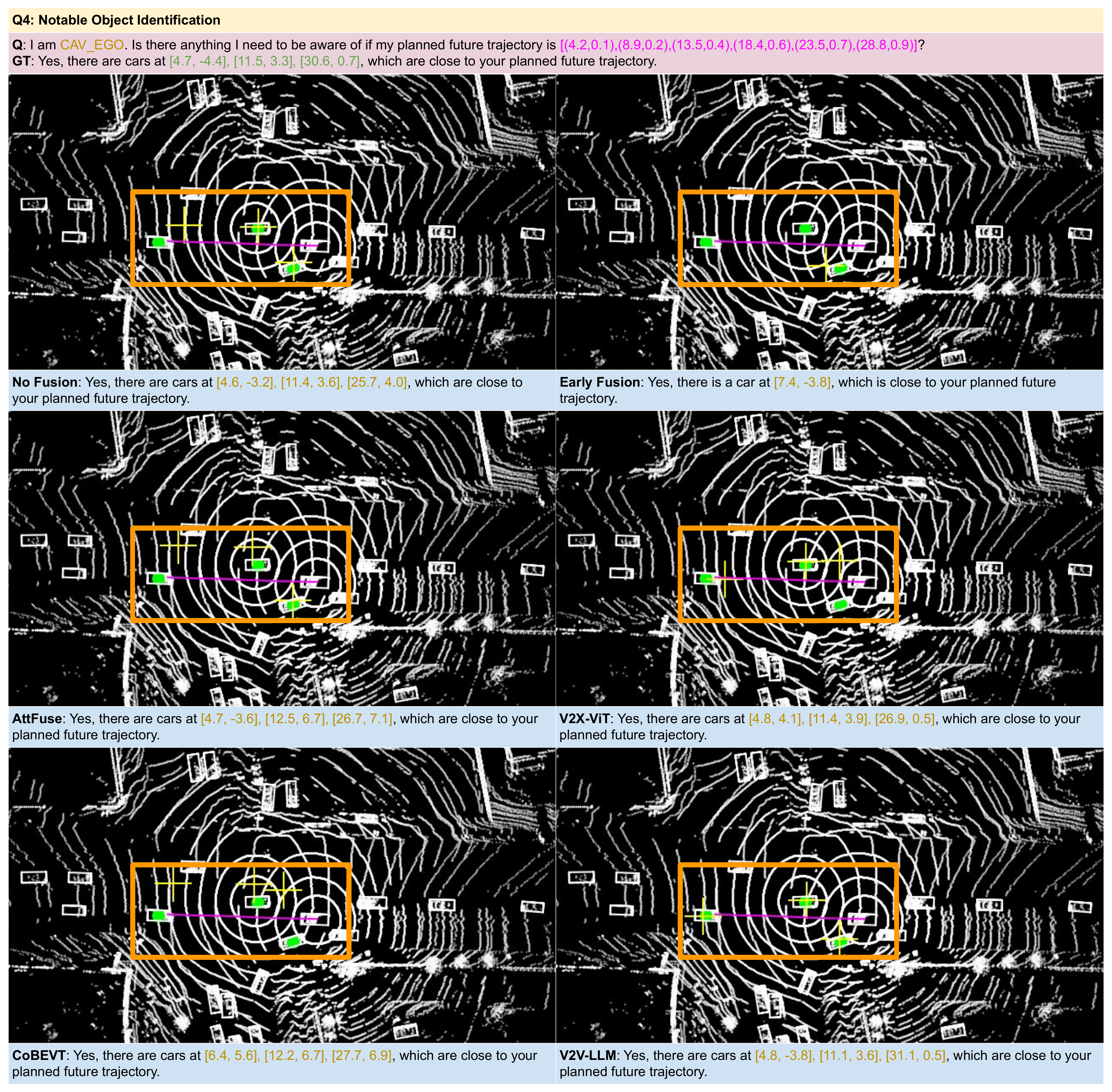}
\caption[]
        {\namemethod~and baseline methods' \textit{notable object identification} results on \namedataset's \namevsplit~testing set.~\textcolor{magenta}{Magenta curve}: planned future trajectories in questions. \textcolor{Green}{Green $\circ$}: ground-truth notable object locations. \textcolor{olive}{Yellow $+$}: model identification outputs.} 
        \label{fig:supp_q4_1}
        \vspace{-5pt}
\end{figure*}
\begin{figure*}[!t]
\centering
\includegraphics[width=1\textwidth]{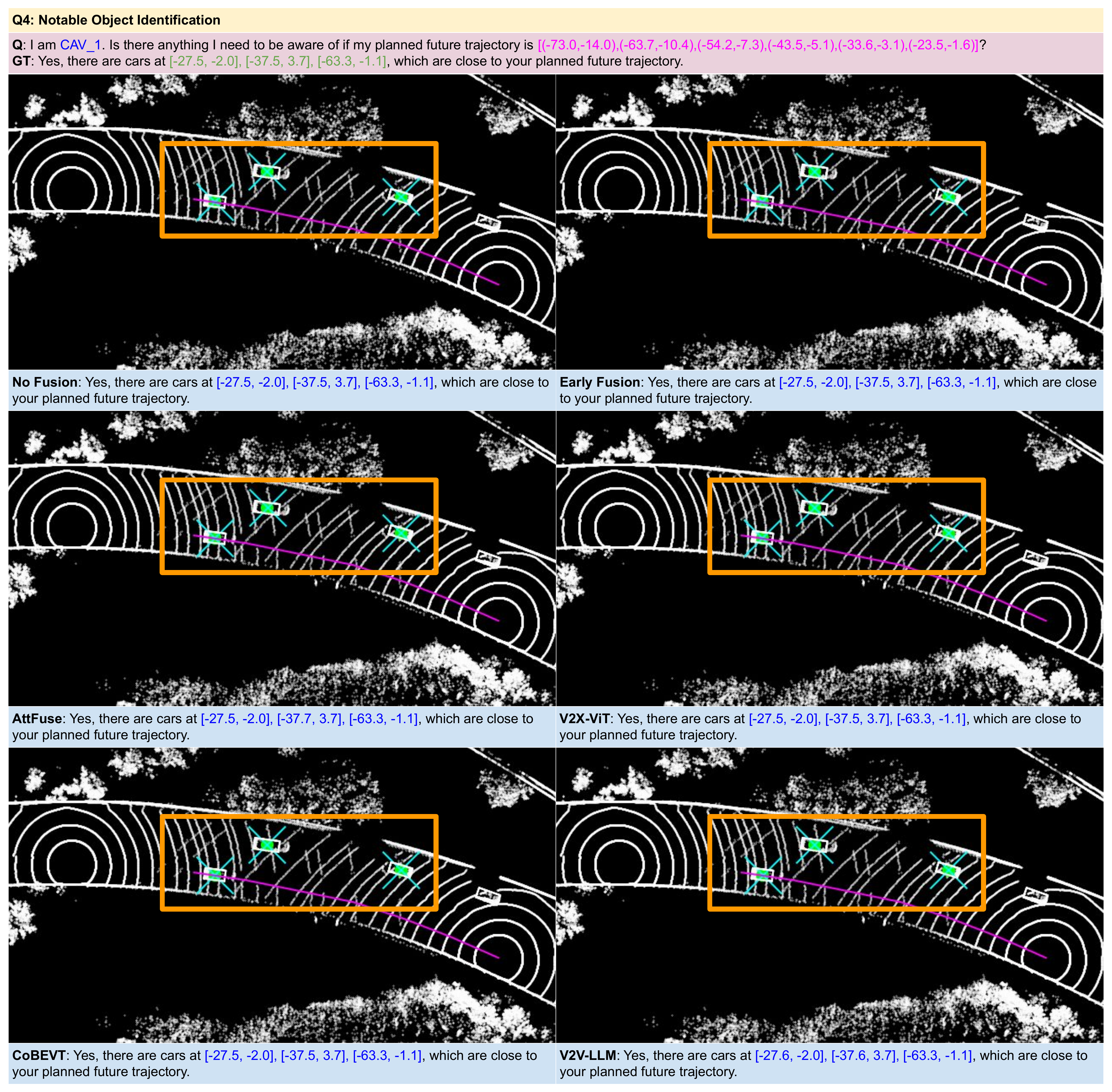}
\caption[]
        {\namemethod~and baseline methods' \textit{notable object identification} results on \namedataset's \namevsplit~testing set.~\textcolor{magenta}{Magenta curve}: planned future trajectories in questions. \textcolor{Green}{Green $\circ$}: ground-truth notable object locations. \textcolor{cyan}{Cyan $\times$}: model identification outputs.} 
        \label{fig:supp_q4_2}
        \vspace{-5pt}
\end{figure*}
\begin{figure*}[!t]
\centering
\includegraphics[width=1\textwidth]{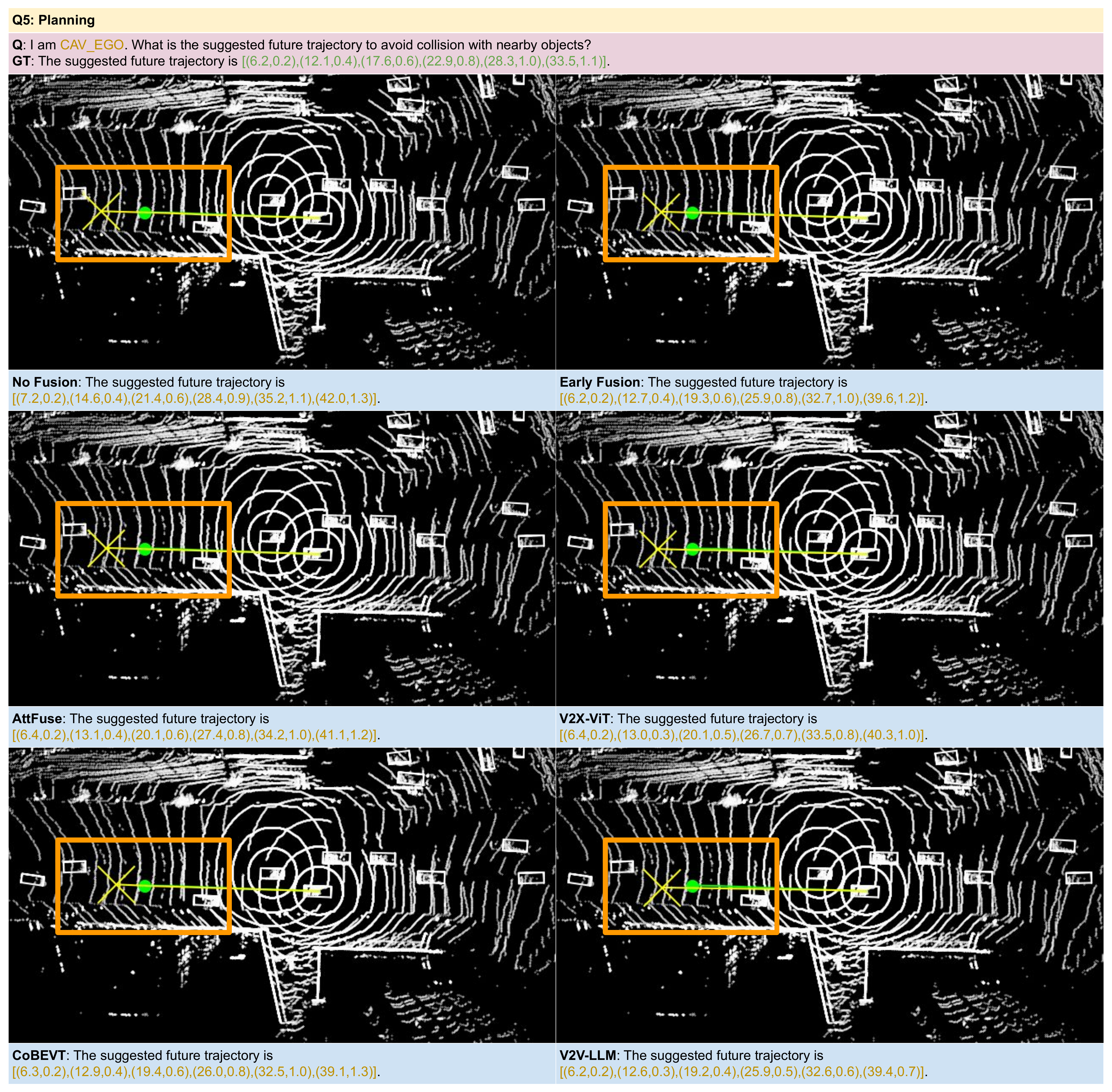}
\caption[]
        {\namemethod~and baseline methods' \textit{planning} results on \namedataset's \namevsplit~testing set.
         \textcolor{Green}{Green curve}: future trajectories in ground-truth answers. \textcolor{Green}{Green $\circ$}: ending waypoints in ground-truth answers. \textcolor{olive}{Yellow curve}: model planning outputs. \textcolor{olive}{Yellow $\times$}: ending waypoints in model outputs.} 
        \label{fig:supp_q5_1}
        \vspace{-5pt}
\end{figure*}
\begin{figure*}[!t]
\centering
\includegraphics[width=1\textwidth]{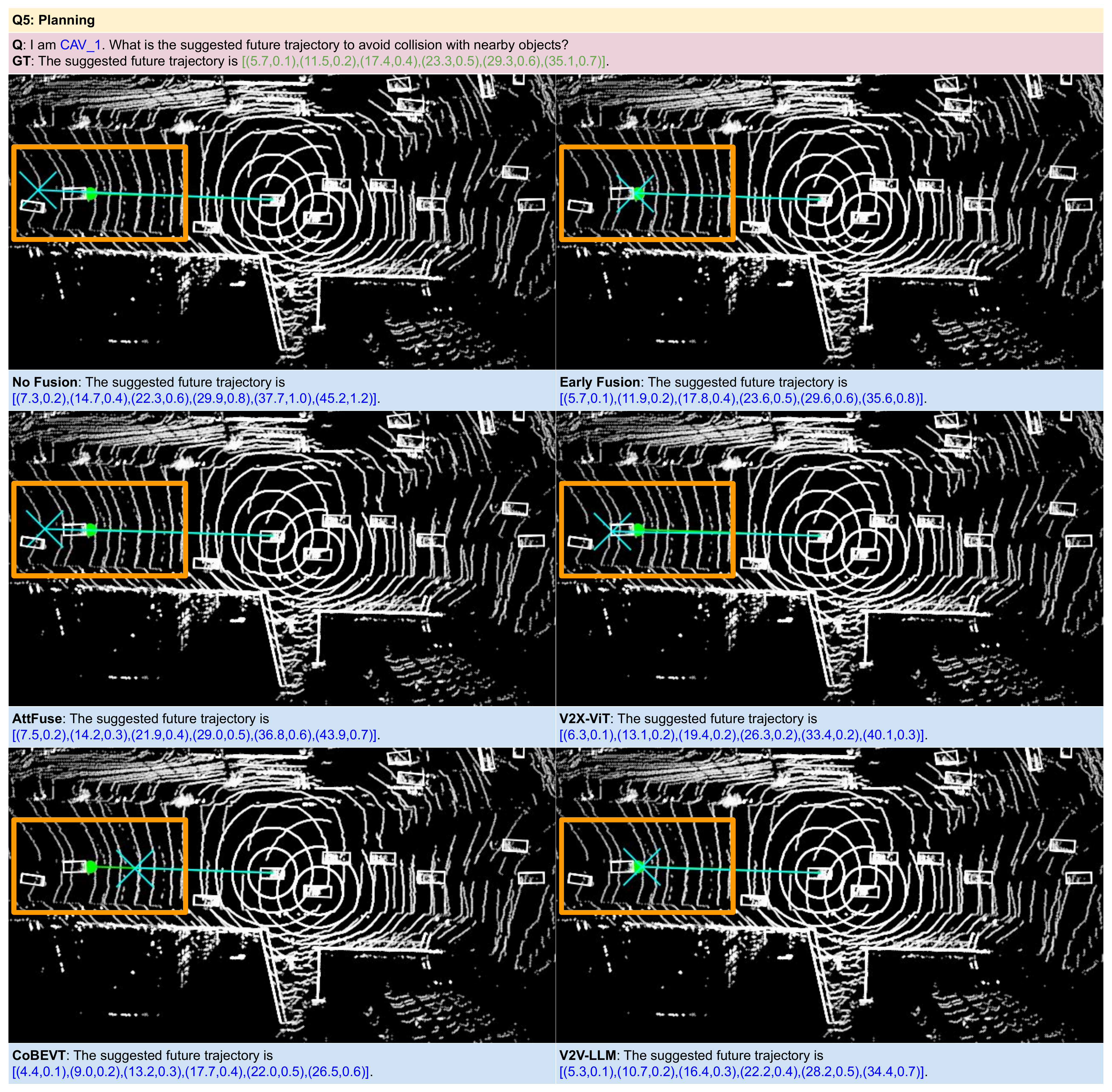}
\caption[]
        {\namemethod~and baseline methods' \textit{planning} results on \namedataset's \namevsplit~testing set.
         \textcolor{Green}{Green curve}: future trajectories in ground-truth answers. \textcolor{Green}{Green $\circ$}: ending waypoints in ground-truth answers. \textcolor{cyan}{Cyan curve}: model planning outputs. \textcolor{cyan}{Cyan $\times$}: ending waypoints in model outputs.} 
        \label{fig:supp_q5_2}
        \vspace{-5pt}
\end{figure*}
\begin{figure*}[!t]
\centering
\includegraphics[width=1\textwidth]{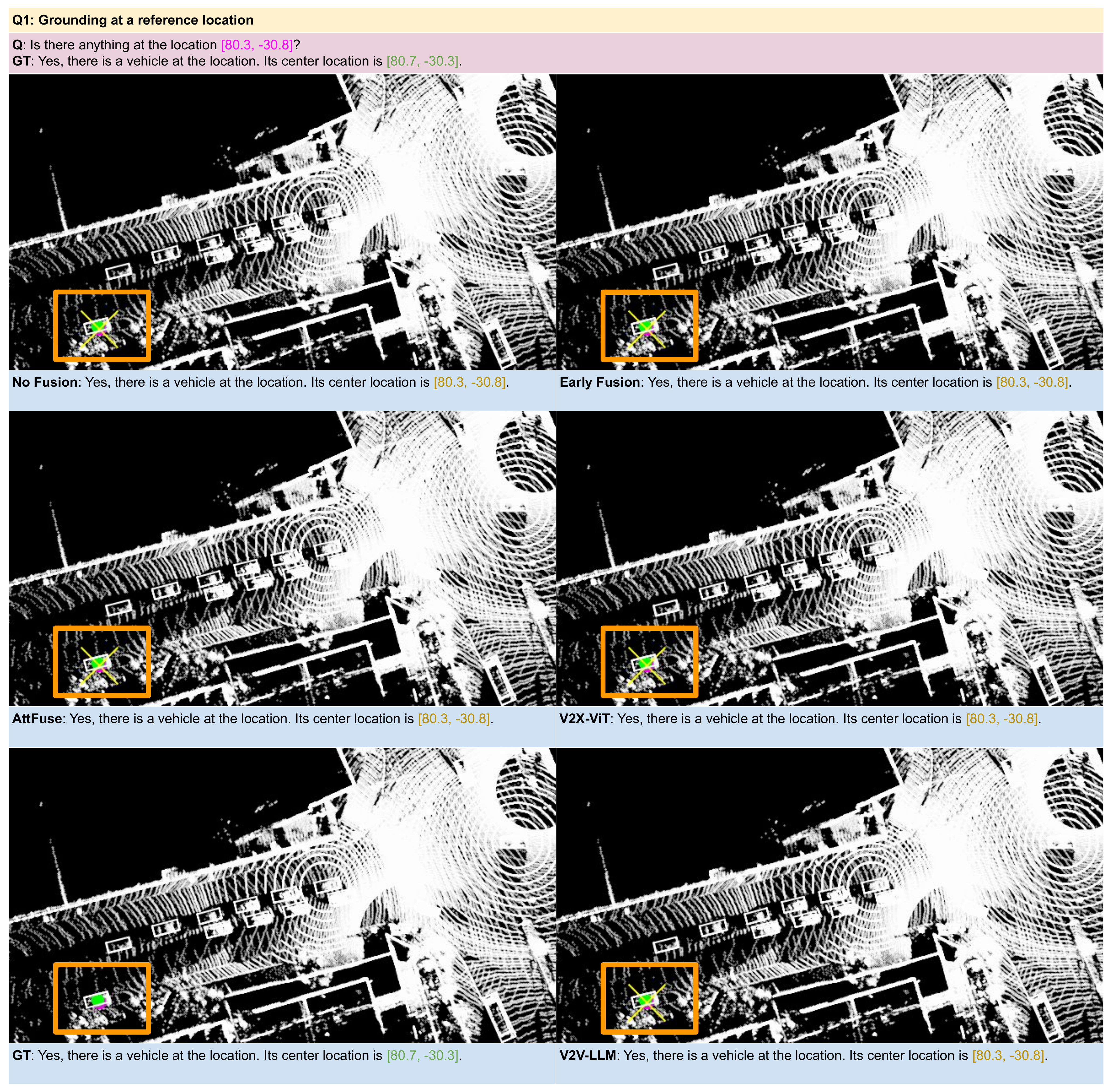}
\caption[]
        {\namemethod~and baseline methods' \textit{grounding} results on \namedataset's \namexsplit~testing set.~\textcolor{magenta}{Magenta $\circ$}: reference locations in questions. \textcolor{olive}{Yellow $+$}: model output locations. \textcolor{Green}{Green $\circ$}: ground-truth answers.} 
        \label{fig:supp_v2x_q1}
        \vspace{-5pt}
\end{figure*}
\begin{figure*}[!t]
\centering
\includegraphics[width=1\textwidth]{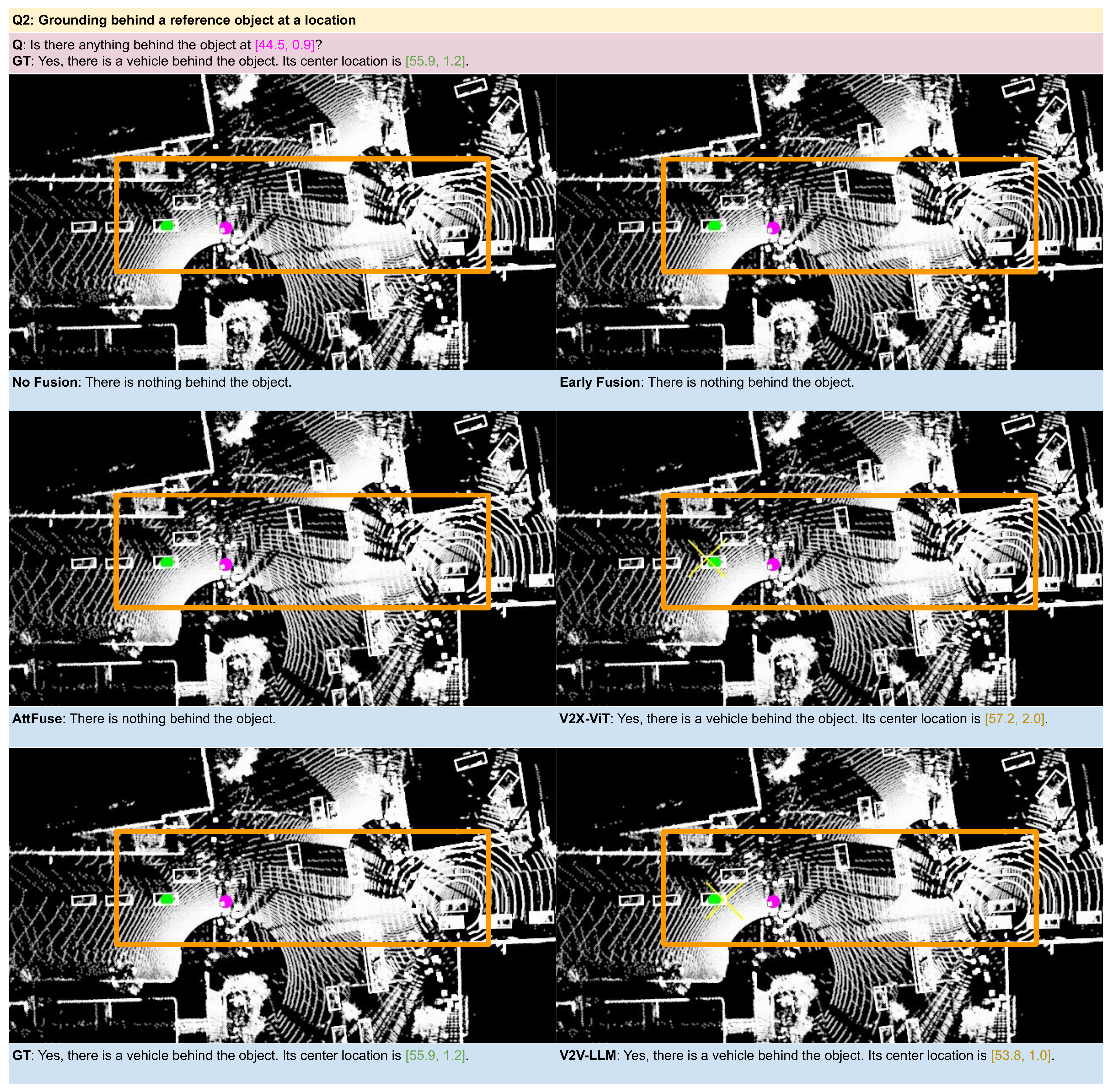}
\caption[]
        {\namemethod~and baseline methods' \textit{grounding} results on \namedataset's \namexsplit~testing set.~\textcolor{magenta}{Magenta $\circ$}: reference locations in questions. \textcolor{olive}{Yellow $+$}: model output locations. \textcolor{Green}{Green $\circ$}: ground-truth answers.} 
        \label{fig:supp_v2x_q2}
        \vspace{-5pt}
\end{figure*}
\begin{figure*}[!t]
\centering
\includegraphics[width=1\textwidth]{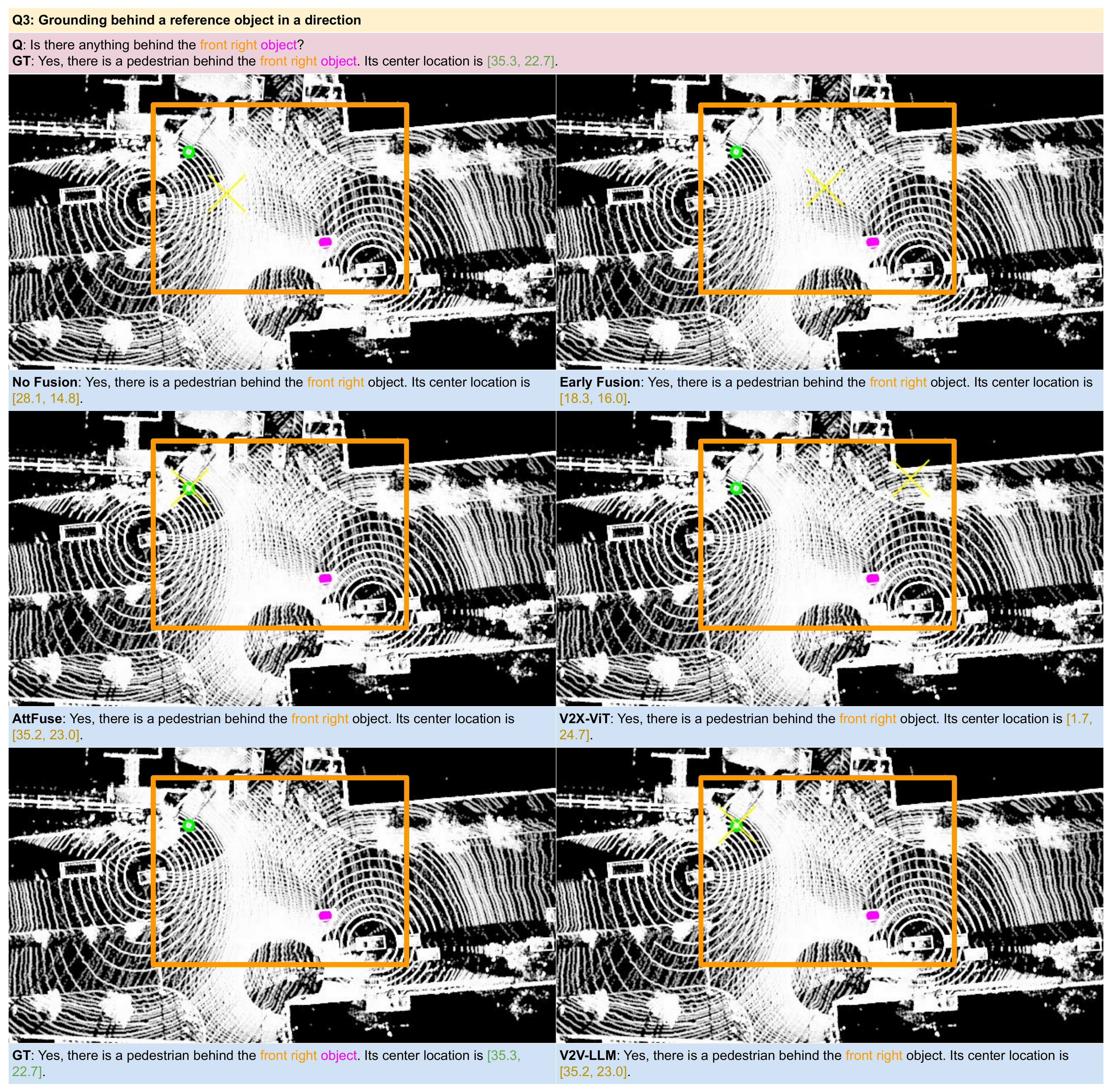}
\caption[]
        {\namemethod~and baseline methods' \textit{grounding} results on \namedataset's \namexsplit~testing set.~\textcolor{magenta}{Magenta $\circ$}: reference locations in questions. \textcolor{olive}{Yellow $+$}: model output locations. \textcolor{Green}{Green $\circ$}: ground-truth answers.} 
        \label{fig:supp_v2x_q3_1}
        \vspace{-5pt}
\end{figure*}
\begin{figure*}[!t]
\centering
\includegraphics[width=1\textwidth]{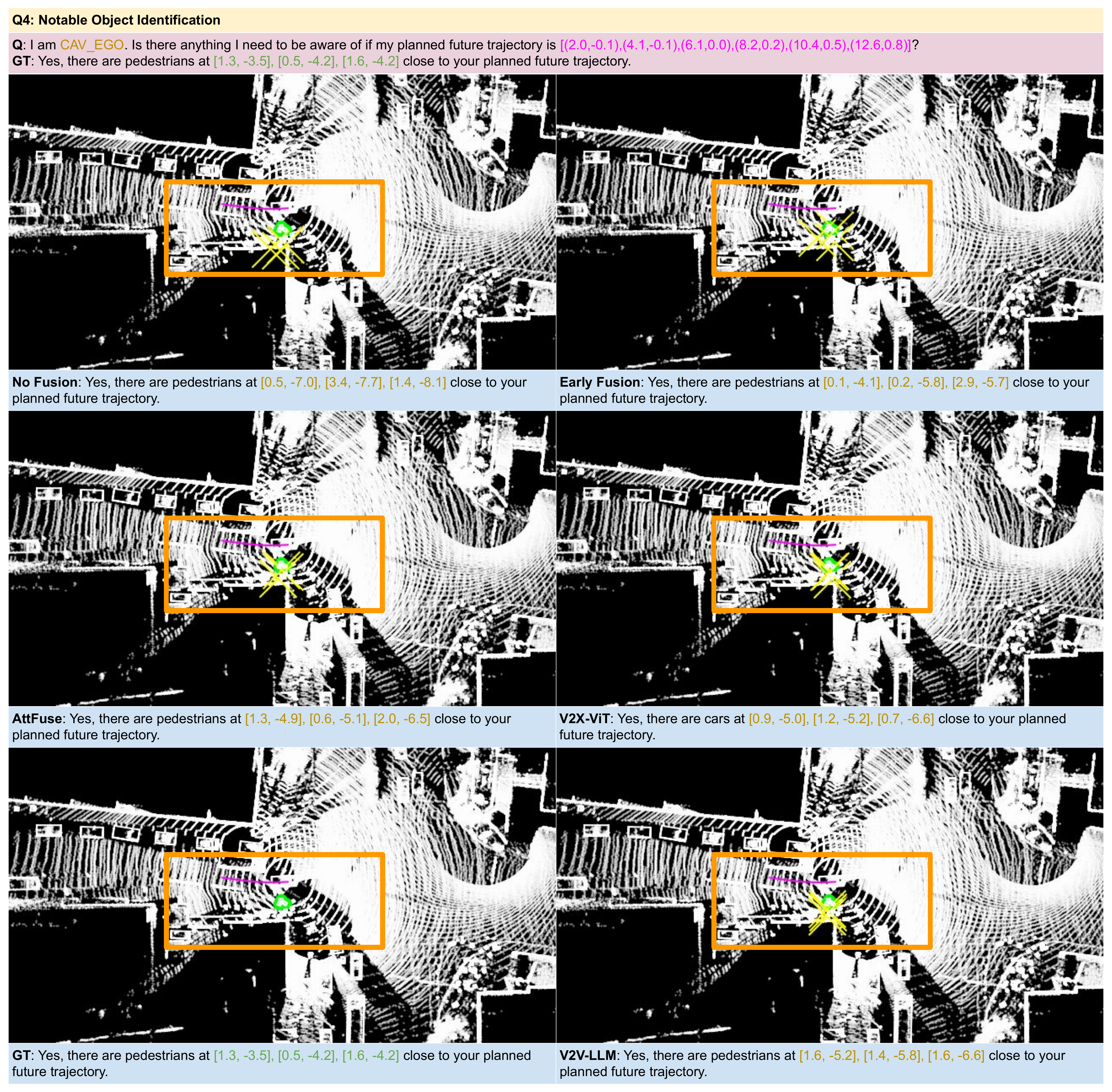}
\caption[]
        {\namemethod~and baseline methods' \textit{notable object identification} results on \namedataset's \namexsplit~testing set.~\textcolor{magenta}{Magenta curve}: planned future trajectories in questions. \textcolor{Green}{Green $\circ$}: ground-truth notable object locations. \textcolor{olive}{Yellow $+$}: model identification outputs.} 
        \label{fig:supp_v2x_q4_1}
        \vspace{-5pt}
\end{figure*}
\begin{figure*}[!t]
\centering
\includegraphics[width=1\textwidth]{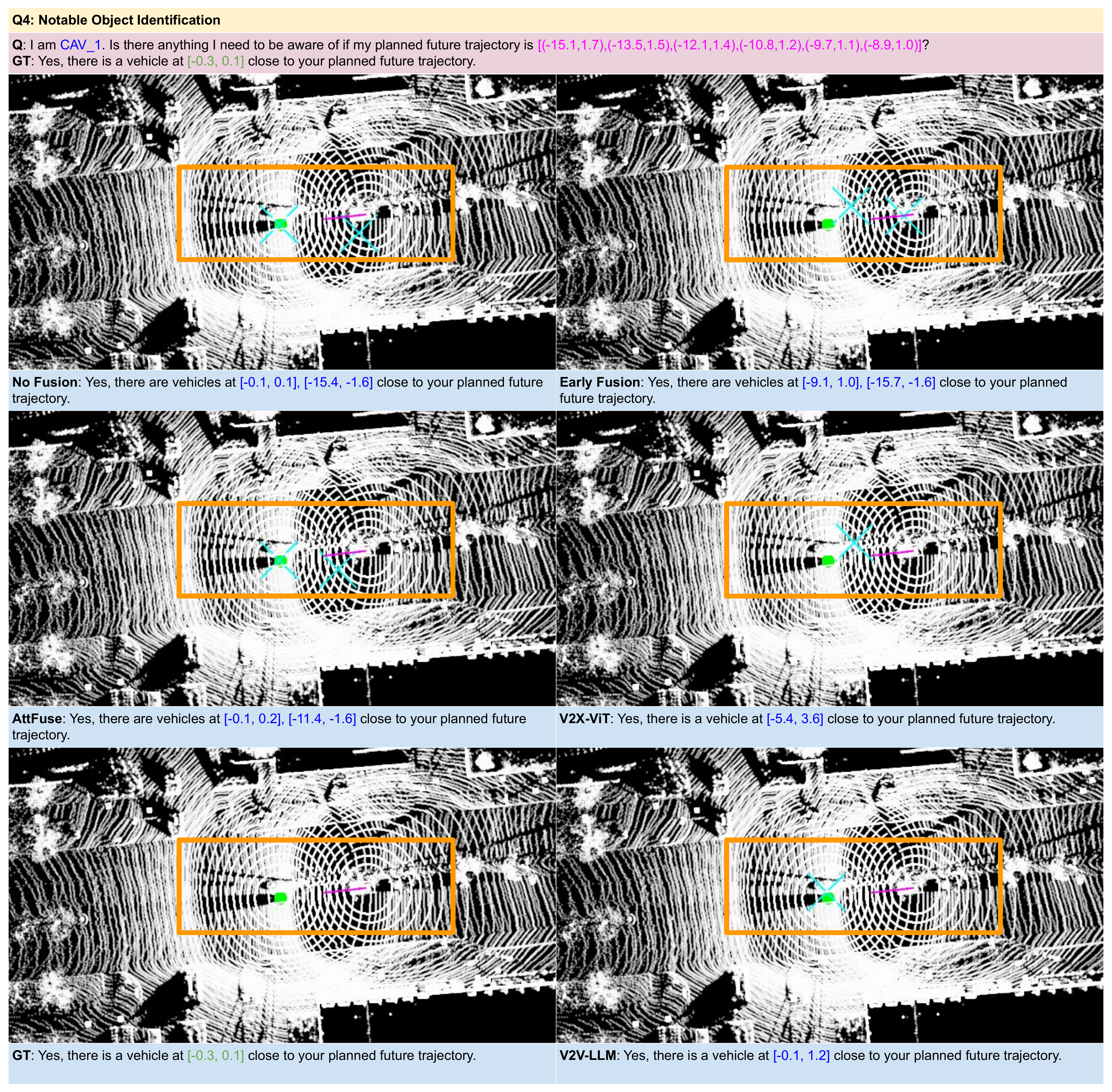}
\caption[]
        {\namemethod~and baseline methods' \textit{notable object identification} results on \namedataset's \namexsplit~testing set.~\textcolor{magenta}{Magenta curve}: planned future trajectories in questions. \textcolor{Green}{Green $\circ$}: ground-truth notable object locations. \textcolor{cyan}{Cyan $\times$}: model identification outputs.} 
        \label{fig:supp_v2x_q4_2}
        \vspace{-5pt}
\end{figure*}
\begin{figure*}[!t]
\centering
\includegraphics[width=1\textwidth]{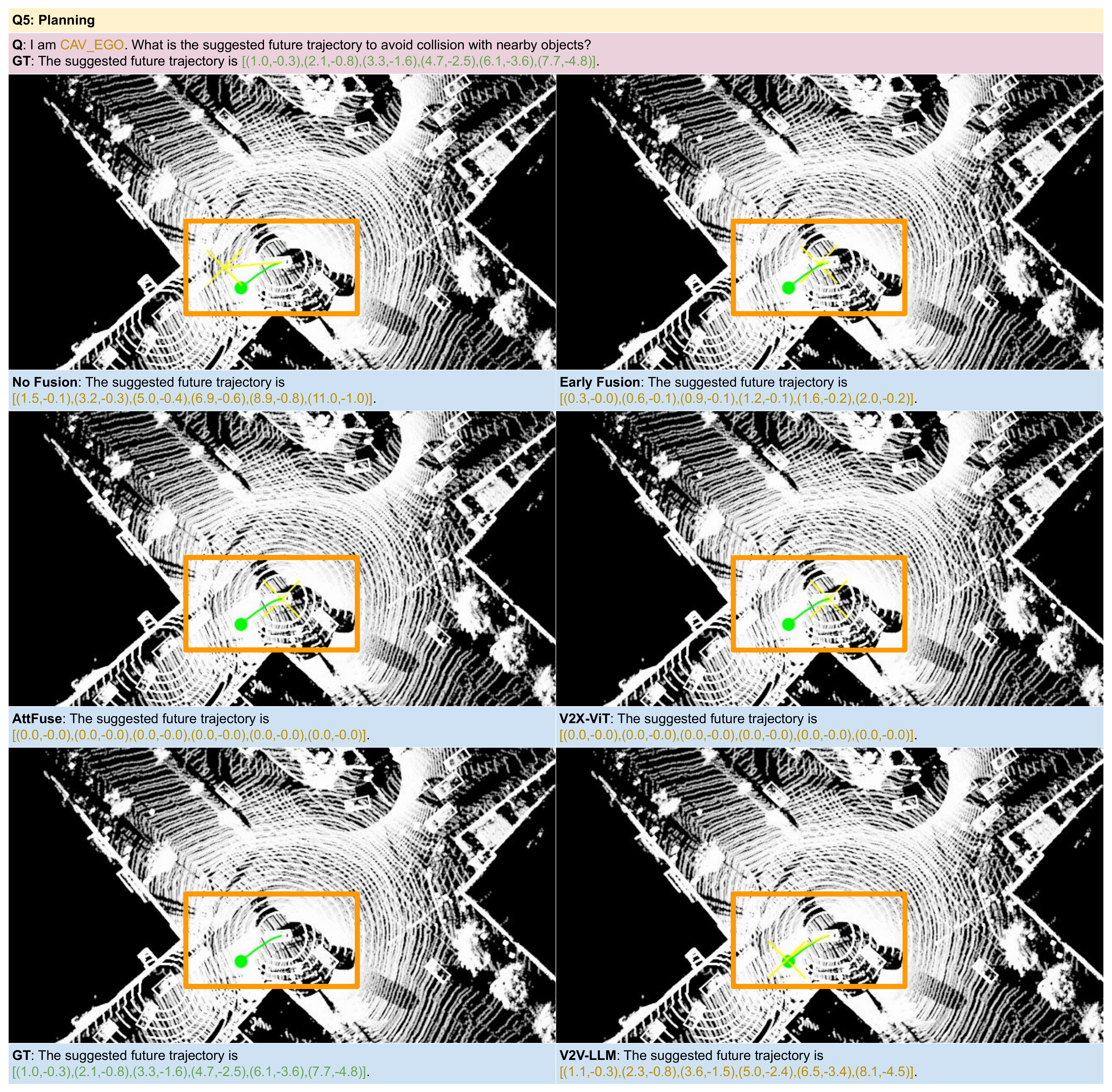}
\caption[]
        {\namemethod~and baseline methods' \textit{planning} results on \namedataset's 
\namexsplit~testing set.
         \textcolor{Green}{Green curve}: future trajectories in ground-truth answers. \textcolor{Green}{Green $\circ$}: ending waypoints in ground-truth answers. \textcolor{olive}{Yellow curve}: model planning outputs. \textcolor{olive}{Yellow $\times$}: ending waypoints in model outputs.} 
        \label{fig:supp_v2x_q5_1}
        \vspace{-5pt}
\end{figure*}
\begin{figure*}[!t]
\centering
\includegraphics[width=1\textwidth]{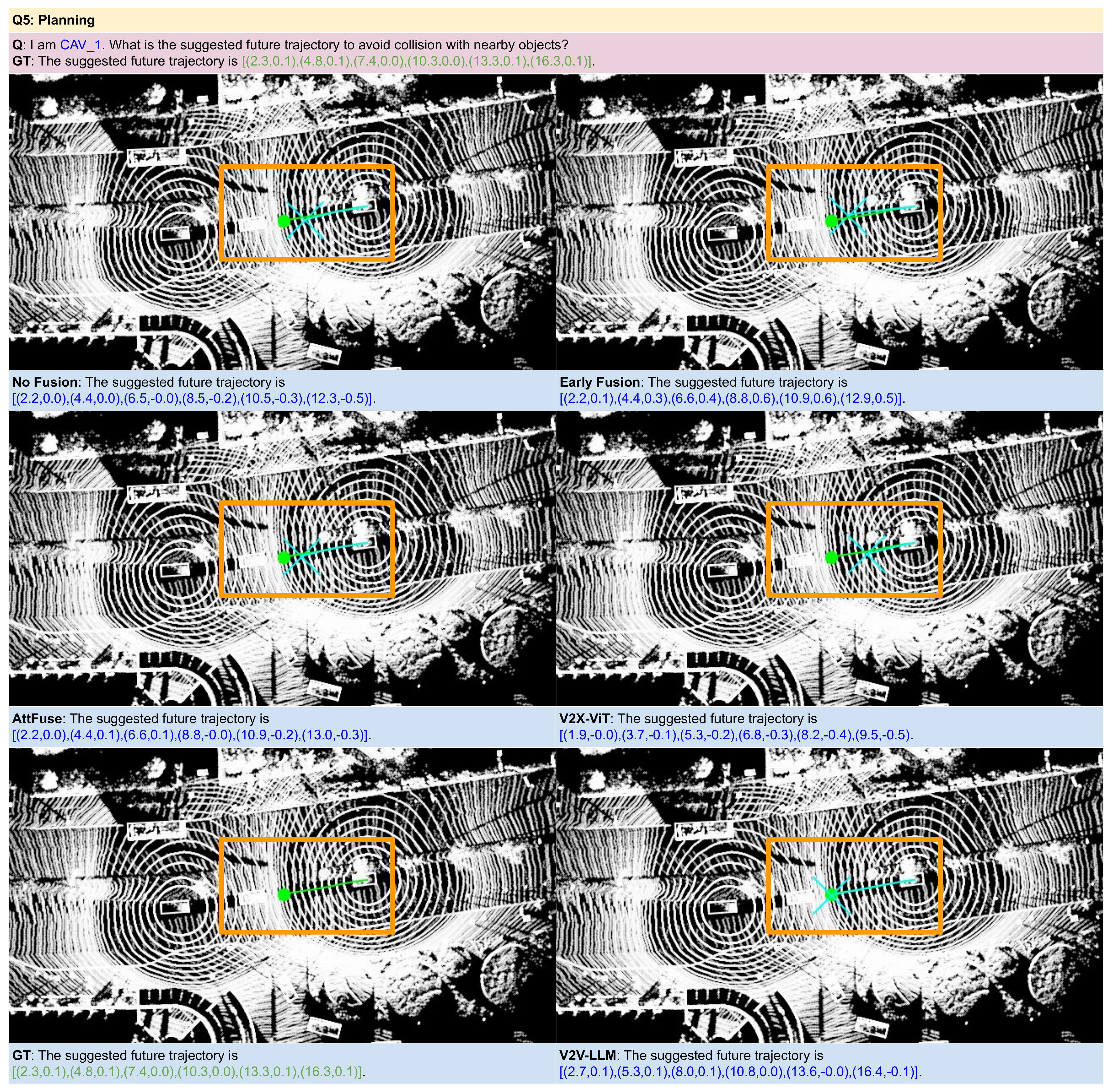}
\caption[]
        {\namemethod~and baseline methods' \textit{planning} results on \namedataset's 
\namexsplit~testing set.
         \textcolor{Green}{Green curve}: future trajectories in ground-truth answers. \textcolor{Green}{Green $\circ$}: ending waypoints in ground-truth answers. \textcolor{cyan}{Cyan curve}: model planning outputs. \textcolor{cyan}{Cyan $\times$}: ending waypoints in model outputs.} 
        \label{fig:supp_v2x_q5_2}
        \vspace{-5pt}
\end{figure*}

We show more qualitative results of our proposed \namemethod~and other baseline methods in the testing set of \namedataset's grounding task in Figures \ref{fig:supp_q1} to \ref{fig:supp_q3_2}, notable object identification task in Figures. \ref{fig:supp_q4_1} to \ref{fig:supp_q4_2}, and planning task in Figures \ref{fig:supp_q5_1} to \ref{fig:supp_q5_2}. The baseline methods include no-fusion, early-fusion, and intermediate-fusion: AttFuse~\cite{xu2022opencood}, V2X-ViT~\cite{xu2022v2xvit}, and CoBEVT~\cite{xu2022cobevt}.
Results of \namexsplit~can be seen in Figs. \ref{fig:supp_v2x_q1} to \ref{fig:supp_v2x_q5_2}.
In general, our proposed ~\namemethod's outputs are closer to the ground-truth answers, in comparison to other baseline methods' results.

\section{Limitation}
\begin{figure*}[!t]
\centering
\includegraphics[width=1\textwidth]{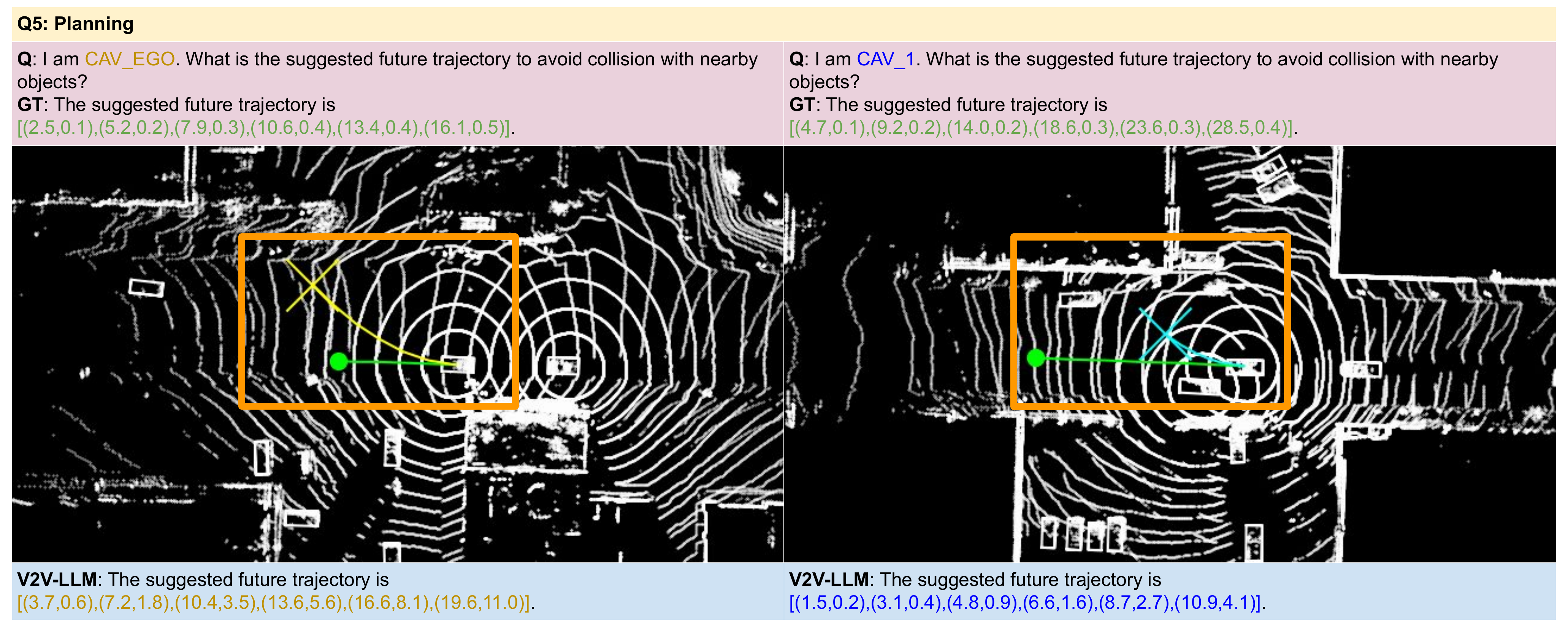}
\caption[]
        {Failure cases of \namemethod's \textit{planning} results on \namedataset's testing set.
         \textcolor{Green}{Green curve}: future trajectories in ground-truth answers. \textcolor{Green}{Green $\circ$}: ending waypoints in ground-truth answers. \textcolor{olive}{Yellow curve} and \textcolor{cyan}{Cyan curve}: model planning outputs corresponding to \textcolor{olive}{CAV\_EGO} and \textcolor{cyan}{CAV\_1}, respectively. \textcolor{olive}{Yellow $\times$} and \textcolor{cyan}{Cyan $\times$}: ending waypoints in model outputs corresponding to \textcolor{olive}{CAV\_EGO} and \textcolor{cyan}{CAV\_1}, respectively.} 
        \label{fig:supp_q5_failure}
        \vspace{-5pt}
\end{figure*}

Figure \ref{fig:supp_q5_failure} shows failure cases of \namemethod's \textit{planning} results on \namedataset's testing set. In a few frames, the model generates future trajectories in the lane of the opposite traffic direction. A potential solution and future work is to include HD map information as additional input to the model. Currently, this approach is not feasible because the base dataset V2V4Real~\cite{xu2023v2v4real} has not released its HD map to the public.

\end{document}